\documentclass[twocolumn,DIV32,10pt]{scrartcl}
 \pdfoutput=1
\usepackage{ctable}
\usepackage{float}
\usepackage{graphicx}
\usepackage{amssymb}
\usepackage{array}
\usepackage{url}
\usepackage{multirow}
\usepackage[round]{natbib}
\usepackage{algorithm}
\usepackage{algorithmic}
\usepackage{subfig}
\usepackage{gensymb}
\usepackage{palatino}
\usepackage{amsmath}
\usepackage{tweaklist}
\usepackage{fixltx2e}
\usepackage{sidecap}
\usepackage[absolute]{textpos}

\usepackage{pifont}
\newcommand{\cmark}{\ding{51}}%
\newcommand{\xmark}{\ding{55}}%

\newcommand{\Alpha}{\mathrm{A}}
\newcommand{\Frequency}{\mathrm{F}}

\renewcommand{\itemhook}{\setlength{\topsep}{0pt}%
  \setlength{\itemsep}{0pt}}

\date{}
\begin{document}
\begin{textblock}{15}(0.4,0.4)

\noindent
\large
 D. Tarapore and J.B. Mouret. Evolvability signatures of generative encodings: beyond standard performance benchmarks. \emph{Information Sciences}, 2015 (in press). 
\end{textblock}
\title{Evolvability signatures of generative encodings: beyond standard performance benchmarks}

\author{Danesh Tarapore and Jean-Baptiste Mouret
\thanks{Danesh Tarapore and Jean-Baptiste Mouret are with the ISIR, Universit\'e Pierre et Marie Curie-Paris 6, CNRS UMR 7222, F-75252, Paris
Cedex 05, France. Contact: daneshtarapore@gmail.com}
}

\maketitle

\begin{abstract}
\bfseries Evolutionary robotics is a promising approach to autonomously synthesize machines with abilities that resemble those of animals, but the field suffers from a lack of strong foundations. In particular, evolutionary systems are currently assessed solely by the fitness score their evolved artifacts can achieve for a specific task, whereas such fitness-based comparisons provide limited insights about how the same system would evaluate on different tasks, and its adaptive capabilities to respond to changes in fitness (e.g., from damages to the machine, or in new situations). To counter these limitations, we introduce the concept of ``evolvability signatures'', which picture the post-mutation statistical distribution of both behavior diversity (how different are the robot behaviors after a mutation?) and fitness values (how different is the fitness after a mutation?). We tested the relevance of this concept by evolving controllers for hexapod robot locomotion using five different genotype-to-phenotype mappings (direct encoding, generative encoding of open-loop and closed-loop central pattern generators, generative encoding of neural networks, and single-unit pattern generators (SUPG)). We observed a \textit{predictive} relationship between the evolvability signature of each encoding and the number of generations required by hexapods to adapt from incurred damages. Our study also reveals that, across the five investigated encodings, the SUPG scheme achieved the best evolvability signature, and was always foremost in recovering an effective gait following robot damages. Overall, our evolvability signatures neatly complement existing task-performance benchmarks, and pave the way for stronger foundations for research in evolutionary robotics.
\end{abstract}


\section{Introduction}
Evolutionary robotics (ER) is a promising approach to achieve one of the prominent long-term goals of artificial intelligence research: creating machines with the adaptive and cognitive abilities of animals. Since the eighties, the ER field has made amazing progress to both design sophisticated artifacts and to endow machines with impressive adaptive abilities. For instance, it allows for the automated construction of modular, three-dimensional, physically locomoting robots, \citep{hornby2003generative}, to synthesize neural networks to control robot behaviors (e.g., \cite{lehman2011improving,mouret2012,Kubota2005403,Santos2001127}), and discover a multitude of walking gaits for multilegged robots following unforeseen mechanical damages \citep{bongard2006resilient,koos2013fast,cully2013}. However, even the most advanced evolved artifacts are still far behind the state of the art in mainstream robotics \citep{siciliano2008springer,bongard2013evolutionary}: conventionally engineered robots are capable of operating successfully in a wide variety of indoor and outdoor environments (e.g., locomotion with the BigDog quadruped robot, \cite{raibert2008bigdog}), whereas the best evolved robots are still only capable of simplistic behaviors (e.g., walking in a straight line on a flat terrain, or avoiding obstacles in an enclosed indoor arena). To progress further, ER needs to go beyond the mere ``stamp collecting'' of proofs of concept, evident in the infancy of many scientific fields \citep{hayes2004undisciplined}, and build strong theoretical and methodological foundations for future research. The objective of the present study is to move in this direction.

In most ER studies, fitness comparison is the main instrument used to compare different evolutionary systems and assess their progress. Such a benchmark-based comparative approach has led to incremental improvements in the robot's performance in specific tasks (e.g., for multilegged robot locomotion, the inclusion of evolved gaits on the commercial release of Sony's AIBO \cite{hornby2005autonomous,valsalam2008modular}, and the progressive improvements in walking speed of the QuadraBot \cite{yosinski2011,lee2013}), and is sufficient if excelling at the given function is the ultimate goal for the robot. Nonetheless, if the evaluated task is treated as a tool to compare different evolutionary systems, and as a stepping stone to harder problems, then a mere comparison of performance does not suffice. This is because such a methodology of comparison only provides a very limited amount of information about the behavior of the system. In particular, it does not provide any insights on, (i)~how efficiently does the evolutionary process explore the search space (e.g., can it also lead to solutions for other similar tasks, or is it biased to the type of solutions useful only for a very specific task?), and (ii)~what capabilities are provided to the evolved population to respond to novel situations (e.g., an unexpected breakage of the multilegged robot's limbs, or changes to its weight distribution). Furthermore, while adaptive evolutionary systems utilize a variety of population-diversity maintenance methods to operate in changing environments \cite{Branke2005}, they are mostly concerned with numerical optimization problems (e.g., \cite{Jong99}), and constrained to fitness-based indices to evaluate available approaches~\cite{weicker2002}. In summary, there is a need for additional metrics when comparing evolutionary systems, especially if one is interested in the adaptive abilities provided by evolution.

In benchmark-based comparative approaches, the fitness value in an evolutionary system is often used as a proxy for the \emph{evolvability} provided by the system \cite{gruau1994automatic,komosinski2001comparison,hornby2003generative,clune2009evolving} --- the capacity of the evolved population to rapidly adapt to novel environments~\cite{Ting10}. However, such a fitness-based proxy provides little information on the potential of the evolutionary system to generate novel phenotypes, and consequently rapidly adapt to new, untested environments. While fitness landscape models can provide interesting insights on search difficulty in the Genotype-to-Fitness map \citep{provine1989sewall,wright1932roles}, the model’s 3D landscape can be deceptive when analyzing  highly multidimensional genotypes \citep{kaplan2008end,gavrilets1997evolution,mccandlish2011visualizing}. Additionally, in NK fitness landscape models \citep{stuart1993origins,tomassini2008complex}, the value of K that controls the degree of epistasis is not easily transferable to more complex and open-ended Genotype-to-Fitness mappings. Also, the individual solutions in all these models are positioned in the landscape solely based on their measured fitness. In the present paper, to counter the limitations of the fitness measure, we introduce a new evolvability metric that features both the quality and quantity of phenotypic variation following genetic change. With this new metric, we can \emph{visualize} evolvability in the behavior-diversity/performance space and \textit{predict} the performance of the population in previously untested environments\footnote{A preliminary study on our approach to visualize evolvability is published in a conference paper \cite{tarapore2014}.}. Such predictive insights on the adaptive characteristics of evolved individuals is  particularly important, since it is difficult if not impossible to consider and evaluate a priori every possible scenario the robot may encounter during its operation.
We employ our new approach to ``signaturize'' evolvability to compare many different encodings of controllers extracted from the literature. Numerous encodings have been proposed in ER, taking inspiration from natural developmental processes, in particular, to evolve control systems for robots (e.g., \cite{gruau1994automatic,kodjabachian1998evolution,clune2009evolving,cheney2013unshackling,lee2013,lewis1992genetic,morse2013}). Given the multitude of available encodings, it is crucial to compare them and understand their differences, so that the ER community can focus on the most promising ones. In the selection of encodings investigated in our study, both direct and generative schemes are considered. Direct encodings encompass a one-to-one mapping between genes and phenotypic traits, and are the simplest form of encoding thus serving as a reference for comparison (e.g., \cite{koos2013fast}). We also evaluate the more complex generative encodings characterized by a one-to-many mapping between genes and phenotypic traits, i.e., a single gene describes several phenotypic traits \citep{stanley2002evolving,stanley2007compositional}. These state of the art encodings are expected to exploit geometric information of the robot morphology to generate regular and modular phenotypic patterns (e.g., \citep{stanley2009hypercube,clune2011,morse2013}). 

Overall, we investigate five encodings for the classical ER problem of legged robot locomotion \cite{lewis1992genetic,gruau1994automatic,hornby2005autonomous,clune2009evolving,bongard2006resilient,Clune2009.1569901.1569995,clune2011,yosinski2011,koos2013fast,lee2013}: (1) open-loop central pattern generator (CPG) evolved with a direct encoding, (2) open-loop CPG based on non-linear oscillators \cite{crespi2013salamandra}, evolved with a Compositional Pattern Generator (CPPN) \cite{stanley2007compositional}, (3) closed-loop CPG evolved with a CPPN, (4) artificial neural network (ANN) evolved with CPPN, inspired by HyperNEAT \cite{stanley2009hypercube,clune2011}, and (5) the recently introduced single-unit pattern generator (SUPG) \cite{morse2013}. For all these encodings, the pertinent questions are the same: are these encodings facilitating evolvability, and are the encoded individuals capable of adapting rapidly to novel situations? Furthermore, does the inclusion of a sensory feedback mechanism improve the evolvability provided, and the adaptive capabilities of the individual? To both answer these questions and evaluate the relevance of our measure of evolvability, our experiments are divided into two phases: first, we compare the evolvability signature obtained with each encoding, and consequently predict their adaptability to novel scenarios, then we evaluate the accuracy of our predictions by analyzing the ability of each encoding to effectively deal with the new scenarios (here, when some of the robot's legs are damaged).

\section{Related work}
\label{sec:relwork}

The section reviews the different Genotype-to-Phenotype mapping schemes implemented in evolutionary computation studies. We also review the functional characteristics of encodings that facilitate evolvability both in natural and artificial systems, and the empirical methods available to estimate evolvability. The encodings compared in this study are described in detail in Sec.~\ref{sec:encanal}.

\subsection{Encoding schemes}
A mapping from the Genotype-to-Phenotype is a model of the process that ``develops'' an individual's phenotype from the information available in the genotype \citep{johannsen1911}. In natural organisms, the elaborate and intricate developmental systems comprise intra-cellular mechanisms of transcription and translation of proteins \citep{freeland1998genetic,novozhilov2007evolution} and regulation of gene expression \citep{albertsmolecular}, and the inter-cellular mechanisms of cell differentiation and specialization \citep{michod2001cooperation,grosberg2007evolution}. Furthermore, these complex developmental processes are a result of billions of years of natural selection \citep{javaux2010organic}.

Many different genetic encodings have been implemented in the field of evolutionary computation, from the early deterministic and fixed Genotype-to-Phenotype mappings \citep{friedberg1958learning,koza1992genetic}, to the recent, plastic and more naturalistic schemes \citep{stanley2007compositional,tonelli2013relationships}. In this review, we broadly classify the available encodings based on their degree of genotypic reuse, i.e., the extent to which genes are allowed to be reused in developing the phenotype, into direct encodings, and generative or developmental encodings (for detailed review see \cite{stanley2003taxonomy}).

\subsubsection{Direct encodings}
Direct encodings are characterized by the complete absence of any genotypic reuse, wherein each gene of the genotype is utilized at most once to determine the phenotype \citep{hornby2002creating}. Such a mapping has been used since the advent of evolutionary computation and genetic programming research \citep{friedberg1958learning}, and is the most widely used encoding scheme in the field. Moreover, direct encodings are easy to implement, and have been applied successfully for the evolution of various robot behaviors such as locomotion gaits for multilegged and tensegrity platforms (e.g., \cite{lewis1992genetic,tellez2006evolving,koos2013fast,iscen2013controlling}), navigation and obstacle avoidance for wheeled robots (e.g., \cite{floreano2001}), body-brain evolution in artificial life systems \citep{hornby2002creating}, and cooperative foraging in robot swarms (e.g., \cite{waibel2006division2009,Dorigo2004,kernbach2013handbook}).

Researchers in evolutionary robotics have recently sought to evolve increasingly complex artifacts \citep{eiben2014grand}. However, in using the direct encoding scheme, as the size of the genotype grows linearly, the possible combination of allelic values for the genotype (the solution search space) grows exponentially \citep{yao1999evolving}. The consequent scalability problem prevents the usage of a direct Genotype-to-Phenotype mapping to evolve solutions for complex problems. For example, the human brain consists of approximately $86$ billion neurons and $100$ trillion neural connections \citep{azevedo2009equal}, and therefore successfully evolving a directly encoded \textit{in silico} brain-like cognitive device is almost surely impossible.

\subsubsection{Generative encodings}
Natural systems are much more sophisticated than the state of the art solutions from artificial evolution. Inspired by the complexity of biological systems \citep{Lipson04principlesof}, researchers have abstracted the underlying developmental processes, to formulate generative or developmental Genotype-to-Phenotype maps for artificial systems~(e.g., \cite{bongard2001repeated,komosinski2001comparison,gruau1994automatic,hornby2002creating,stanley2007compositional,mouret2010importing,Devert2011}). The generative encodings can be broadly defined as, \textit{encodings that map the genotype to the phenotype through a process of growth from a simple genotype in a low-dimensional search space to a complex phenotype in a high-dimensional space}. These encodings have been applied successfully to various application problems, from allowing a computer to design antennas for satellites \citep{hornby2006automated}, designing tables \citep{hornby2005measuring,hornby2004functional} and creating tessellating tile shapes \citep{bentley1999three}, to evolving locomotion gaits for both soft robots (e.g., \cite{cheney2013unshackling}) and conventional multilegged robots (e.g., \cite{clune2009evolving}).

The many implemented generative encodings are all commonly characterized by a one-to-many mapping between an element of the genotype and many elements of the phenotype. In the resulting Genotype-to-Phenotype mappings, each gene of the genotype encodes for many phenotypic traits, thus allowing for \textit{genotype reuse}. Consequent to the capability of the encoding to reuse parts of the genotype to affect different phenotypes, generative encodings in comparison to direct encodings, (i) exhibit higher efficiency in representing complex phenotypes, (ii) operate in a more tractable solution search space, (iii) scale well to large phenotypic spaces, and (iv) are capable of generating regular and modular control architectures~\citep{stanley2003taxonomy,Clune2009.1569901.1569995,stanley2009hypercube}. However, while direct encodings are easy to implement, generative encodings follow more complex implementations, determined by their level of abstraction of development.

Existing generative encodings model the underlying natural developmental processes at several different levels of abstraction, ranging from the low-level cell chemistry simulations to the high-level grammatical approaches~\citep{stanley2007compositional}. The cell chemistry methods simulate the local intra-cellular and inter-cellular interactions between genes and protein products, modulated by signals from a gene regulatory network~(e.g.,~\cite{furusawa1998emergence}). These microscopic approaches are based on the philosophy that the vital functions that allow development to assemble complex phenotypes are located in the low-level cellular interactions occurring in a developing embryo. By contrast, the grammatical approaches simulate development with a set of high-level symbol replacement rules~(e.g.,~\cite{mjolsness1991connectionist}). These approaches grow a final structure from a single seed symbol, by the repeated application of grammar rules on specified target symbols. Consequently, local interactions and temporal unfolding control the phenotype development through the grammars~\citep{stanley2003taxonomy}.

In most studies comparing encoding schemes, the generative encoded individuals frequently outperform their directly encoded counterparts for a range of diverse tasks such as, designing 3D objects (e.g.,~\cite{hornby2005measuring,clune2011evolving}), game playing (e.g.,~\cite{reisinger2007acquiring,gauci2010autonomous}), pattern matching (e.g., \cite{clune2011}), and multilegged robot locomotion~(e.g.,~\cite{hornby2002creating,seys2007genotype}). The generatively encoded individuals achieve a better task performance, and a faster rate of evolution (e.g.,~see~\cite{gruau1994automatic,komosinski2001comparison}). Furthermore, generative encodings also attain a higher proportion of beneficial genetic mutations (e.g.,~\cite{clune2009evolving,reisinger2007acquiring}), and are capable of exploring a larger range of phenotypes from genetic change~(e.g.,~\cite{reisinger2005towards,lehman2013evolvability}). 

\subsection{Evolvability}
The process of evolution in natural systems comes from the cooperation of, (i) exploratory genotypic variation, (ii) the corresponding phenotypic variation, and (iii) selection operators that preserves the improvements in heritable phenotypic traits over previous generations. The crucial coordination between these three forces yields the evolvability of an evolutionary system \citep{Alberch1991,Ting10,pavlicev2012coming}. 

The first formal definition of evolvability stems from research in computer science. In experiments with optimization algorithms using genetic programming, Lee Altenberg defined evolvability as ``the ability of a population to produce variants fitter than any yet existing'' \citep{altenberg1994evolution}. In natural evolutionary systems, Kirschner and Gerhart \citep{kirschner1998evolvability} describe evolvability, also called evolutionary adaptability, as ``the capacity to generate heritable, selectable phenotypic variation''. Marrow \citep{Marrow99evolvability} considers evolvability as a characteristic relevant to both artificial and natural evolutionary systems, and viewed as the capability of a population to evolve. In a summary of results from both evolutionary biology and evolutionary computer science, Wagner and Altenberg \citep{wagner1996perspective} view evolvability as ``the ability of random variations to sometimes produce improvements''. These incremental improvements are critically dependent on the Genotype-to-Phenotype encoding. Mappings facilitating evolvability, confer on the individual a robustness to lethal mutations, and exhibit a modular architecture wherein genes preferably only affect traits with the same function~\citep{pavlicev2012coming}. 

Although the concept of evolvability is still very much under discussion, for our study we adopt the definition pertaining to adaptability, and the generation of major phenotypic breakthroughs~\cite{pigliucci2008evolvability,clune2013evolutionary}: \emph{Evolvability is the capability of a population to rapidly adapt to novel and challenging environments.}

The measurement of evolvability conferred by an encoding is a complex and difficult problem. Phenotypic fitness or task performance is a directly observable measure, and a criteria for selection. However, the \emph{potential} to generate a better fitness, evolvability, is a less tangible type of observable and is more difficult to measure \citep{Ting10}. While a formal method to quantify evolvability has not yet been agreed upon in the literature, some empirical methods have been proposed notwithstanding.

In Gerhart and Kirschner's theory of facilitated variation \cite{gerhart2007theory}, which unifies most earlier findings of cellular and developmental processes with characteristics of evolvability, the capacity of an individual to evolve is considered to have two functional components: (i)~to curtail the proportion of lethal mutations; and (ii)~to decrease the number of mutations necessary to evolve diverse or novel phenotypes. Nonetheless, most studies measuring evolvability focus mainly on one only of these two aspects. Most comparisons estimate evolvability solely as the proportion of mutations that are beneficial to an individual~(e.g., \cite{hornby2003generative,clune2011,hornby2002creating,clune2009evolving,reisinger2007acquiring}), and irrespective of the phenotypic novelty of the resultant offspring. In other work, Ib{\'a}{\~n}ez-Marcelo and Alarc{\'o}n \cite{ibanez2014topology} characterize evolvability as the number of viable mutations required to reach different phenotypes, but do not take into account the novelty or the amount of diversity between the different phenotypes. By contrast,~\cite{lehman2011improving,reisinger2005towards,lehman2013evolvability} quantify evolvability only on the basis of the amount of phenotypic diversity resulting from genetic change, usually without considering the quality of the change. However, both factors are essential to quantify evolvability, to discount for, (i)~mutations that generate very diverse phenotypes, but prove lethal to the organism, and (ii)~mutations resulting in small increments in performance that improve on a trait, but may not be able to generate novel phenotypes. Therefore, for our comparison between encodings, evolvability is visualized by characterizing both the nature of the genetic mutation, and the quantity of generated phenotypic variation.

\section{Evolvability signature}
\label{sec:evob}

In this study, the evolvability provided by a Genotype-to-Phenotype mapping is described by a distinct signature featuring information on the effect of genetic mutations on both, the \textit{quality} of mutated individuals (their viability), and the \textit{quantity} of generated phenotypic variation. Our signature pictures these two features as a statistical distribution of fitness (quality-feature) and behavior diversity (quantity-feature), following multiple independent, randomly sampled mutations. The two features are treated separately instead of being combined into a single quantitative measure of evolvability, to consider the trade-offs between them in their individual influence on evolvability \cite{deb2010multi}.

\textbf{Feature 1:~Deleteriousness of mutations.}~The first feature in our signature of evolvability is computed as the proportion decrease in the fitness of a mutated individual.

For an individual $i$ and the mutant $i^\prime$, we have,

\small
\begin{equation}
f_1 = \frac{F_i^\prime - F_i}{F_i} \label{eqn:evob_f1}
\end{equation}
\normalsize

\noindent where $F_i$ and $F_i^\prime$, are the fitness values before and after the application of a random genetic mutation, respectively.

The feature $f_1$ reflects the behavior quality following beneficial ($f_1 > 0$), neutral ($f_1 \approx 0$), and deleterious ($f_1 < 0$) genetic change. Additionally, mutations that prove lethal are associated with $f_1$ values less than $-1$, reflecting a $100\%$ or larger decrease in individual fitness.

\textbf{Feature 2:~Diversity of behaviors.}~Following the theory of facilitated variation \cite{gerhart2007theory}, the second feature in our signature of evolvability evaluates the diversity of phenotypes than can be reached from a given individual. The phenotype can here be understood in two ways: in an evolutionary biology perspective, the phenotype can describe both morphological traits and behaviors \cite{Arnold1992,Dawkins1999}, whereas in a evolutionary robotics perspective, only morphological traits are considered to be parts of the phenotype (e.g., the parameters and the topology of an evolved networks form the phenotype) \cite{stanley2003taxonomy}. The distinction between phenotype and behaviors avoids potential confusions when working on developmental encodings (genotype-phenotype maps), which focus on morphological traits, or when working on selective pressures, which often focus more on the behavior than on the representation \cite{doncieux2014}.

In the present study, we focus on the diversity of \emph{behaviors}, as done in evolutionary biology, because it best distinguishes promising individuals from the poor performers when evolving robot controllers \cite{lehman2011abandoning,mouret2012}. For instance, all the neural-networks that are not connected to the robot's actuators lead to the same stopped-robot behavior, whereas the morphological traits (synaptic parameters and topology of the neural-network) can be widely different. A second advantage of looking at behaviors instead of morphological traits is that the behavior representation can be independent of the implementation of the controller, thus allowing us to compare the evolvability of very different controllers like CPGs, neural networks, and SUPG controllers. 

Measuring behavioral differences recently received a lot of attention in evolutionary robotics because several experiments showed that explicitly encouraging the diversity of evolved behaviors helps to mitigate the issue of premature convergence \cite{mouret2011NS,mouret2012,doncieux2013behavioral,doncieux2014}. It is also the main driving force in the Novelty Search algorithm, which leads to high-performing individuals in deceptive domains by only searching for novel behaviors and disregarding task-fitness values \cite{lehman2011abandoning}. Following this interest in measuring behavioral differences, many behavioral diversity metrics have been proposed, ranging from task-specific metrics (e.g., difference between end points of a robot's trajectory), to more task-agnostic measures (e.g., differences in the robot sensory-motor flow), and various information theoretic measures~(detailed review in~\cite{doncieux2014} and in~\cite{mouret2012}).

Among the investigated measures, the mutual information diversity metric provides a general approach to compute a non-linear, non-monotonic relationship between behaviors, that is applicable to numerical and symbolic behavioral representations, both in the continuous and discrete domains \cite{Thomas1991,kraskov2004}. For our signature of evolvability, we compute the behavioral diversity as the normalized mutual information between behaviors of an individual, before and after its genome is mutated.

Assuming that the behavior of an individual $i$ can be represented as a discrete vector $B_i$ (details in \cite{mouret2012}), for the behaviors $B_i$ and $B_i^\prime$, of individual $i$ and mutant~$i^\prime$, we have:

\small
\begin{subequations}
\begin{align}
H(B_i) &= -\sum_{b_i \in B_i} p(b_i) \log p(b_i) \label{eqn:evob_f2a}\\
H(B_i, B_i^\prime) &= -\sum_{b_i \in B_i} \sum_{b_i^\prime \in B_i^\prime} p(b_i,b_i^\prime) \log p(b_i,b_i^\prime) \label{eqn:evob_f2b}\\
f_2 &= 1 - \frac{H(B_i) + H(B_i^\prime) - H(B_i,B_i^\prime)}{max(H(B_i), H(B_i^\prime)} \label{eqn:evob_f2c}
\end{align}
\end{subequations}
\normalsize

\noindent where $H(B_i)$ is the entropy of the behavior $B_i$ comprising the individual states $b_i$ with probability $p(b_i)$, $H(B_i,B_i^\prime)$ is the joint entropy between behaviors $B_i$ and $B_i^\prime$ with joint probability density function $p(b_i,b_i^\prime)$, and $f_2$ denotes the inverse of the normalized mutual information between the two behaviors.

The entropy and joint entropy are computed by first approximating $p(b_i)$ and $p(b_i,b_i^\prime)$, by counting the number of instances of each behavior state. Systematic errors in the probability estimates, consequent to the limited number of available data samples, is compensated for by adding a corrective term $E$ to the computed entropy: $E=\left(S_i-1\right)/2T$ (where $T$ is the size of the temporal window over which the entropy is computed, and $S_i$ is the number of states for which $p(b_i) \neq 0$), and $E =\left(S_i + S_{i^\prime} - S_{i,i^\prime} - 1\right)/2T$ to the joint entropy (where $S_{i}$, $S_{i^\prime}$, $S_{i,i^\prime}$, and $T$ have an analogous meaning to the previous case). These corrective terms compensate for the systematic and random errors in the observed entropy of a series, that systematically bias downwards the expected value of the observed entropy from the true entropy (derived in \cite{roulston1999estimating}). Integrating the corrective term to the equations for entropy and joint entropy, we have:

\small
\begin{subequations}
\begin{align}
H(B_i) &= -\sum_{b_i \in B_i} p(b_i) \log p(b_i) + \frac{S_i-1}{2T} \label{eqn:evob_f2a_corrc}\\
H(B_i, B_i^\prime) &= -\sum_{b_i \in B_i} \sum_{b_i^\prime \in B_i^\prime} p(b_i,b_i^\prime) \log p(b_i,b_i^\prime) + \frac{S_i + S_{i^\prime} - S_{i,i^\prime} - 1}{2T} \label{eqn:evob_f2b_corrc}
\end{align}
\end{subequations}
\normalsize

Estimates of the corrected entropy (eq.~\ref{eqn:evob_f2a_corrc}) and joint entropy (eq.~\ref{eqn:evob_f2b_corrc}) are then used to update the mutual information distance between behaviors. The resulting feature $f_2$ represents the quantity of behavioral variation following genetic change, and is indicative of the ability of the evolutionary system to produce novel behaviors.

\section{Hexapod robot locomotion problem}
\label{sec::hexapodlocomotion}

The evolution of locomotion gaits for multilegged robots is a classical problem in evolutionary robotics, addressed in many studies utilizing both direct and generative encodings, on bipedal~(e.g.,~\cite{liu2004hierarchical}), quadrupedal (e.g., \cite{clune2011,hornby2005autonomous,Risi2013,tellez2006evolving,valsalam2008modular}), and hexapedal robots~(e.g.,~\cite{valsalam2008modular,zykov2004evolving,barfoot2006experiments}) -- employed here for the comparison of different encodings. In most existing studies on evolved locomotion gaits, the performance of an individual is analyzed solely by its walking speed and the required number of generations of evolution. The rate of evolution and evolved performance has also been linked to evolvability provided by the encoding scheme, wherein controllers achieving a higher task fitness and requiring fewer generations to evolve are considered more evolvable~(e.g.,~see~\cite{hornby2003generative,clune2009evolving,gruau1994automatic,komosinski2001comparison}). While these approaches provide interesting insights on the performance of the Genotype-to-Phenotype mapping, they largely ignore its capabilities to generate viable phenotypic variations (diverse gaits in case of legged robots). However, the diversity of evolved walking gaits is important for the legged robot to recover rapidly from faults such as, the loss of one or more limbs, or motor malfunctions~\citep{koos2013fast}, and for the robot to adapt to previously unencountered environmental changes. Furthermore, an efficient recovery is particular relevant for hexapedal legged robots, wherein the probability of component failure is high, consequent to the large number of moving parts.

\begin{figure}
\centering
\subfloat[Hexapod robot]{\includegraphics[width=4.205cm, height=4.16cm]{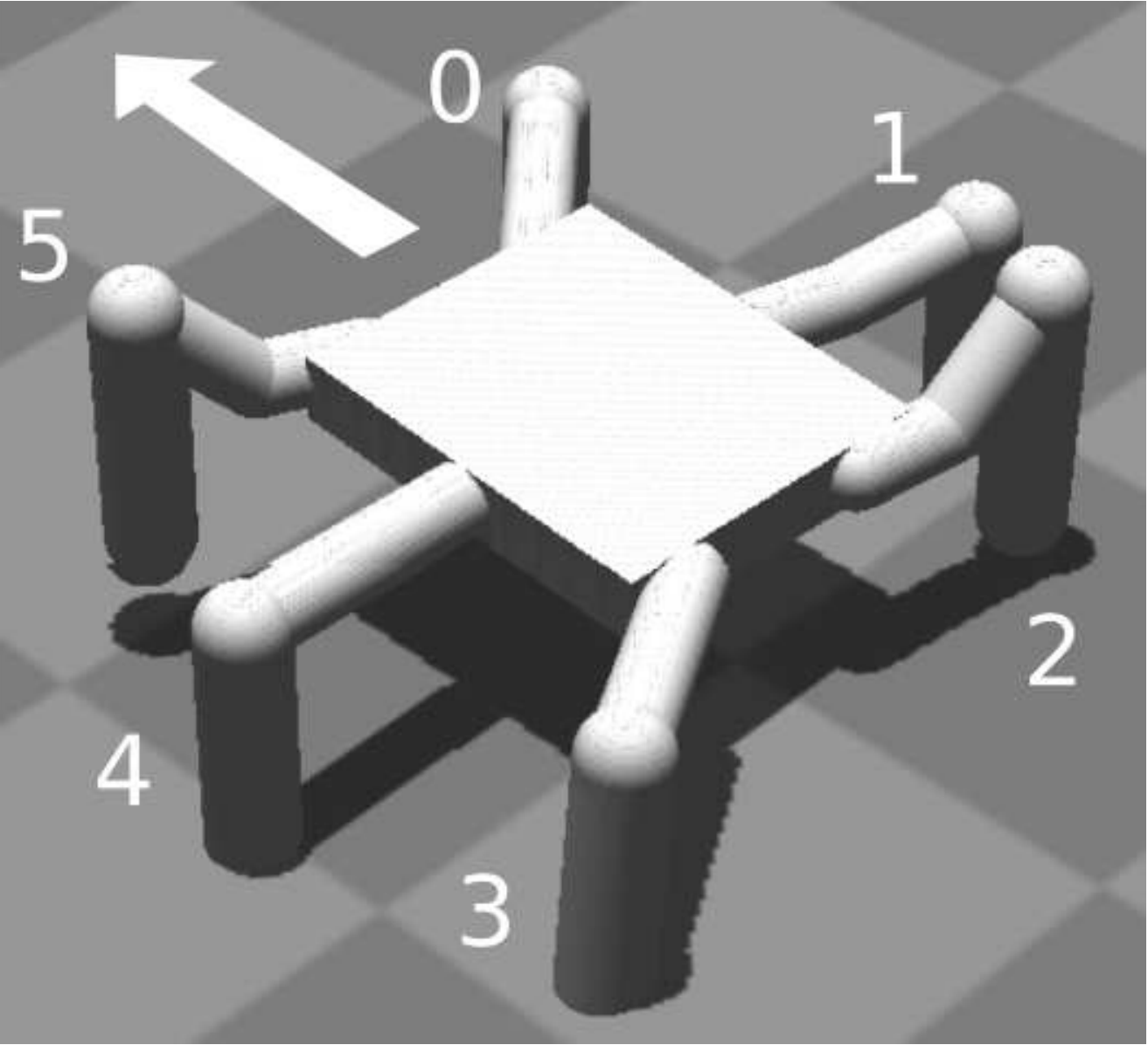}}\hfil
\subfloat[Kinematic scheme]{\includegraphics[width=4.205cm, height=4cm]{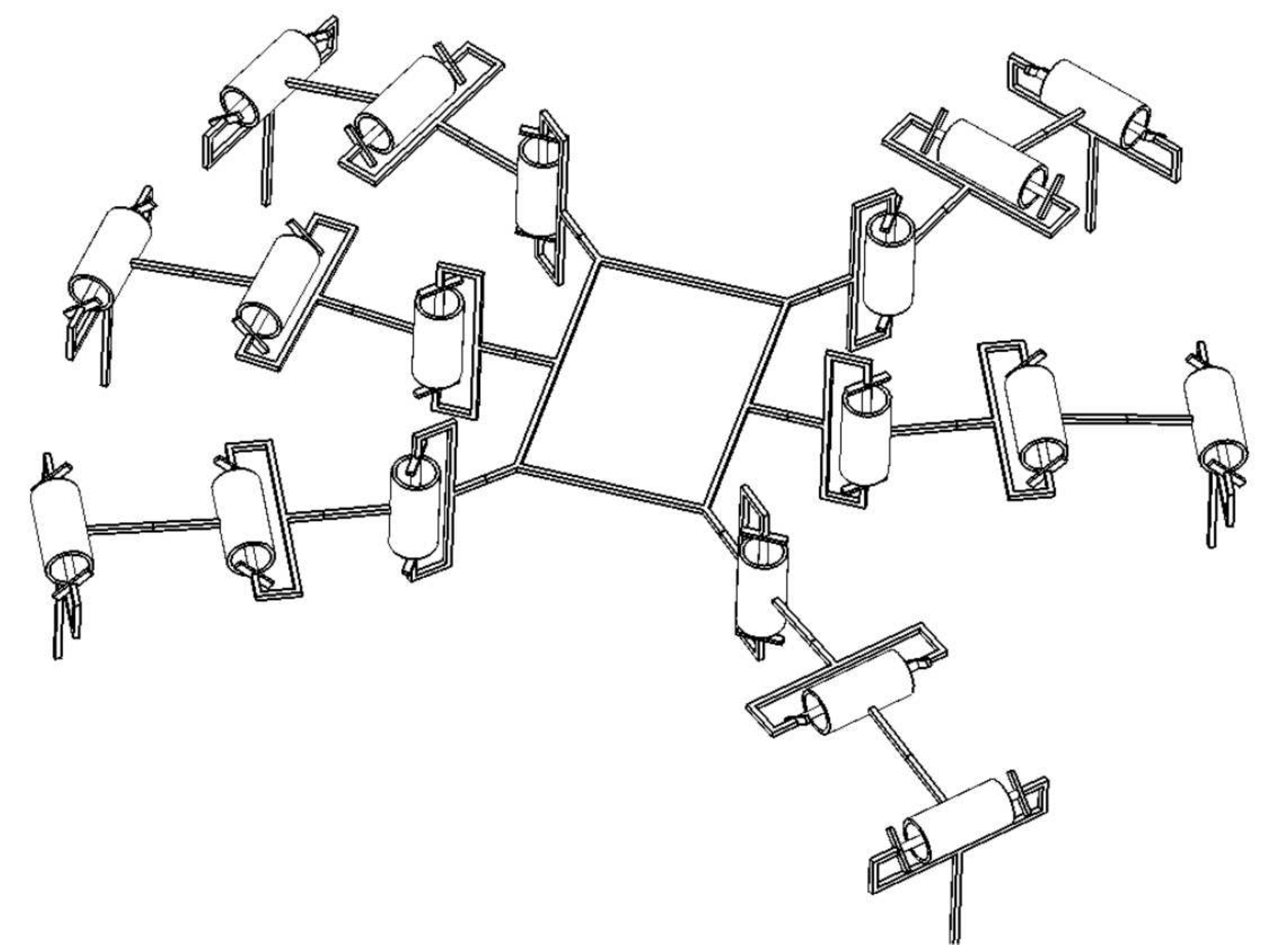}}
\caption{(a) Snapshot of an 18-DOF simulated hexapod robot walking on a horizontal surface, with contacts simulated. (b) Kinematic scheme of the robot, with cylinders representing actuated pivot joints. The three servos on each leg, $\mathtt{s_1}$, $\mathtt{s_2}$ and $\mathtt{s_3}$, are labeled in increasing order of distance to robot torso.}
\label{fig:robot}
\end{figure}

\textbf{Hexapod platform details:}~The hexapod robot is simulated on a flat, horizontal surface (Fig.~\ref{fig:robot}a), with the Open Dynamics Engine\footnote{\url{http://www.ode.org}} (ODE) physics simulator. The robot has $18$ Degrees of Freedom (DOF), $3$ for each leg (Fig.~\ref{fig:robot}b), and each DOF is actuated by a single servo. The first servo on each leg ($\mathtt{s_1}$) actuates the horizontal orientation of the leg within range $[-\pi/8, \pi/8]$~radians. The second ($\mathtt{s_2}$) and third ($\mathtt{s_3}$) servos control the leg elevation and extension, respectively, each within the range of $[-\pi/4, \pi/4]$~radians.

Angular positions are sent to the $18$ servos once every $15$~ms. Furthermore, in order to maintain the last subsegment of each leg vertical (for enhanced stability), the control signal for the third servo ($\mathtt{s_3}$) is always in antiphase to that of the second servo ($\mathtt{s_2}$). Consequently, the robot is reduced to a $12$ DOF system, despite being actuated by $18$ motors.

\textbf{Hexapod gait representation:}~The behavioral diversity in our signature of evolvability corresponds to the inter-gait diversity in the hexapod robot locomotion problem. For this diversity, a hexapod gait is represented using a gait diagram~\citep[p.~379]{siciliano2008springer}, comprising a binary matrix $C$ of leg-surface contacts:

\small
\begin{equation*}
  C_{tl}=\begin{cases}
    1 & \text{if leg $i$ makes surface contact at time-step $t$,}\\
    0 & \text{otherwise.}
  \end{cases}
\end{equation*}
\normalsize

\noindent where $t \in \{0 \dotsc T\}$, the gait is evaluated for $T$ time-steps, and the hexapod legs $l \in \{0 \dotsc 5\}$.

The hexapod gait for an individual $i$ is represented by binary vector $B_i$, comprising the contacts in $C$ concatenated in row-major order, $B_i = [C_{00}, C_{10} \dotsc C_{T5}]$. Diversity between two gaits is measured as the normalized mutual information between the corresponding gait vectors~(eq.~\ref{eqn:evob_f2c}).

\section{Encoding schemes analyzed}
\label{sec:encanal}

Generative encodings for evolving our hexapod locomotion controllers are based on CPPNs~\citep{stanley2007compositional}. The CPPN abstracts the processes of embryonic development by determining the attributes of phenotypic components as a function of their geometric location in the individual, instead of simulating complex inter-cellular interactions and chemical morphogen gradients to determine component location~\citep{carroll2005endless}. In nature, cells differentiate themselves into different lineages, influenced by their immediate environment~(the epigenetic landscape, \cite{goldberg2007epigenetics}). Analogously, the CPPN genome outputs the fate of an organismal component as a function of its geometric coordinates in the individual.

The CPPN genome is represented as a directed graph, comprising a set of \textit{Sine}, \textit{Gaussian}, \textit{Sigmoid}, and \textit{Linear} type of nodes, connected by weighted links. The node type indicates the activation function applied to the sum of its weighted inputs, to compute the node output. Selected activation functions can succinctly encode a wide variety of phenotypic patterns, such as symmetry (e.g., a Gaussian function) and repetition (e.g., a Sine function), that evolution can exploit. Mutations to the CPPN genome can change the connection weights and node type, and add or remove nodes from the graph. Consequently, the topology of the CPPN is unconstrained, open-ended, and can represent any possible relationship between the input coordinates of the phenotypic component and its output attributes (see details in~\cite{stanley2007compositional}).

\begin{table*}
\caption{Summary of encoding schemes to evolve controllers for hexapod robot locomotion.}\label{tab:encodings}
\centering
\begin{scriptsize}
\setlength{\extrarowheight}{3pt}
\begin{tabular}{||l|l|c|m{6.5cm}|p{5cm}||} 
\hline
\textbf{Encoding} & \textbf{Signal generator} & \textbf{Feedback} & \textbf{Summary} & \textbf{References} \\
\hline
Open-loop CPG & Phase oscillator & \xmark & An amplitude-controlled phase oscillator is used to generate locomotion gaits. The $12$ amplitude parameters, and $11$ inter-oscillator phase bias parameters of the oscillator, are generatively encoded with a CPPN. &\cite{ijspeert2007swimming,righetti2006design,crespi2013salamandra,ijspeert2008central}\\
\hline
Closed-loop CPG & Phase oscillator & \cmark & A generatively encoded amplitude-controlled phase oscillator is extended with the inclusion of a phase resetting mechanism. The extension introduces a sensory feedback mechanism that modulates the oscillations produced by the CPG, adapting the oscillation period depending on the locomotion gait and terrain.  &\cite{aoi2005locomotion,fukuoka2003adaptive,righetti2008pattern}\\
\hline
HyperNEAT & ANN & \xmark & A widely used generative encoding scheme, used to evolve large-scale ANNs. The weights of the neural network are encoded with a CPPN. Evolved ANNs have been successfully deployed to generated symmetric and coordinated gaits for both simulated and physical quadruped robots. & \cite{stanley2009hypercube,clune2011,yosinski2011,lee2013}\\
\hline
SUPG & CPPN & \cmark & A recently developed encoding scheme, wherein a CPPN encodes the attributes of a SUPG. The SUPG is a macro-neuron, that upon receiving an external trigger, produces a single cycle of an oscillatory signal. Repeated triggering of the SUPG generates a periodic gait for a multilegged robot. In previous work, the SUPG outperformed HyperNEAT encodings in evolving locomotion gaits for a simulated quadruped robot, thus encouraging further study. & \cite{morse2013}\\
\hline
Direct & Phase oscillator & \xmark & A simple locomotion controller, used as reference for comparison between encoding schemes. The robot is controlled with a amplitude controlled phase oscillator, the $12$ amplitude parameters, and $11$ inter-oscillator phase bias parameters, are directly encoded. & \cite{koos2013fast,cully2013}\\
\hline
\end{tabular}
\end{scriptsize}
\end{table*}

In this study, the CPPN genotype is mapped to four very different phenotypes to control hexapod locomotion, open-loop CPGs, closed-loop CPGs, ANNs (minimal HyperNEAT), and SUPGs (summarized in Table~\ref{tab:encodings}). The SUPG is a new type of macro-neuron introduced by~\citeauthor{morse2013}~\cite{morse2013} to genetically encode spatio-temporal oscillatory patterns.

\subsection{Encoding CPGs with CPPNs}
A CPG is a biological neural network, comprising a distributed array of neurons located along the spinal cord of vertebrates~\citep{frigon2006experiments}. The spinal neural centers exhibit rhythmic activity, generating complex high-dimensional signals for the control of coordinated periodic movements required for animal locomotion. In recent years, bio-inspired engineering approaches have led researchers to model these CPGs as coupled dynamical oscillatory systems~\citep{ijspeert2008central}, to generate locomotion control policies for biped~\citep{aoi2005locomotion,taga1994emergence}, quadruped \citep{righetti2008pattern,righetti2006design,fukuoka2003adaptive} and hexapod robots \citep{ren2014multiple}, as well as modular~\citep{sproewitz2008learning} and swimming/walking salamander-like robots~\citep{ijspeert2007swimming,crespi2013salamandra}. 

The design of CPG-based control policies for robot locomotion offers many advantages~\citep{ijspeert2008central}. The main advantage of using such controllers is the stable limit cycle behaviors of the coupled dynamical system. Additionally, the CPGs maintain a smooth transient response to changes to external inputs (e.g., from sensory feedback). Another advantage of CPGs is the intrinsic properties of inter-oscillator synchronization, allowing for coordinated interactions between the robot and its environment. Finally, using CPGs for generating coordinated policies reduces the search space of the control problem, as opposed to using other approaches such as continuous-time recurrent neural networks~\citep{beer1992evolving}. 

In most studies investigating CPGs for robot locomotion, the parameters of the model are hand-tuned by the designer (e.g.,~\cite{aoi2005locomotion,righetti2008pattern,crespi2013salamandra}). While such an approach does provide interesting coordinated robot movements, the system is unable to autonomously explore a rich diversity of different locomotion gaits. Furthermore, the parameters of the model are required to be re-tuned in response to changes in the environment, or to damages incurred by the robot. To overcome these issues, we seek to evolve parameters of generatively encoded CPGs, thus combining the advantages of CPG models with that of generative encodings. The following two CPPN-encoded CPG models are investigated: (i)~an open-loop model wherein the control policies are not influenced by any sensory feedback; (ii)~a closed-loop model, with sensory feedback from touch sensors on each of six hexapod legs influencing the resultant gait.

\begin{figure}[h]
\centering
\def\tabularxcolumn#1{m{#1}}
\begin{tabularx}{9.5cm}{@{}rX@{}}
\begin{tabular}{c}
\subfloat[CPPN encoding CPG parameters.]{\includegraphics[width=3.5cm, height=4cm]{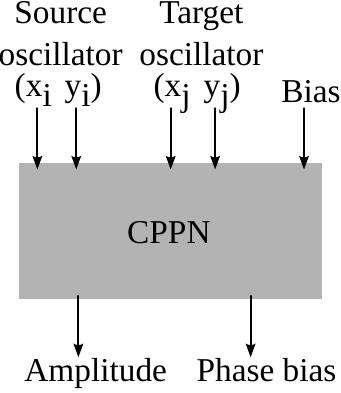}}
\end{tabular}
&
\subfloat[CPG substrate.]{\includegraphics[width=4.5cm, height=4cm]{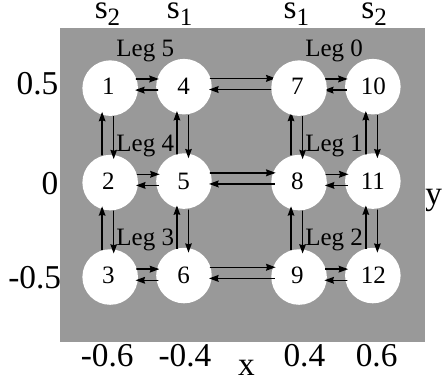}}
\end{tabularx}
\caption{Encoding CPGs with CPPNs. The intrinsic amplitude $\Alpha_{i}$ of each oscillator is encoded by the CPPN as a function of its position in the substrate. Phase biases $\phi_{i,j}$ of inter-oscillator couplings are determined by querying the CPPN with the coordinates of every pair of adjacent source $\left(x_i,y_i\right)$ and target $\left(x_j,y_j\right)$ oscillators. For every pair of adjacent oscillators, the query is made only once as $\phi_{i,j} = -\phi_{j,i}$.}
\label{fig:cpg}
\end{figure}

\subsubsection{Open-loop CPG}
\label{sec:cpg}
The open-loop generatively encoded CPG system is comprised of $12$ coupled amplitude-controlled phase oscillators~\citep{ijspeert2007swimming}, governing the actuation of the $12$ servos ($\mathtt{s_1}$ and $\mathtt{s_2}$, on each of $6$ robot legs). Each oscillator is modeled by the following set of ordinary differential equations:

\begin{subequations}
\begin{align}
{\dot{\theta_{i}}} &= 2\pi \Frequency_{i} + \sum\limits_{j} \alpha_{j} w_{i,j} \sin \left(\theta_{j} - \theta_{i} - \phi_{i,j} \right) \label{eqn:cpg_a}\\
{\ddot{\alpha_{i}}} &= \beta \left(\frac{\beta}{4} \left(\Alpha_{i} - \alpha_{i} \right) - \dot{\alpha_{i}} \right) \label{eqn:cpg_b}\\
{\gamma_{i}} &= \alpha_{i} \cos \left(\theta_{i} \right) \label{eqn:cpg_c}
\end{align}
\end{subequations}

\noindent where $\alpha_{i}$ and $\theta_{i}$ denote the amplitude and phase of the oscillator $i$, $\Alpha_{i}$ represents its intrinsic amplitude (in radians), and $\Frequency_{i}$ the intrinsic frequency (in Hz). The oscillator amplitude $\alpha_{i}$ converges to the intrinsic amplitude $\Alpha_{i}$ at a rate determined by positive constant $\beta$ (set to $10$~rad/s for rapid convergence). Couplings between oscillators are defined by weights $w_{i,j}$, and phase biases $\phi_{i,j}$. The coupling weights define the strength of the coupling, and influence the time to synchronize between oscillators (all weights prefixed at $20$ to enable rapid inter-oscillator synchronization). The phase bias $\phi_{i,j}$ determines the phase difference between the oscillators $i$ and $j$. The output of the oscillator is denoted by $\gamma_{i}$, and is computed using the Euler integration method with a step-size of $20$~ms.

Oscillator equations (eq.~\ref{eqn:cpg_a},b and c) were designed such that the output of the oscillator $\gamma_{i}$ exhibits a limit cycle behavior, producing a stable periodic output. The first equation (eq.~\ref{eqn:cpg_a}) determines the time evolution of the phase $\theta_{i}$ of the oscillators. In this study, all $12$ oscillator of the CPG have the same frequency ($\Frequency_{i}=1$~Hz), and the bidirectional coupling between oscillators is such that $\phi_{i,j} = -\phi_{j,i}$. Furthermore, we ensure that the sum of all the phase biases in every closed loop of inter-oscillation couplings is a multiple of $2\pi$, so that all the phase biases in a loop are consistent. In such a parameter regime, the oscillator phases grow linearly at a common rate $\Frequency_{i}$ and with an inter-oscillator phase difference of $\phi_{i,j}$. The second equation (eq.~\ref{eqn:cpg_b}) is a critically damped second order linear differential equation, with $\Alpha_{i}$ as the stable fixed point. Therefore, from any initial condition the oscillator amplitude $\alpha_{i}$ will asymptotically and monotonically converge to the intrinsic amplitude $\Alpha_{i}$, allowing the amplitude to be smoothly modulated. Finally, utilizing the amplitude and the phase of the oscillator (eq.~\ref{eqn:cpg_c}), the output of the oscillator $\gamma_{i}$ governs servo actuation.

In this first generative encoding evaluated, a CPPN encodes the intrinsic amplitudes $\Alpha_{i}$ and inter-oscillator phase biases $\phi_{i,j}$ of $12$ oscillators of the CPG~(Fig.~\ref{fig:cpg}a). The oscillators are placed in a $2$-D Cartesian grid termed the \textit{substrate}, so that each oscillator has a distinct $\left(x,y\right)$ coordinate, and so as to reflect the  hexapod robot morphology~(Fig.~\ref{fig:cpg}b). The intrinsic amplitude of each oscillator $i$ is obtained by inputting to the CPPN the coordinates $\left(x_i,y_i\right)$, and setting the inputs $\left(x_j,y_j\right)$ to $0$. Amplitudes output are scaled to the allowable angular range of the corresponding motors. In the CPG, adjacent oscillators are coupled together~(Fig.~\ref{fig:cpg}b). The phase bias for every pair of adjacent oscillators $i$ and $j$ is obtained by querying the CPPN with inputs $\left(x_i,y_i\right)$ and $\left(x_j,y_j\right)$, and scaling the output to range $\left[0, 2\pi\right]$. Furthermore, the following two constraints are introduced: (i) couplings are bilaterally symmetrical, i.e., $\phi_{i,j} = -\phi_{j,i}$. For every pair of adjacent oscillators, the phase bias is queried only once; (ii) phase biases $\phi_{2,1}$, $\phi_{2,3}$, $\phi_{7,4}$, $\phi_{9,6}$, $\phi_{10,11}$ and $\phi_{12,11}$ are not queried, but computed such that the sum of phase biases in every closed loop of the CPG is a multiple of $2\pi$ (oscillators numbered in Fig.~\ref{fig:cpg}b). Therefore, the total number of CPG parameters generatively encoded by the CPPN is $23$ ($12$ intrinsic amplitude and $11$ phase bias parameters).

\begin{figure*}[h]
\centering
\subfloat[Normal step cycle]{\includegraphics[width=5.1cm, height=3.4cm]{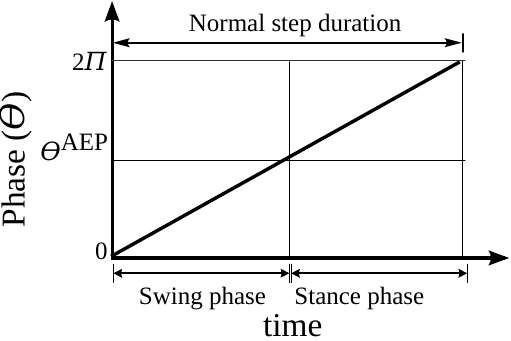}}
\subfloat[Extended step cycle]{\includegraphics[width=5.7cm, height=3.4cm]{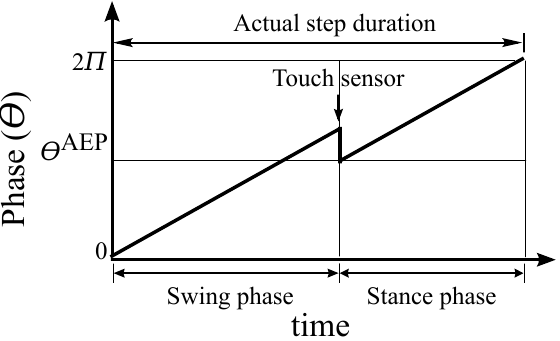}}
\subfloat[Attenuated step cycle]{\includegraphics[width=4.8cm, height=3.4cm]{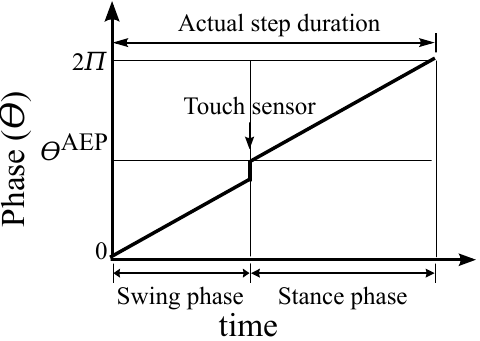}}
\caption{An illustration of the normal and actual phase $\theta_{i}$ of oscillator $i$ over time. The phase of an oscillator governing leg horizontal orientation (servo $\mathtt{s_1}$) is reset to $\theta^{AEP}$ following a signal from the touch sensor, resulting in changes to the step cycle.}
\label{fig:cpg_phasereset}
\end{figure*}

\subsubsection{Closed-loop CPG}
The second generative encoding scheme evaluated is an extension of the open-loop scheme (see Sec.~\ref{sec:cpg}). While the generatively encoded CPG parameters and the CPPN encoding remains the same as in the open-loop model (Fig.~\ref{fig:cpg}), the modification introduced is a sensory feedback mechanism that modulates the oscillations produced by the CPG. In this closed-loop encoding scheme, feedback signals from the touch sensors attached to each of the six legs of the hexapod trigger a phase-resetting mechanism \cite{aoi2005locomotion}, that adapts the oscillation period depending on the gait and the terrain. For the phase-resetting mechanism, two extreme positions of the horizontal orientation of the robot leg are introduced with respect to the robot trunk, (i)~the anterior extreme position (AEP), where the swing phase transitions to the stance phase, and (ii)~the posterior extreme position (PEP), where the stance phase transitions to the swing phase. 

The phase of the oscillator at the AEP, $\theta^{AEP}$ is determined by the following equations. 

\begin{align*}
\beta &= \frac{T_{ST}}{T_{ST} + T_{SW}}\\
\theta^{AEP} &= 2\pi\left(1 - \beta\right) 
\end{align*}

\noindent where $T_{SW}$ is the duration of the swing phase, $T_{ST}$ is the duration of the stance phase, and $\beta$ ($0<\beta<1$) is the duty ratio, i.e., the ratio between the stance phase duration and the total step cycle~\citep{siciliano2008springer}. In this study, $\beta$ is prefixed at $0.5$, resulting in $\theta^{AEP}$ to be at $\pi$.

Utilizing $\theta^{AEP}$ for the oscillators governing horizontal orientation on each of six legs of the robot (oscillators $i \in  \{4, 5 \dotsc 9\}$, see Fig.~\ref{fig:cpg}), the phase equation (eq.~\ref{eqn:cpg_a}) is modified while keeping the equations for the amplitude (eq.~\ref{eqn:cpg_b}) and the oscillator output (eq.~\ref{eqn:cpg_c}) unchanged.

\begin{subequations}
\begin{align}
{\dot{\theta_{i}}} &= 2\pi \Frequency_{i} + \sum\limits_{j} \alpha_{j} w_{i,j} \sin\left(\theta_{j} - \theta_{i} - \phi_{i,j}\right) + g_{i} \label{eqn:cpgfb_a}\\
{g_{i}} &= \left(\theta^{AEP} - \theta_{i}\pmod{2\pi} \right) \delta\left(t_{i}^{land} - t\right)\label{eqn:cpgfb_b}
\end{align}
\end{subequations}
\noindent where $t_{i}^{land}$ is the time when the foot of the leg being horizontally oriented by oscillator $i$ lands on the ground, and $\delta\left(\cdot\right)$ is the Dirac delta function ($[\cdot] : R \to [0,\infty)$) defined as $0$ for all arguments except at the origin, where it is $1$. The function $g_{i}$ resets the phase $\theta_{i}$ of oscillator $i$ to the normal value $\theta^{AEP}$ instantaneously when the foot of the leg lands on the ground.

When the foot touches the ground, the phase of the corresponding oscillator is reset to $\theta^{AEP}$, the normal phase value where the transition from the swing phase to the stance phase is supposed to occur. Consequently, in the oscillator's step cycle the actual duration of the swing phase is not fixed, but is dependent on the timing of the foot touching the ground (see Fig.~\ref{fig:cpg_phasereset}), and the resulting oscillator's step cycle is thus modulated by sensory feedback.

\subsection{Encoding ANNs with CPPNs (minimal HyperNEAT)}
The third generative encoding scheme evaluated is a simplified version of HyperNEAT indirect encoding\footnote{The CPPN is evolved with a simple multiobjective evolutionary algorithm, instead of the NEAT method (details in \cite{tonelli2013relationships}).} In previous work, the CPPN has been used successfully to evolve modular and regular patterns in the connection space of the ANN, resulting in symmetric and coordinated gaits for both simulated and physical quadruped robots~\citep{stanley2009hypercube,clune2011,yosinski2011,lee2013}. The results encourage us to include the HyperNEAT encoding in our comparative study.

\begin{figure*}
\begin{center}
\includegraphics[width=16cm]{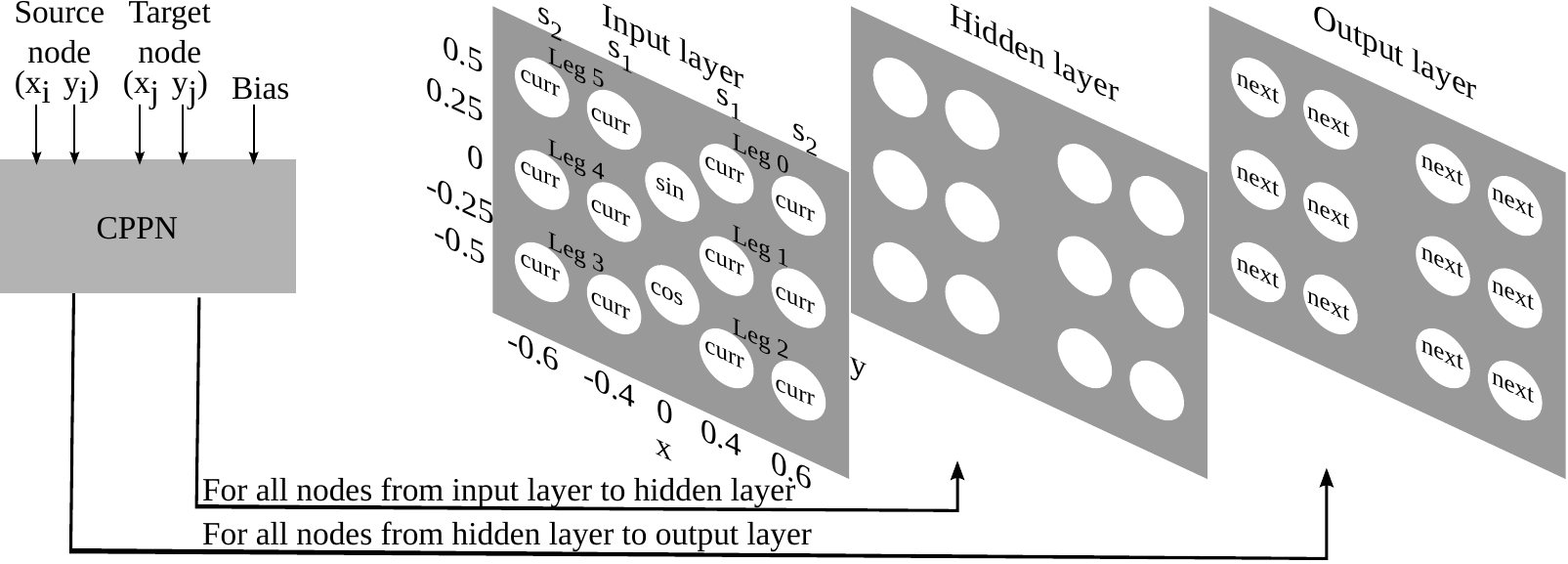}
\end{center} 
\caption{Encoding ANNs with CPPNs (inspired by~\cite{clune2011,yosinski2011,lee2013}). Inter-neuron connection weights are encoded as function of the position of source and target neurons of each neural connection. The CPPN outputs the weights of input-hidden and hidden-output neuron connections, for each source $\left(x_i,y_i\right)$ and target $\left(x_j,y_j\right)$ neuron in proximal layers. The ANN is input the requested angles of the previous time-step for the first two servos ($\mathtt{s_1}$ and $\mathtt{s_2}$) on each leg, and a sine and cosine wave. The output neurons specify the new joint angles for the current time-step.}
\label{fig:ann_substrate}
\end{figure*}

The CPPNs encode the weights of a fixed topology, single-layer feedforward ANN, featuring $2$-D Cartesian grids of inputs, hidden and output neurons~(Fig.~\ref{fig:ann_substrate}). Neurons of the ANN are positioned in the substrate, in accordance with the hexapod robot morphology. Using the encoding, the CPPN is iteratively queried the positions of all source $\left(x_1,y_1\right)$ and target $\left(x_2,y_2\right)$ neurons in proximal layers, along with a constant bias, and it outputs the corresponding weights of the input-hidden and hidden-output neuron connections.

The ANN receives as input the previously requested angles (actual angles unknown) for each of the $12$ pivot joints of the hexapod robot ($\mathtt{s_1}$ and $\mathtt{s_2}$, for $6$ legs). In addition, sine and cosine waves of frequency $1$~Hz are also input to the ANN, to facilitate periodic oscillations at the output neurons. The output from the ANN at each time-step are $12$ numbers (one for each of $\mathtt{s_1}$ and $\mathtt{s_2}$, on each of $6$ legs) in interval $\left[-1, 1\right]$, that are scaled to the allowable angular range of the corresponding motors, and indicate the next position of each motor.

In preliminary experiments, the HyperNEAT encoding evolved ANNs that exhibited high frequency output oscillations (in excess of $20$~Hz).~In the resultant gaits, the robot could traverse large distances by vibrating its legs rapidly, and in unison. Such high frequency pronking gaits would clearly not be replicable on the physical robots, and result in overly taxed servos~\cite{yosinski2011}. To resolve this problem, and as suggested by \cite{lee2013}, we generated joint angular positions with a time-step of $15$~ms, by averaging over four consecutive pseudo-positions generated at $3.75$~ms intervals. The number of evolved ANN controllers outputting high frequency oscillation was thus reduced, with oscillations at $2.44\pm1.95$~Hz (Median$\pm$IQR in $20$ replicates, with the frequency at each replicate averaged across $18$ servos).

\subsection{Encoding SUPGs with CPPNs}

In the fourth generative encoding scheme evaluated, the CPPN encodes the attributes of a SUPG. The SUPG is a macro-neuron that upon receiving an external trigger, produces a single cycle of a CPPN encoded oscillatory signal. Consequently, the repeated triggering of the SUPG generates a periodic pattern, used to govern the actuation of a multilegged robot. In a previous work, the SUPG outperformed HyperNEAT encodings in evolving locomotion gaits for a simulated quadruped robot~\citep{morse2013}. The resultant SUPG gaits appeared faster and steadier in extended evaluations, encouraging us to study the encoding both in terms of performance, and the evolvability provided.

\begin{figure}
\centering
\def\tabularxcolumn#1{m{#1}}
\begin{tabularx}{9.5cm}{@{}rX@{}}
\begin{tabular}{c}
\subfloat[A single SUPG.]{\includegraphics[width=3cm, height=2.14cm]{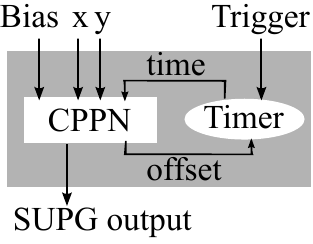}}\\
\subfloat[Timer of SUPG.]{\includegraphics[width=3cm, height=1.36cm]{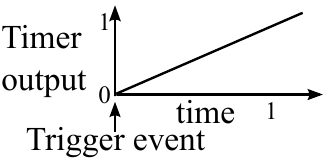}}\\
\end{tabular}
&
\subfloat[SUPG substrate.]{\includegraphics[width=4.5cm, height=4.1cm]{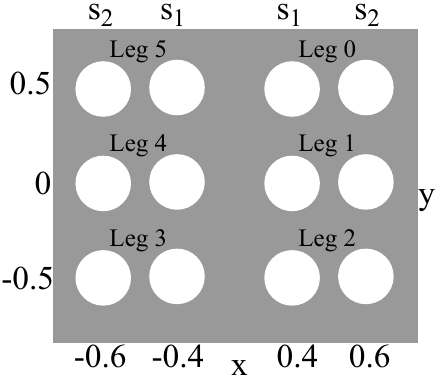}}
\end{tabularx}
\caption{Encoding SUPGs with CPPNs (inspired by~\cite{morse2013}). (a) The SUPG output is a function of its coordinates $\left(x,y\right)$ in the substrate, and the elapsed time since last trigger (output of Timer). The time of first trigger is determined by an offset. (b) Once triggered, the SUPG timer ramps upward linearly from $0$ to $1$ and stays there, until it is re-triggered. (c) Positions of the $12$ SUPGs in the substrate, outputting the desired angles for the first two servos ($\mathtt{s_1}$ and $\mathtt{s_2}$), on each leg of the hexapod.}
\label{fig:supg}
\end{figure}

In this encoding, the CPPN is input the position $(x,y)$ of the SUPG in the substrate, and the elapsed time since the SUPG was last triggered (Fig.~\ref{fig:supg}a). The elapsed time is recorded by an internal timer, individual to each SUPG, and ramps upwards from an initial value of $0$ to a maximum value of $1$, in one period of the SUPG's output signal (Fig.~\ref{fig:supg}b).~Therefore, the SUPG's output is a function of both, its position in the substrate, and the time since its last cycle was triggered. Applying the SUPGs for hexapod locomotion, the substrate comprises $12$ SUPGs positioned to reflect the robot morphology (Fig.~\ref{fig:supg}c). The outputs of the SUPGs at each time-step specify the desired angles for the first and second servos ($\mathtt{s_1}$ and $\mathtt{s_2}$), on each leg of the robot. 

The internal timer of the SUPG can be restarted from $0$, following the occurrence of an external trigger event (Fig.~\ref{fig:supg}b). Consequently, the SUPG cycle does not need to match the length of an optimal robot-walking step. Rather, the oscillation period can be adjusted depending on the gait and the terrain by restarting the SUPG whenever its associated foot touches the ground, thus producing a closed-loop control. In our hexapod robot, the two SUPGs actuating each leg (Fig.~\ref{fig:supg}c), are simultaneously triggered by the corresponding foot touching the ground. 

At the start of the simulation, all six legs of the robot are in contact with the ground, resulting in all the SUPGs being triggered simultaneously. To avoid the resulting hopping gaits, the first trigger to each SUPG is delayed by an offset. The offset output of the CPPN is determined for the $\mathtt{s_1}$ SUPG on each leg by supplying its coordinates as input, and setting the time input to $0$. The same offset value is also applied to the $\mathtt{s_2}$ SUPG on the leg, allowing both the oscillators on each leg to start simultaneously.

\subsection{Direct encoding}
Locomotion controllers evolved with direct encoding are designed to be simple, wherein the actuation along each DOF of the robot is governed by the periodic signal of an open-loop amplitude controlled phase oscillator. With this encoding, hexapod leg actuation is governed by the differential equation model (eq.~\ref{eqn:cpg_a}, b and c). There are $12$ evolved parameters for the intrinsic amplitudes of oscillators, governing the actuation of the two servos $\mathtt{s_1}$ and $\mathtt{s_2}$, on each of six legs of the hexapod. In addition, $11$ inter-oscillator phase bias parameters are also evolved (see Fig.~\ref{fig:cpg}b, and Sec.~\ref{sec:cpg} for details on constraints on phase biases). Consequently, a directly encoded controller for the hexapod robot is fully represented by $23$ parameters.

\section{Experiments}
\label{sec:results}

We conducted $8,000$ generations of artificial selection in populations consisting of $100$ individuals ($8,000 \times 100$ evaluations).~Our aim was to evolve controllers for the hexapod robot to walk forward, evaluated for a period of $5$~s ($334$ time-steps). The Nondominated sorting genetic algorithm II~\citep{deb2002fast} was used to simultaneously optimize the following three objectives:

\begin{equation}
  \text{Maximize} \begin{cases}
    -F_i\\
    -\lvert\Theta_i\rvert\\
    \frac{1}{N} \sum_{j=0}^{j=N} D(B_i,B_j)
  \end{cases}
  \label{eqn:mo}
\end{equation}

\noindent where for individual $i$ in the population, $F_i$ is the fitness computed as the distance between the final position of $i$ and a goal located $25$~m directly in front of the robot's initial position, $\Theta_i$ denotes the angle of the robot's trajectory with respect to the normal forward walking direction, $D(B_i,B_j)$ is the hamming distance between the binary gait vectors of individual $i$ and $j$, and $N$ is the size of the population.

In eq.~\ref{eqn:mo}, the first and second objectives reward individuals to walk forward large distances towards a goal, unattainable by the robot within the experiment evaluation time. The third objective is introduced to facilitate the exploration of diverse solutions and avoid premature convergence to suboptimal solutions at local minima \cite{mouret2012}.

Artificial selection was conducted in $20$ independent replicates, for the \textbf{Direct} encoding, and the four generative encodings, (i)~\textbf{CPG} (open-loop controller), (ii)~\textbf{CPG-f/b} (closed-loop controller), (iii)~\textbf{ANN} (minimal HyperNEAT), and (iv)~\textbf{SUPG}\footnote{A single evolution replicate required about $24$h of computational time on a 8-cores Intel Xeon E5520 at $2.27$~Ghz.}. Performance and evolvability analysis are reported for the best individual of each replicate, selected to have the highest fitness in the population, and with an angle of trajectory in the range of $\pm1^{\circ}$ (simulation source code can be downloaded from \url{http://pages.isir.upmc.fr/evorob_db}.)

\subsection{Performance}
\begin{figure*}[h]
\centering
\subfloat[Performance time-line.]{\includegraphics[width=0.4\textwidth, height=0.3125\textwidth]{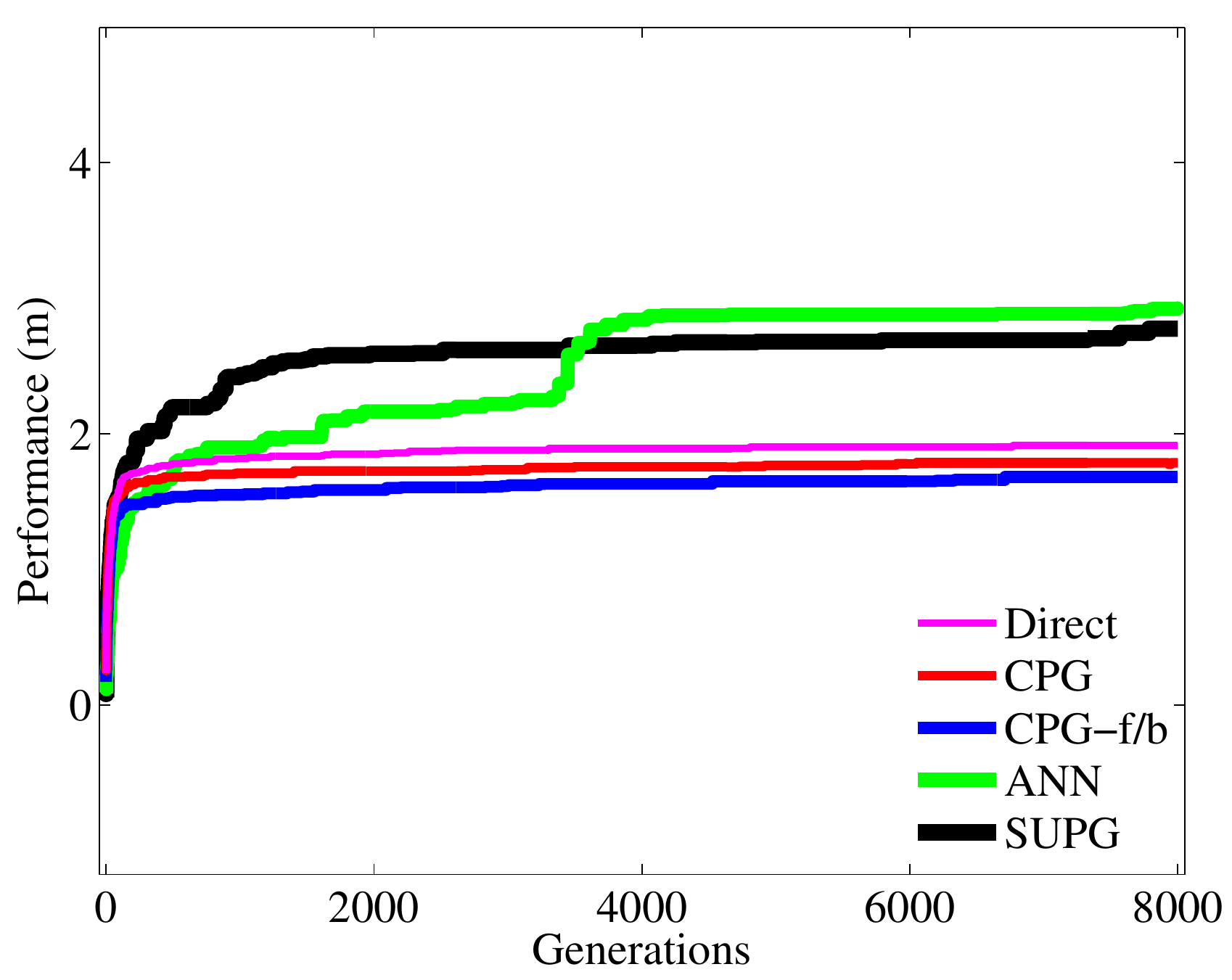}}
\subfloat[Final performance.]{\includegraphics[width=0.4\textwidth, height=0.3125\textwidth]{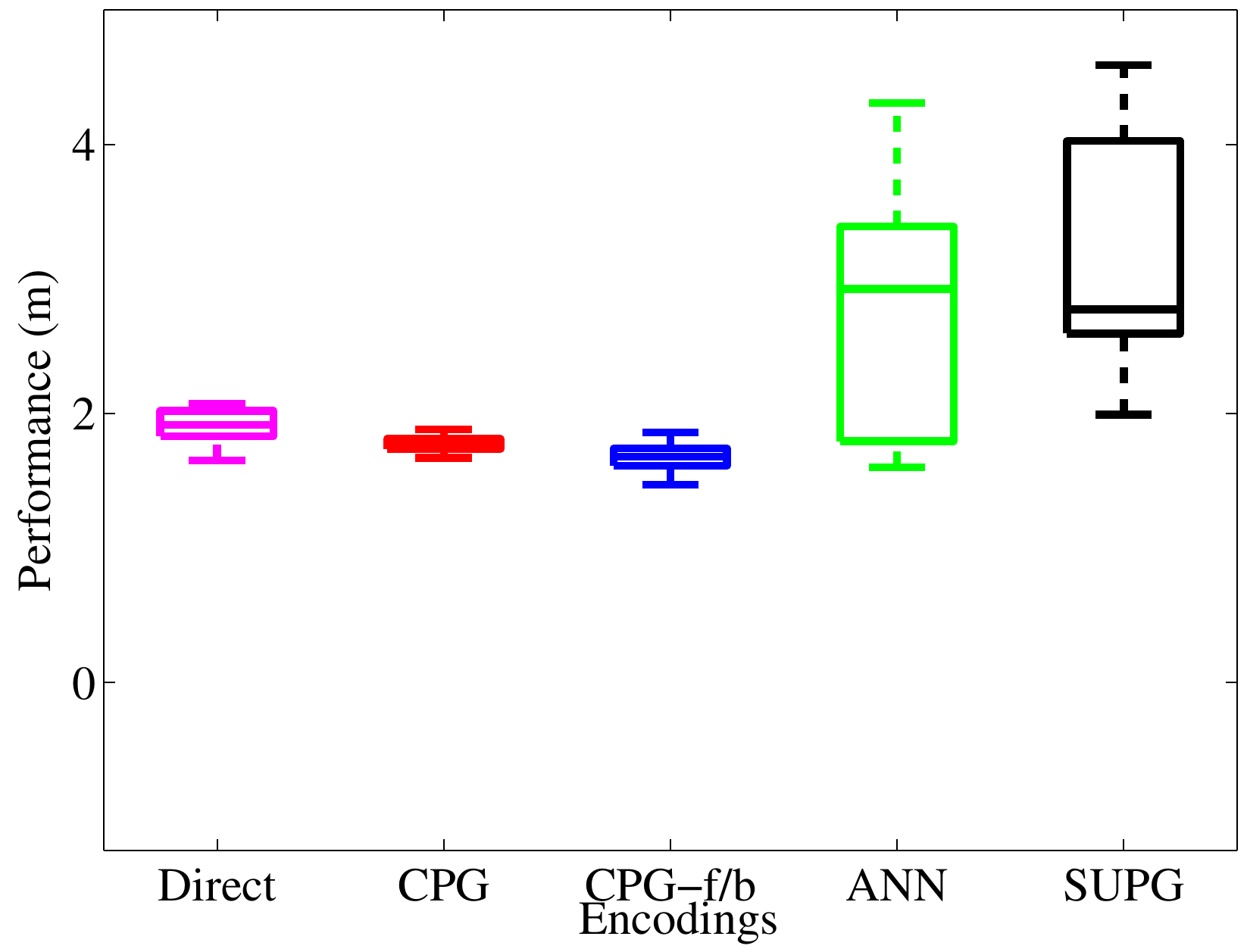}}
\caption{Performance in forward displacement for the Direct, CPG, CPG-f/b, ANN and SUPG encoding schemes: (a) Median performance for $8,000$ generations of selection; and (b) Performance of encodings at generation $8,000$.\protect\footnotemark}
\label{fig:performance}
\end{figure*}

In all five encodings, the performance of the best individuals rapidly increased with a quasi-stable equilibrium being reached with less than $5,000$ generations of selection~(Fig.~\ref{fig:performance}a).~Additionally, individuals with evolved CPGs (Direct, CPG and CPG-f/b) converged more rapidly as compared to those encoded with the ANN and SUPG schemes (Fig.~S4, generations $0$ to $8,000$). After $8,000$ generations, the performance in forward displacement of the Direct, CPG, CPG-f/b, ANN and SUPG encodings was $1.92\pm0.19$, $1.79\pm0.08$, $1.68\pm0.13$, $2.93\pm1.60$ and $2.78\pm1.43$ m, respectively (Median$\pm$IQR, see Fig.~\ref{fig:performance}b, Kruskal-Wallis test: $p < 0.001$). The ANN and SUPG schemes achieved the highest performance values across all five encodings ($d.f.= 38$, all $p < 0.001$, using Matlab's Mann-Whitney test, which is the default statistical test unless otherwise specified), but with no significant difference in performance between them. Furthermore, amongst the Direct, CPG, and CPG-f/b encodings, a decrease in performance was detected with a generative encoding, and with the inclusion of a feedback mechanism ($d.f.=38$, $p < 0.001$). However, this decrease in performance was small, and did not exceed $12\%$ (detailed statistical comparison provided in Table~S1).

Importantly, intrinsic inter-encoding differences existed in the frequencies of oscillation governing leg actuation. The frequency of the CPG oscillations was prefixed at $1$~Hz, irrespective of the sensory feedback provided, and the direct or generative nature of the encoding. By contrast, the individuals evolved with ANN and SUPG schemes were capable of expressing higher frequency oscillations ($2.44\pm1.95$~Hz for ANN, and $3.81\pm0.73$~Hz for SUPG), and the frequency of the gait may itself be under selection. \emph{Consequently, an assessment of the encodings solely on the basis of the performance is biased, and other measures are needed to compare encodings}.

\subsection{Evolvability analysis}
The evolvability provided by the encoding schemes is analyzed by mutating the best individual of each replicate at generation $8,000$, and reporting the following: (i)~The proportion decrease in performance consequent to the mutation~(eq.~\ref{eqn:evob_f1}); and (ii)~The gait diversity, computed as the mutual information between gait vectors of the original and mutated individual~(eq.~\ref{eqn:evob_f2c}). The individual is mutated at a predetermined mutation rate as used during selection, in $1,000$ separate and independent instances. Finally, a kernel density estimation\footnote{A commonly used non-parametric technique for the estimation of the probability density function of a random variable, from a finite data sample.}~\citep{scott2009multivariate} is used to visualize the resultant landscape of $20,000$ data points ($1,000$ mutations~$\times$~$20$ replicates), pooled together from all replicates.\footnote{Bivariate density estimation, with Gaussian type kernels over a grid of $100 \times 100$ equidistant points.} 

\begin{figure*}[ht]
\begin{center}
\includegraphics[width=16.5cm]{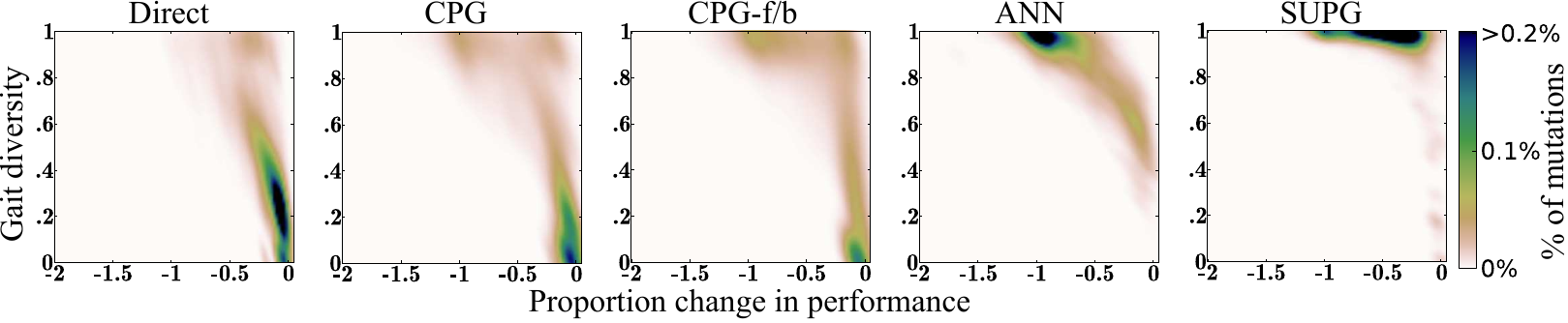}
\end{center}
\caption{Evolvability signatures for different encodings: Gait diversity and the proportion decrease in performance, following $20,000$ independent mutations of the best individuals, for the Direct, CPG, CPG-f/b, ANN, and SUPG encoding schemes after $8,000$ generations of selection, pooled from all $20$ replicates. In good evolvability signatures, mutations are located in the upper-right corner of the signature space, indicating high gait diversity, and a robustness to deleterious and lethal mutations.}
\label{fig:evolvability}
\end{figure*}

Distinct evolvability signatures were exhibited by the Direct, CPG, CPG-f/b, ANN, and SUPG encoding schemes, after $8,000$ generations of selection (see Fig.~\ref{fig:evolvability}). In evolvability analysis with a direct encoding, a conservative exploration of the phenotype, limited to solutions close to the unmutated individuals was found ($11.9\%$ and $0.33$, median decrement in performance and gait diversity, respectively). A generative encoding of the CPG model had only a minor effect on the evolvability provided (performance decrement of $12.7\%$ and gait diversity of $0.36$). The inclusion of a feedback mechanism in the generatively encoded CPG model resulted in more diverse gaits ($0.74$), but with not much change in the performance loss ($16.6\%$) following mutations. By contrast, the generative encoded ANN and SUPG schemes were much more aggressive in the exploration of the phenotypic landscape, with the gait diversity of mutated individuals at $0.95$ for ANN, and $0.99$ for the SUPG encodings. However, differences existed in the severity of negative effects of mutations amongst the two encoding schemes. The ANN encoded individuals were sensitive to the effects of deleterious mutations, resulting in a $78.9\%$ drop in performance. In comparison, individuals evolved with the SUPG encoding were much more resilient to the negative effects of mutations, with a smaller decrement of $43.1\%$ in performance following mutation. 

\begin{figure*}[h]
\centering
\subfloat[Beneficial mutations generated by different encodings.]{\includegraphics[width=0.4\textwidth]{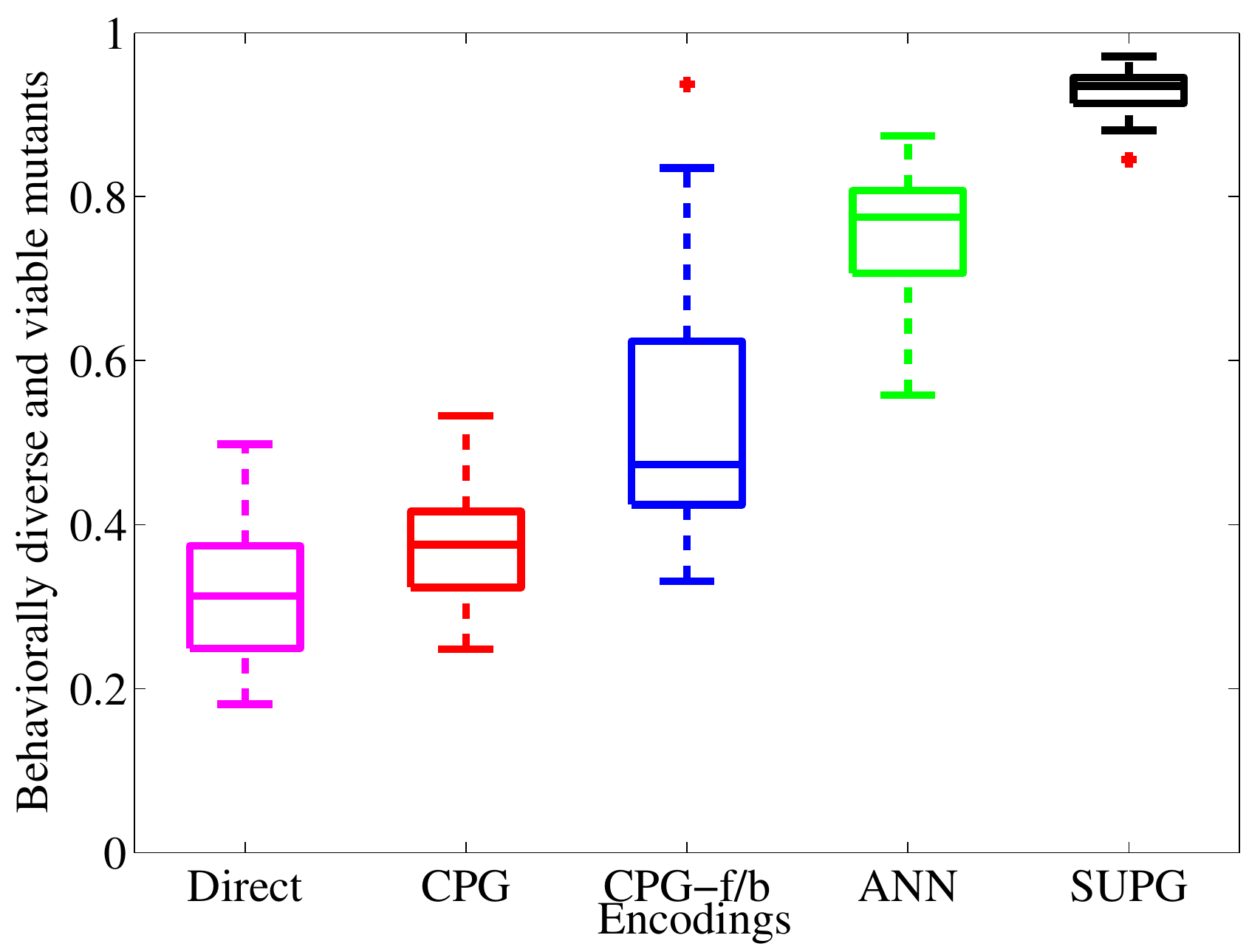}}\hspace{0.75cm}
\subfloat[Shaded region of beneficial mutations.]{\includegraphics[width=0.25\textwidth]{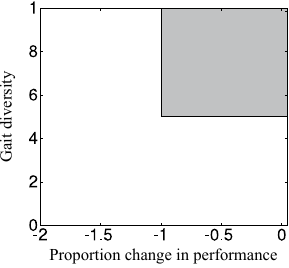}}
\caption{Proportion of viable mutants with gait diversity in excess of $0.5$, from $1,000$ independent mutations of the best individuals at generation $8,000$ in each of $20$ replicates, for the Direct, CPG, CPG-f/b, ANN and SUPG encoding schemes. These beneficial mutations are counted from the shaded region of each encoding's evolvability signature.}
\label{fig:evolvability_viabdivmut}
\end{figure*}

The evolvability provided by the encodings is further analyzed by computing the number of mutations in the evolvability signature (Fig.~\ref{fig:evolvability}), that are both non-lethal and result in diverse locomotion gaits. Mutations are classified as lethal if they result in performance decrement in excess of $100\%$ ($f_1 < -1$, see eq.~\ref{eqn:evob_f1}), corresponding to the failure of any forward movement by the robot. Similarly, a mutation is considered to generate a diverse gait, if the inter-gait diversity exceeds $0.5$ ($f_2 > 0.5$, see eq.~\ref{eqn:evob_f2c}).

The proportion of viable and diverse-gait generating mutations was affected by the encoding scheme (see Fig.~\ref{fig:evolvability_viabdivmut} and Table~S2, Kruskal-Wallis test: $p < 0.001$). Across the five encodings, the SUPG scheme was most efficient at generating such mutations ($d.f.= 38$, all $p < 0.001$). Both the ANN and the CPG-f/b encodings led to an intermediate number of viable and diversity generating mutations (ANN significantly higher than CPG-f/b, and both different from all other encodings, all $p < 0.001$). The lowest mutation count was achieved by the Direct and CPG encoding schemes (not significantly different from each other $p = 0.03$, but different from the three other encodings, all three $p < 0.001$). Thus, across the five encoding schemes, the SUPG approach provided the highest evolvability, with the capability to explore very different but viable gaits. Additionally, with a stricter definition of viable mutants resulting in no more than $50\%$ drop in performance, the SUPG still achieved the highest beneficial mutation count (all $p < 0.001$), with no difference between the other four encodings (see Fig.~S1).

\footnotetext{On each box, the mid-line marks the median, and the box extends from the lower to upper quartile below and above the median. Whisker outside the box generally indicate the maximum and minimum values, except in case of outliers, which are shown as crosses. Outliers are data points outside of $1.5$ times the interquartile range from the border of the box.}

\subsubsection{Evolvability under varying mutation intensities}
In this section, we study the sensitivity of our signature of the evolvability provided by the Direct, CPG, CPG-f/b, ANN and SUPG encodings with respect to the parameters of the variation operator used to generate mutants. The main questions are, if and how differences in the mutation operator affect our conceived signature of evolvability? We ran a series of experiments to access evolvability, with genetic mutants generated at different intensities. Mutations were considered at the standard mutation rate and mutation step-size as used during selection (medium intensity), and at a four-fold decrease (low intensity) and a four-fold increase (high intensity) of the standard mutation operator parameters of both rate and step-size.

\begin{figure*}
\begin{center}
\includegraphics[width=16.5cm]{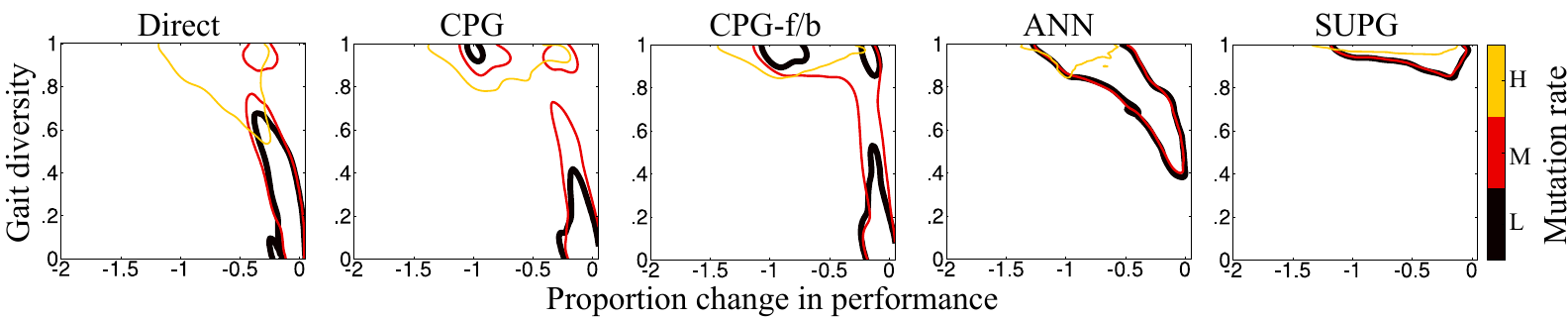}
\end{center}
\caption{Contours of the evolvability signature for the Direct, CPG, CPG-f/b, ANN, and SUPG encoding schemes with mutations generated at low (L), medium (M), and high (H) intensity, of the best individuals at generation $8,000$ of selection. The low, medium, and high mutation intensities were quantitatively $0.25$, $1$, and $4$ times respectively, the mutation rate and step-size used during selection. Individual contours encompass a $0.025\%$ or higher density of mutants.}
\label{fig:evolvability_mutrate}
\end{figure*}

In order to analyze the effect of the variation operator on our signature of evolvability, in Fig.~\ref{fig:evolvability_mutrate} we outline the perimeter of the evolvability signatures generated from low, medium and high intensity mutations (see Fig.~S2 for interior of signature). In all five encodings, the distribution of mutants shift towards more diverse gaits and is accompanied by larger loss in performance, following increments in the mutation intensity. A $16$-fold increment in the mutation rate and step-size (low to high mutation intensity) resulted in a $0.94$, $0.96$ and $0.98$ difference in mutant gait diversity for the Direct, CPG and CPG-f/b encodings respectively (see Fig.~S3a, b and c). By contrast, the ANN and SUPG schemes achieved highly diverse gaits at the lowest mutation intensity ($0.94$ for ANN and $0.99$ for SUPG), and an increment from low to high mutation intensity only resulted in a $6.0\%$ for ANN, and $0.4\%$ for SUPG further increase in the mutated gaits diversity (Fig.~S3d and e). The increment from low to high mutation intensity also resulted in a drop in mutant performance of $63.7\%$, $90.5\%$, $93.5\%$, $98.6\%$ and $54.9\%$, for the Direct, CPG, CPG-f/b, ANN, and SUPG encodings respectively (Fig.~S3f-j). In summary, the SUPG encoding facilitates the exploration of diverse gaits even when mutants are generated at a low mutation intensity. Furthermore, across all five encodings, the SUPG scheme provides the most resilience against the deleterious nature of high intensity mutations.

\subsection{Damage recovery}
The significance of our evolvability signatures of the Direct, CPG, CPG-f/b, ANN and SUPG encodings was investigated by analyzing the adaptation of the evolved robot's gait, following the removal of one or more of its legs. We expect that for the encodings registering a better evolvability signature, the corresponding evolved individuals would require fewer generations to recover an effective walking gait.

\begin{figure*}
\centering
\subfloat[Scenario 1: Removal of right-middle leg.]{\includegraphics[width=4.205cm, height=4.16cm]{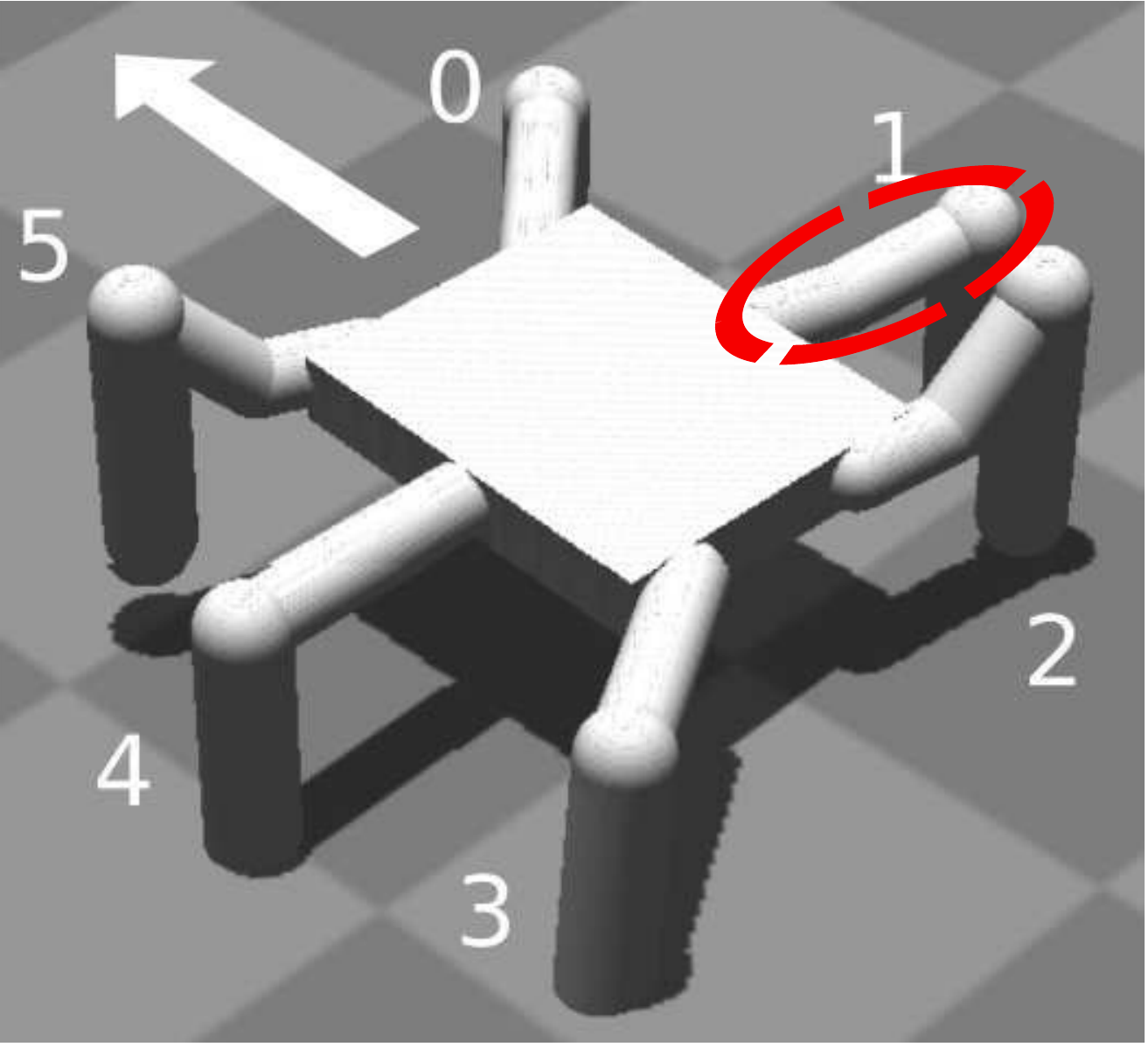}}\hfil
\subfloat[Scenario 2: Removal of right-middle and left-middle legs.]{\includegraphics[width=4.205cm, height=4.16cm]{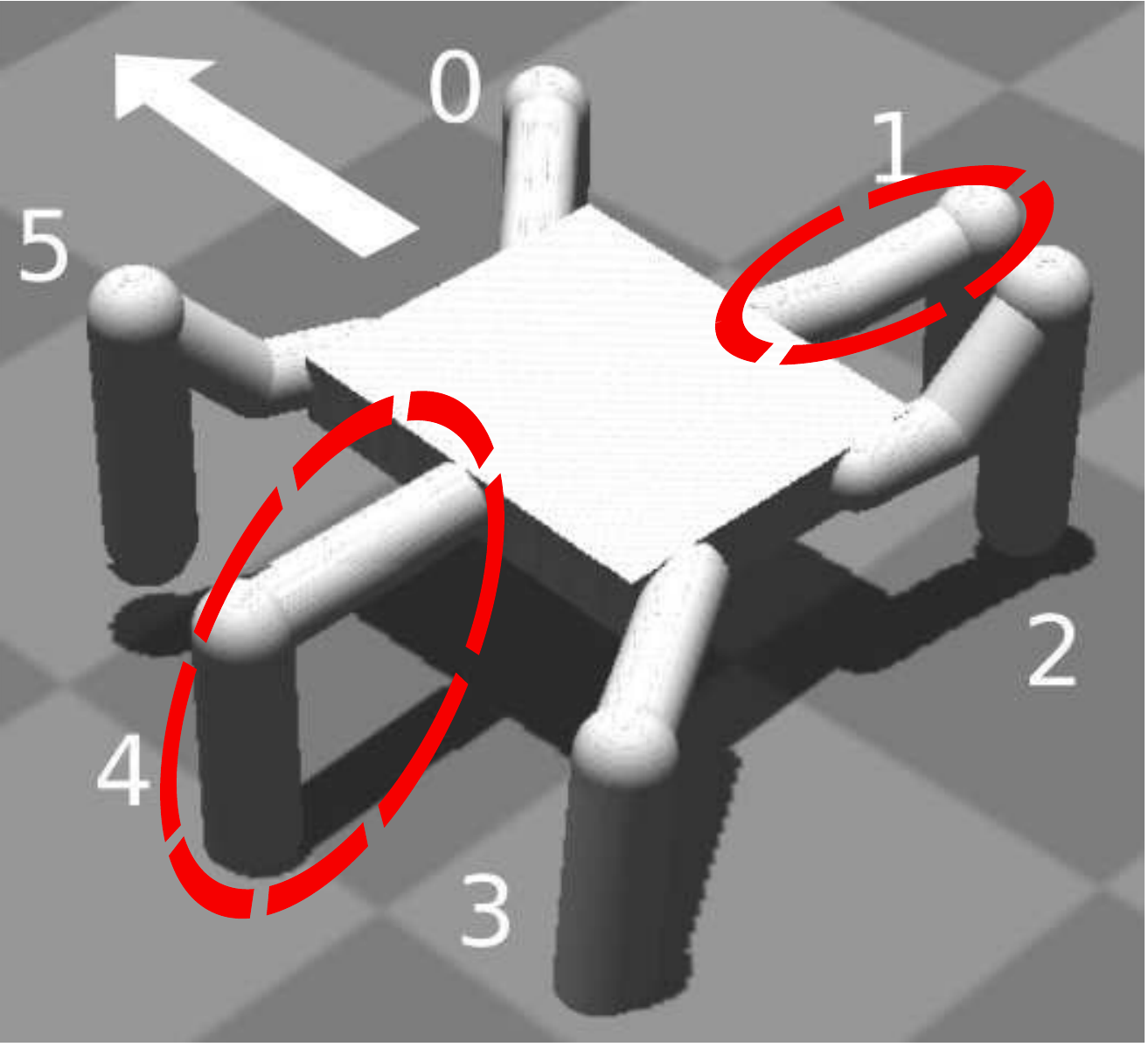}}\hfil
\subfloat[Scenario 3: Removal of right-middle and left-rear legs.]{\includegraphics[width=4.205cm, height=4.16cm]{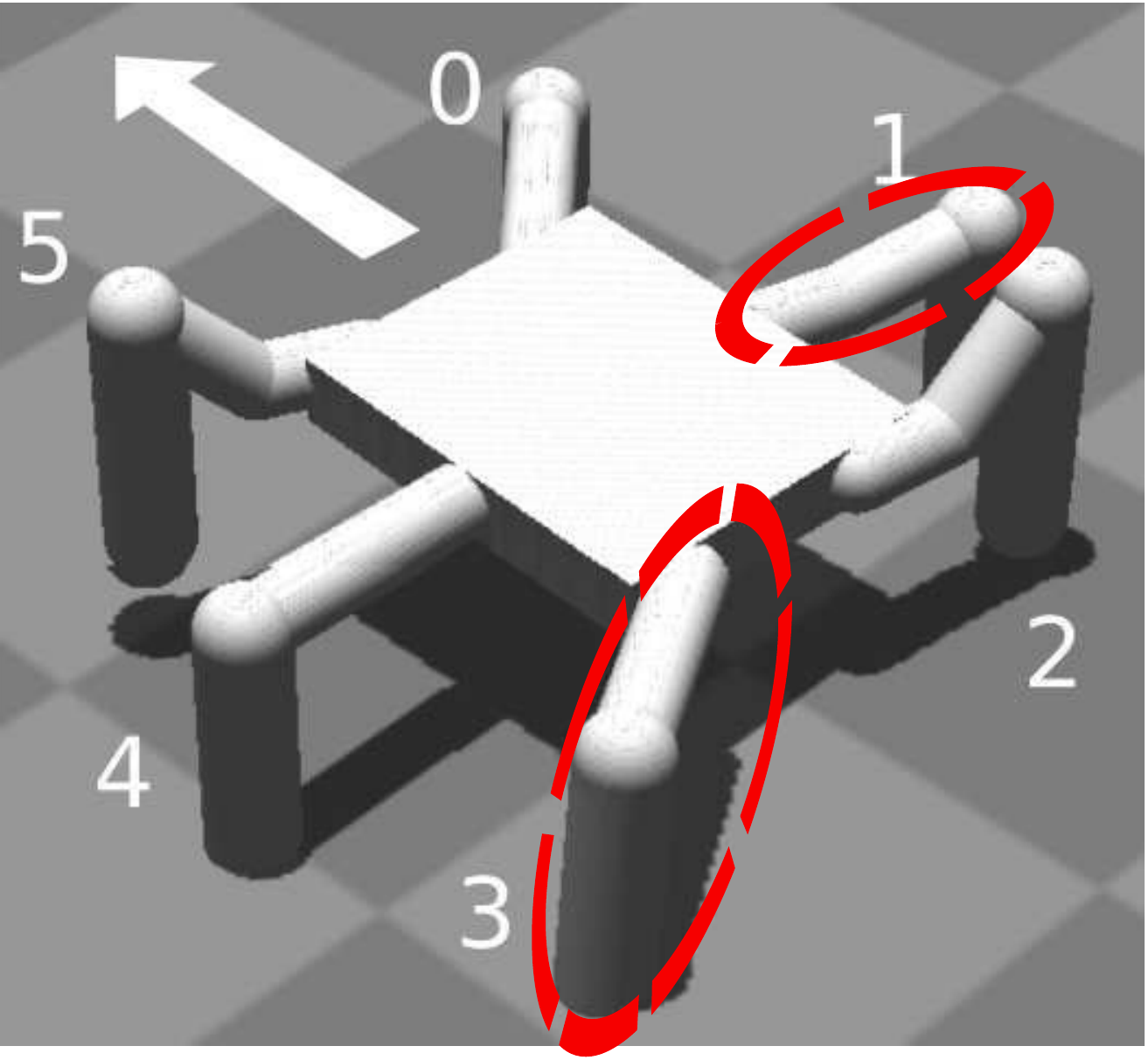}}
\caption{The three damage scenarios imposed on the hexapod robot (undamaged robot in Fig.~\ref{fig:robot}).}
\label{fig:damaged_hexapod}
\end{figure*}

In these experiments, the new (damage recovery) populations were comprised of $100$ mutated individuals of the best individual of each replicate at generation $8,000$ of selection. In separate preliminary experiments, the use of the entire population at generation $8,000$ (instead of the best individuals) did not change our results of the adaptability provided by different encodings. Individuals in the damage recovery population were mutated at the standard mutation rate and step size used during selection. A further $10,000$ generations of artificial selection was conducted on the populations of amputee hexapods for each of the following three damage scenarios: (i)~an asymmetrical damage, following the removal of one leg of the robot (leg $1$, Fig.~\ref{fig:damaged_hexapod}a); (ii)~a symmetrical damage occurs, wherein the two middle legs on either side of the robot are removed (legs $1$ and $4$, Fig.~\ref{fig:damaged_hexapod}b); and (iii)~a highly asymmetrical damage occurs consequent to the removal of the middle leg on one side and the rear leg on the opposing side of the hexapod (legs $1$ and $3$, Fig.~\ref{fig:damaged_hexapod}c). The number of generations required to regain an effective gait and the proportion of the original performance (undamaged robot's performance at generation $8,000$) recovered for each of the three damage scenarios is analyzed.

\begin{figure*}[h]
\centering
\subfloat[Scenario 1: Removal of right-middle leg.]{\includegraphics[width=0.3\textwidth, height=0.237\textwidth]{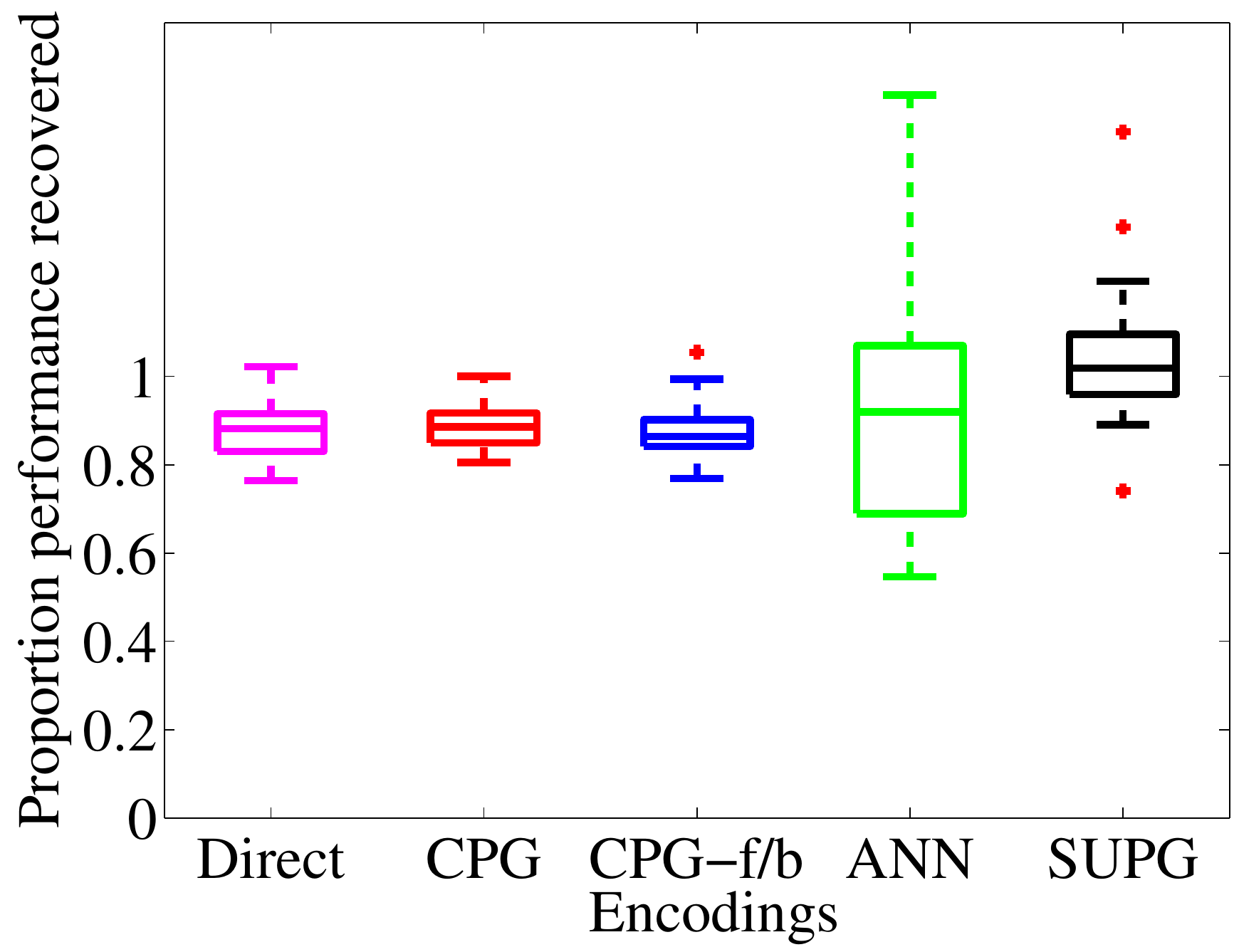}}\hfil
\subfloat[Scenario 2: Removal of right-middle and left-middle legs.]{\includegraphics[width=0.3\textwidth, height=0.237\textwidth]{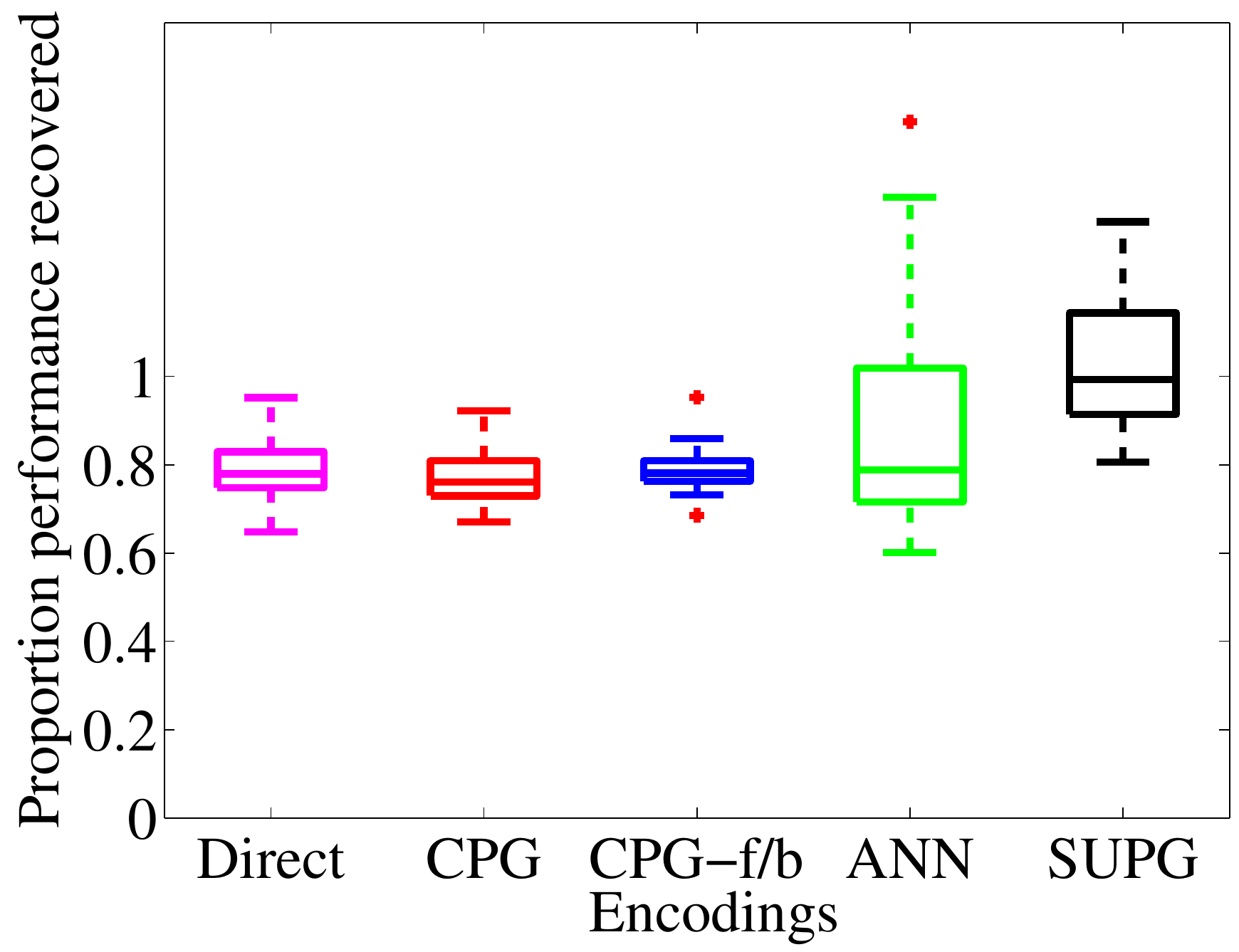}}\hfil
\subfloat[Scenario 3: Removal of right-middle and left-rear legs.]{\includegraphics[width=0.3\textwidth, height=0.237\textwidth]{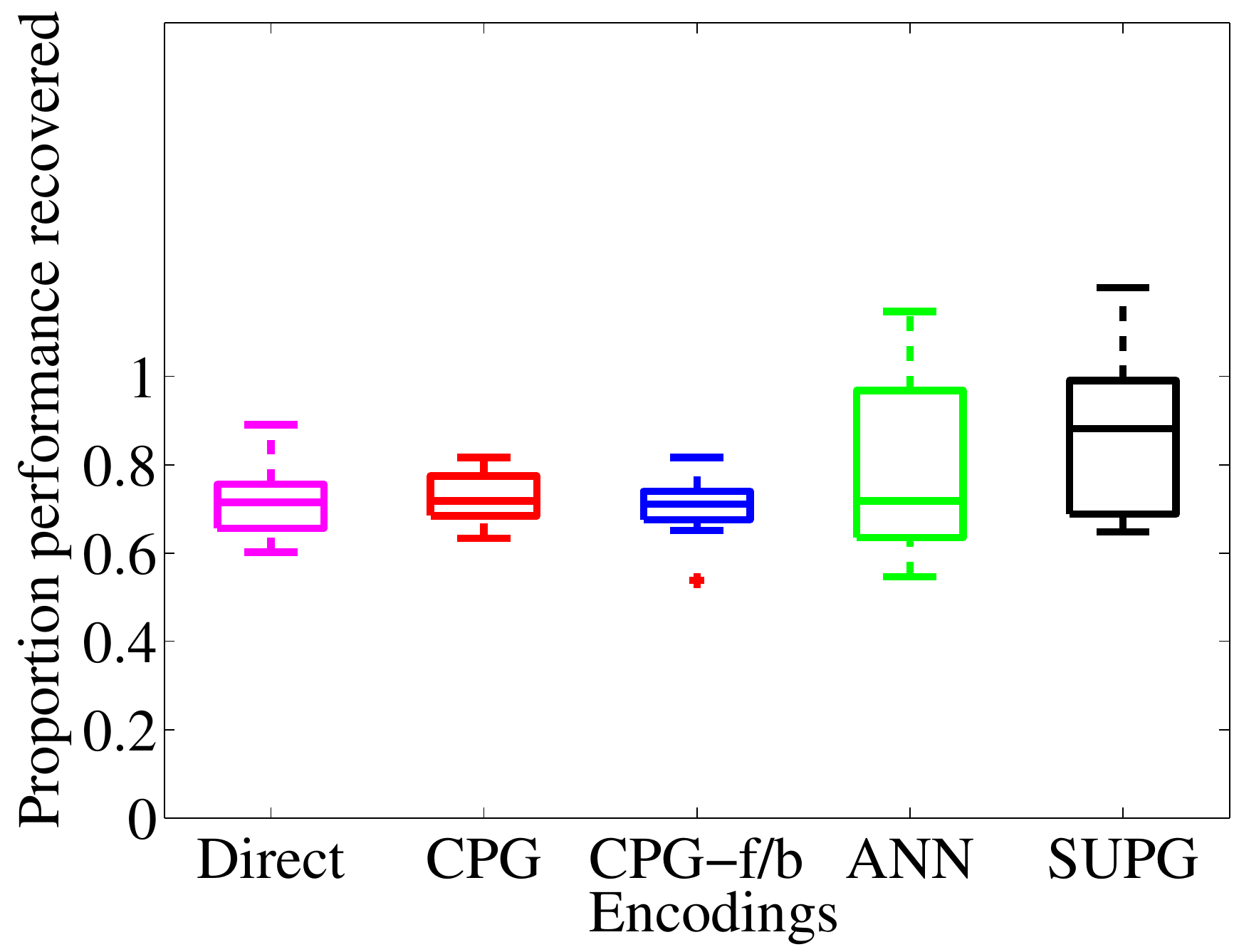}}
\caption{Proportion of the original performance in forward displacement restored $10,000$ generations after the three damage
scenarios, across $20$ replicates, for the Direct, CPG, CPG-f/b, ANN and SUPG encoding schemes.}
\label{fig:recov_performance_finalperform}
\end{figure*}

In the $10,000$ generations of selection, all the five encodings were capable of recovering a majority of their original performance in forward displacement, irrespective of the damage to the hexapod robot (see Fig.~S4). After $10,000$ generations post robot damage, the Direct, CPG, CPG-f/b, ANN and SUPG schemes all recovered the highest proportion of their original performance in the first damage scenario ($0.89\pm0.07$, Median$\pm$IQR across all encodings), followed by an intermediate recovery in the second ($0.78\pm0.07$), and third  ($0.72\pm0.05$) scenarios (Fig.~\ref{fig:recov_performance_finalperform}a, b and c, Kruskal-Wallis test: $d.f.=4$, $p < 0.001$ for first two scenarios, and $p = 0.018$ for third scenario). Across all five encodings, the SUPG was most efficient in recovering its original performance in the first ($1.02\pm0.14$) and second ($0.99\pm0.23$) scenarios (for both, all $p < 0.001$), while in both scenarios no significant difference in recovery was registered between the remaining four encodings (Fig.~\ref{fig:recov_performance_finalperform}a and b). In the third scenario which was the hardest, the SUPG again achieved the highest performance recovery ($0.88\pm0.3$), although performance was no longer significantly different between encodings (Fig.~\ref{fig:recov_performance_finalperform}c).

\begin{figure*}[h]
\centering
\subfloat[Scenario 1: Removal of right-middle leg.]{\includegraphics[width=0.3\textwidth, height=0.237\textwidth]{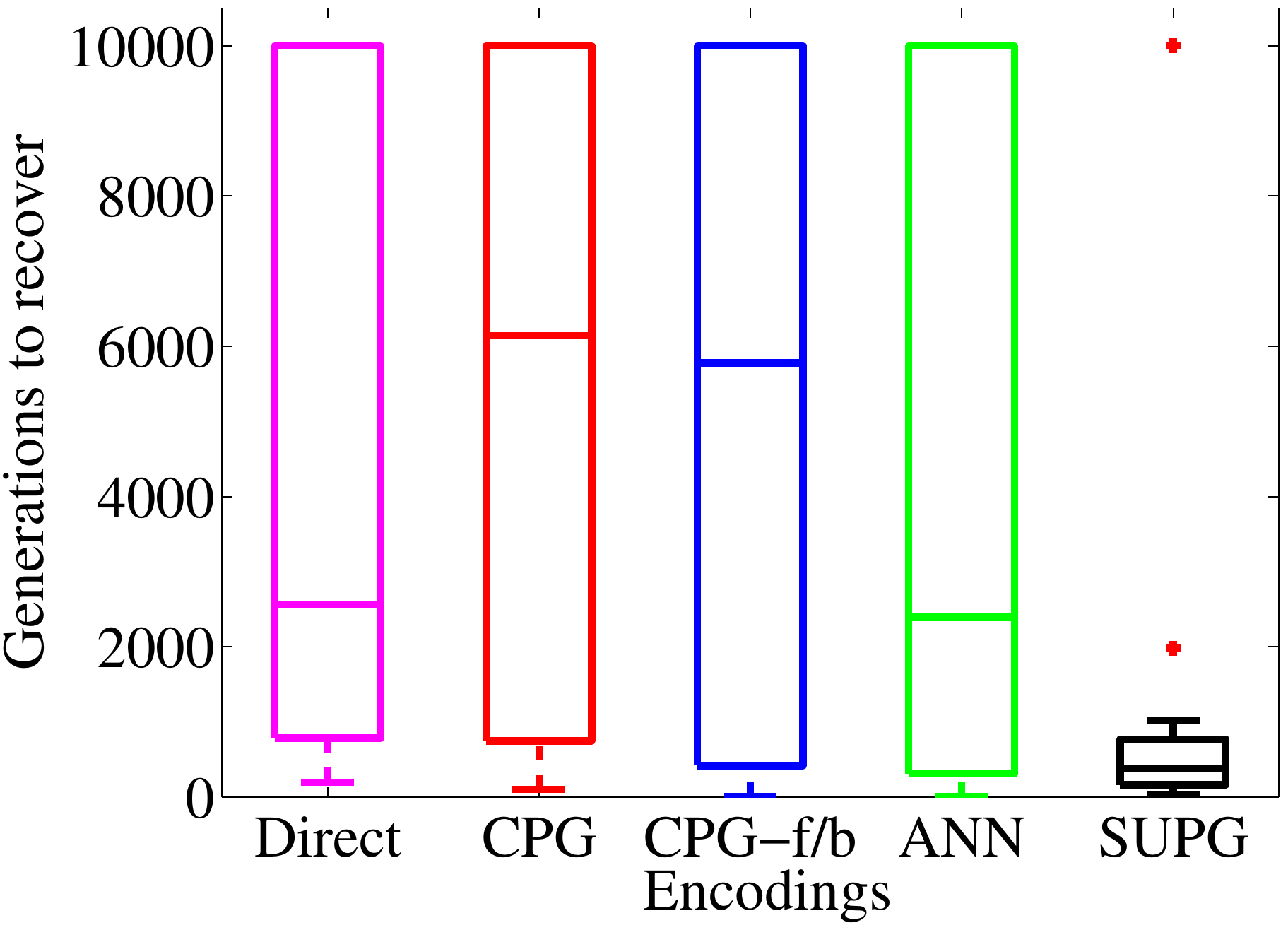}}\hfil
\subfloat[Scenario 2: Removal of right-middle and left-middle legs.]{\includegraphics[width=0.3\textwidth, height=0.237\textwidth]{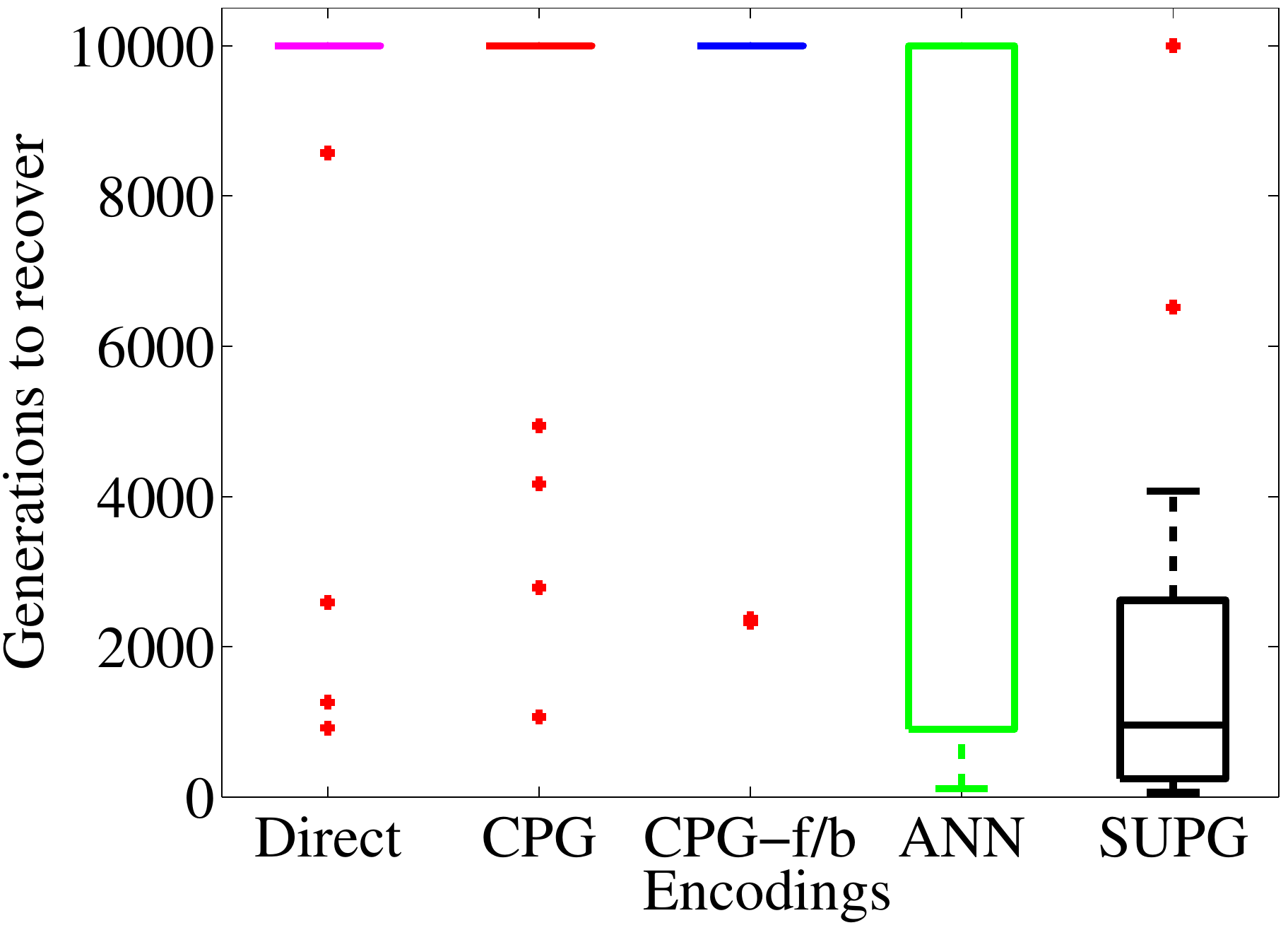}}\hfil
\subfloat[Scenario 3: Removal of right-middle and left-rear legs.]{\includegraphics[width=0.3\textwidth, height=0.237\textwidth]{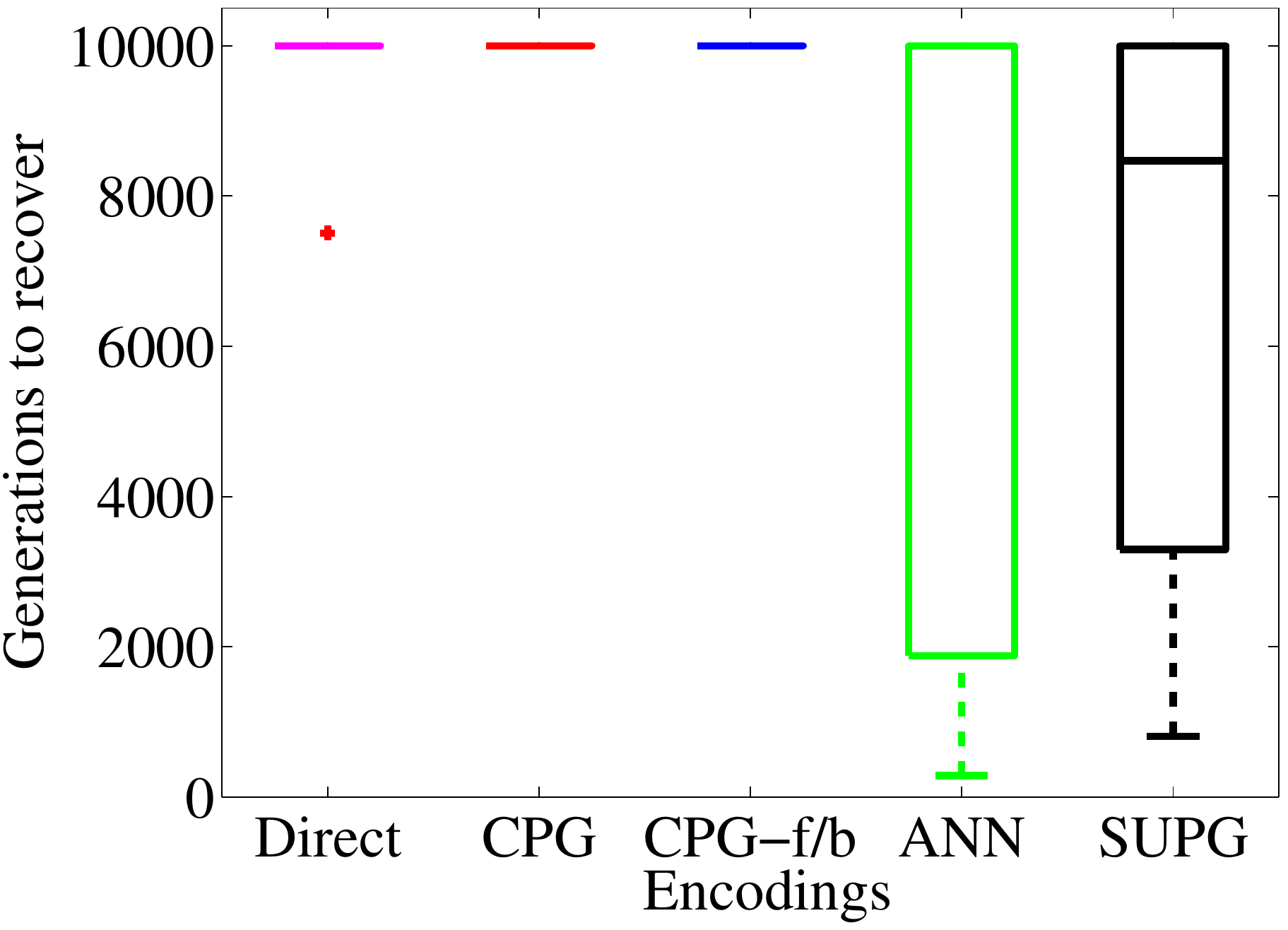}}
\caption{The number of generations of selection required to restore $85\%$ of the original performance of the undamaged hexapod in forward displacement across $20$ replicates, for the Direct, CPG, CPG-f/b, ANN, and SUPG encoding schemes. In replicates unable to attain the $85\%$ mark, the recovery time was set to the upper limit of $10,000$ generations.}
\label{fig:recov_performance_numgen}
\end{figure*}

In order to analyze the time required by damaged hexapods to recover an effective gait, in Fig.~\ref{fig:recov_performance_numgen} we have plotted the number of generations required to restore  $85\%$ of the original performance in forward displacement. Across the three robot-damage scenarios, amputee hexapods in the first scenario achieved the highest recovery rate ($13\pm3.25$ of $20$ replicates restored performance in $10,000$ generations, across all encodings), followed by an intermediate recovery rates in the second ($4\pm7.75$ replicates), and third scenarios ($1\pm8.25$ replicates). In the first two scenarios (Kruskal-Wallis test: $d.f.=4$, both $p < 0.001$), the SUPG encoded individuals recovered at least an order of magnitude faster ($373$ and $957$ generations in scenarios $1$ and $2$, respectively) than  individuals with the Direct, CPG, CPG-f/b and ANN encodings (see Fig.~\ref{fig:recov_performance_numgen}a and b, all $p < 0.001$). In both scenarios, no significant difference in recovery was registered between these four encoding schemes. In the third scenario (Kruskal-Wallis test: $d.f.=4$, $p < 0.001$), the SUPG encoded amputee hexapods continued to exhibit the fastest recovery ($8466.5$ generations), although it was no longer significantly different from ANN encoded hexapods (see Fig.~\ref{fig:recov_performance_numgen}c, $p=0.25$). Furthermore, in this scenario, all three CPG based encodings (Direct, CPG, and CPG-f/b) performed poorly, with only one replicate making the $85\%$ mark in the $10,000$ generations of selection. In summary, across all five encoding schemes, the SUPG encoded individuals had the fastest recovery, despite being in increasingly difficult robot-damage scenarios wherein most of the individuals encoded by the other encodings failed to recover an effective walking gait (detailed comparison of damage recovery in Tables~S3, S4 and S5).

\section{Discussion}
\label{sec:discussion}
In this study, we introduce a methodology to compare genetic encoding schemes, utilizing our novel approach to signaturize evolvability. The evolvability provided by the generatively encoded CPG, CPG-f/b, ANN~\citep{clune2011,yosinski2011} and SUPG~\citep{morse2013} schemes and a direct encoding, is characterized by quantifying both, (i)~the quality of the genetic mutations, and (ii)~the quantity of behavioral variation generated from genetic change.~Our evolvability based inter-encoding comparison methodology provides insights on the adaptability of evolved individuals to novel scenarios. To this effect, the significance of our conceived signature of evolvability is evaluated by the individual adaptation response to morphological changes previously unencountered by the hexapod robot.

Our results revealed a direct relationship between the estimated evolvability provided by the encodings, and the capability of the evolved individuals to adapt to severe changes in morphology, simulated by the amputation of one or more of the hexapod legs. Amongst the five encodings evaluated, the SUPGs \citep{morse2013} had the best evolvability signature, and their encoded individuals were also foremost to recover following sustained damages. In both the easy and the intermediate robot-damage scenarios (scenarios 1 and 2), the SUPG encoded individuals were capable of recovering $85\%$ of their performance on the undamaged robot in all but two replicates, and did so more than an order of magnitude faster than the other four encodings. Furthermore, even in the most difficult robot-damage scenario (scenario 3), the SUPG scheme achieved the fastest recovery in the majority of the evaluated replicates. 

The ANN encoding scheme (minimal HyperNEAT) was capable of producing highly diverse hexapod gaits, following genetic mutations. The high behavioral diversity generated is consistent with the earlier use of this encoding to evolve gaits for the QuadraBot robot \citep{clune2009evolving,clune2011}. However, in our ANN implementation, most of genetic mutations producing diverse gaits were highly deleterious, and resulted in little forward hexapod movement. Consequently, an estimate of evolvability solely on the basis of the generated behavioral diversity \citep{lehman2011improving,reisinger2005towards,lehman2013evolvability} is not reliable, and both the quality (individual viability) and quantity of phenotypic variation consequent to genetic change is required to characterize evolvability. Furthermore, the poor evolvability signature for the ANN encoding scheme is reflected in its poor recovery from sustained robot damages. 

In studies on evolution of multilegged robot locomotion, the generative encodings exploit the symmetry of the robot morphology to generate regular and coordinated gait patterns that often outperform gaits evolved with direct encodings (e.g., \cite{clune2011}). Generative encodings also facilitate scalability, wherein evolution in the low-dimensional genetic search space is capable of evolving complex phenotypes comprising of many more dimensions \citep{stanley2003taxonomy}. However, no difference in performance was registered between the directly and generatively encoded CPGs for our hexapod locomotion problem, perhaps consequent to the already low-dimensional search space for the directly encoded locomotion controllers. For example, our directly encoded CPGs for hexapod locomotion comprise $23$ amplitude and phase bias parameters, in contrast to the $800$ \textit{Fixed-Topology NEAT} (FT-NEAT) encoded neural weight parameters for quadruped locomotion controllers \citep{clune2009evolving}. Thus, the potential benefits of phenotypic scalability in utilizing generative encoding schemes are reduced in our study.

For our signature of evolvability, we mutated the individuals with a predetermined mutation rate, tuned to allow a speedy convergence of the evolved solutions. This is a critical consideration as variations to the mutation rate can affect the viability and gait diversity of generated mutants. A comparison of the evolvability provided by encodings at low and high mutation intensities suggests that with an increase in mutation rate, the peak of the distribution of mutants shifts towards more diverse gaits with a larger decrease in task performance. However, the overall shape of the distribution, highlighting desirable regions of the evolvability landscape, remained the same for all five encodings. Importantly, across the five encodings, the SUPG scheme continued to provide the highest resilience to deleterious genetic change, despite a $16$-fold increase in mutation intensity.

The high evolvability and rapid recovery provided by the SUPGs may be consequent to the closed-loop control ingrained in the encoding scheme. Such a feedback mechanism provides an adaptive period of the SUPG oscillations, that can be adjusted to the new step size of a gait more appropriate for a four or five legged robot, after the hexapod has suffered damages. Alternatively, the better performance of SUPGs may be consequent to the open-ended encoding of control signals by these oscillators. In contrast to the simple sinusoidal waves of the CPG-based schemes wherein only the signal amplitude and phase difference is encoded, no constrains are imposed on the CPPNs encoding the SUPG output signals. The resulting unconstrained encoding may help for example in adjusting the duty ratio for each oscillator to match the new swing and stance phase durations of the remaining undamaged hexapod legs. Furthermore, since the inclusion of feedback in the CPG encoding (CPG-f/b) registered no improvement in task-performance, evolvability or damage recovery, a combination of the SUPG closed-loop system and the open-ended encoding of its oscillatory signal may be responsible for its high performance and adaptive capabilities.

In our evolvability signatures, the phenotypic variation from genetic change was associated with the mutual information between hexapod gaits. The diversity may also be computed for the gaits of bipedal and quadrupedal type of robots. Similarly, behavioral diversity may be computed for other benchmark problems such as, the final position of a robot in a maze navigation task \citep{lehman2011improving,mouret2012}, the final positions of balls in an arena for the robot ball-collecting task \citep{Doncieux2009,Doncieux2010}, and a vector of board piece moves in game playing tasks \citep{reisinger2007acquiring,gauci2010autonomous}. Consequently, our approach to estimate evolvability is easily applicable to a wide range of tasks, commonly used in evolutionary robotics experiments.

The systematic building and organizing of knowledge, a requirement in any scientific discipline, can not be achieved without a wide assortment of quantifiable measures to compare and contrast concepts, hypotheses, testable explanations and predictions. In the field of evolutionary robotics, task-performance has been prominently and often solely used as such a quantifiable measure to form links between the different available evolutionary systems. However, the evaluated fitness by its very nature, is limited to the specific problem for which the individual solutions are tested. By contrast, an estimate of evolvability facilitated by an evolutionary process, may be applicable to a much broader scope of scenarios, allowing the formation of a more generic relationship between existing evolutionary system implementations. An evolvability-based approach of comparison between evolutionary processes may help to extrapolate to unevaluated problem regions in-between existing evaluations, and consequently better lend itself to building a strong theoretical foundation for future research in the field.

\section{Conclusion}
\label{sec:conclusion}
A novel approach to compare encoding schemes is proposed in this paper, using evolvability instead of task-fitness, and demonstrated for the problem of evolving locomotion gaits for a hexapod robot. From the different direct and generative encodings evaluated, the single-unit pattern generator had the best evolvability signature. Furthermore, its encoded individuals were foremost in adapting to various new and challenging scenarios. Our biologically-relevant ``fingerprint'' of the evolvability provided by an encoding accounts for both the quantity and quality of phenotypic variation resulting from genetic change, and results in distinct evolvability signatures for the different encodings, despite similar task-fitness values. These evolvability signatures allow for a discrimination between encoding schemes, based on their capability of producing highly adaptable individuals for novel, and importantly, a priori unknown scenarios. In summary, our evolvability signatures for encoding comparison serves as a comprehensive analysis tool beyond the standard task-fitness performance benchmarks commonly employed in the evolutionary robotics field.

\section*{Acknowledgment} This study was supported by the ANR Creadapt project (ANR-12-JS03-0009).

\section*{Supplemental data} Movies of the hexapod robot walking behaviors are available online at \url{http://goo.gl/uyY1RX}.

\bibliographystyle{plainnat}
\bibliography{is_evolvability}

\newpage

\pagestyle{empty}

\renewcommand{\thefigure}{S\arabic{figure}}
\setcounter{figure}{0}

\renewcommand{\thetable}{S\arabic{table}}
\setcounter{table}{0}

\begin{figure*}[h]
\section*{Supplemental data}
\centering
\subfloat[Highly beneficial mutations generated by different encodings.]{\includegraphics[width=0.5\textwidth]{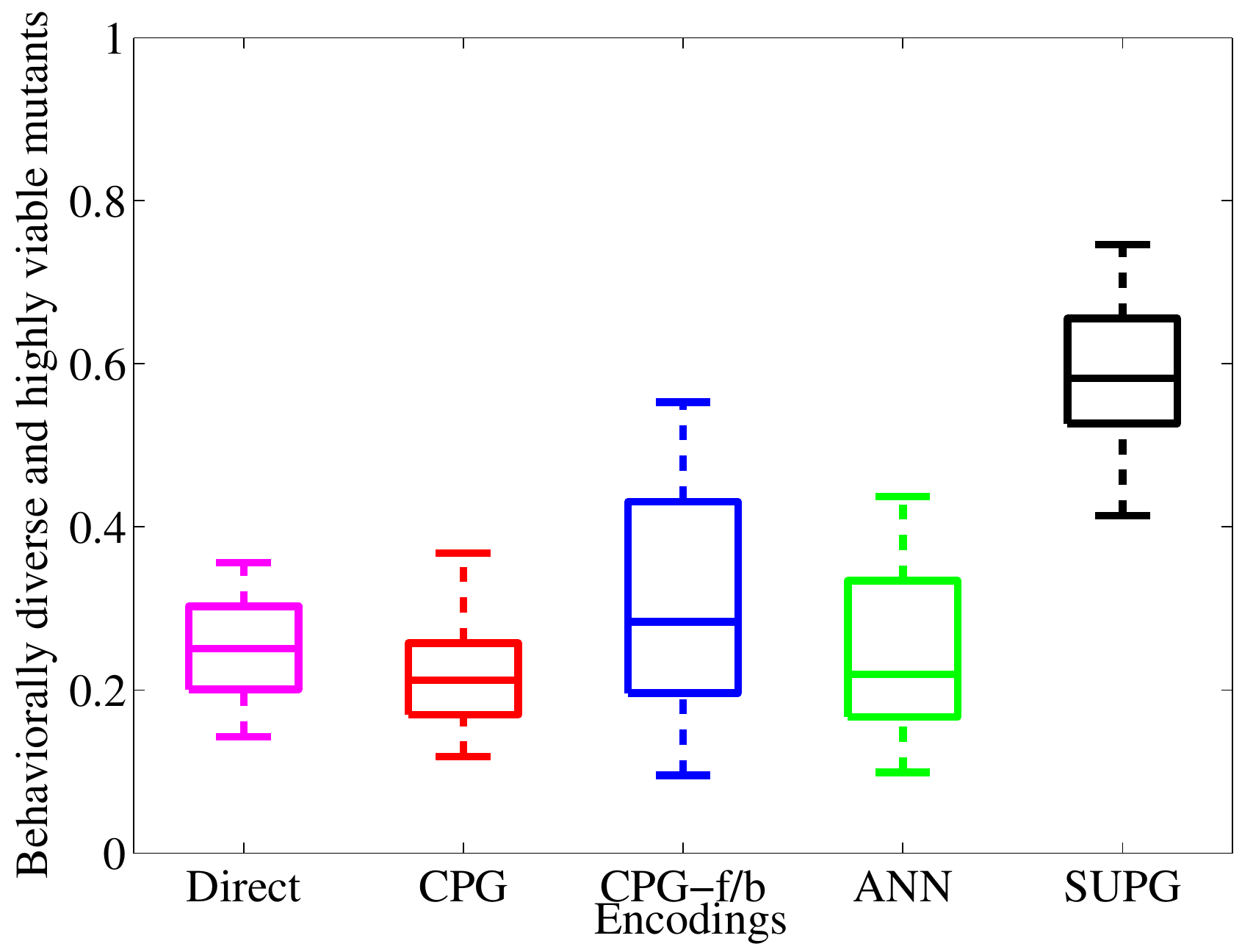}}\hfil
\subfloat[Shaded region of highly beneficial mutations.]{\includegraphics[width=0.35\textwidth]{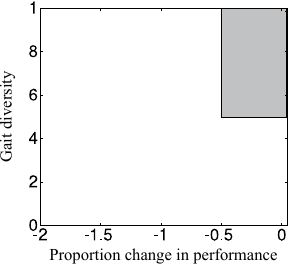}}
\caption{Proportion of mutants generating gait diversity in excess of $0.5$, and performance loss not exceeding $50\%$, from $1,000$ independent mutations of the best individuals at generation $8,000$ in each of $20$ replicates, for the Direct, CPG, CPG-f/b, ANN and SUPG encoding schemes. These highly beneficial mutations are counted from the shaded region of each encoding's evolvability signature.}
\label{fig:evolvability_viabdivmut}
\end{figure*}

\begin{figure*}[ht]
\begin{center}
\subfloat[Low mutation intensity]{\includegraphics[width=16.5cm]{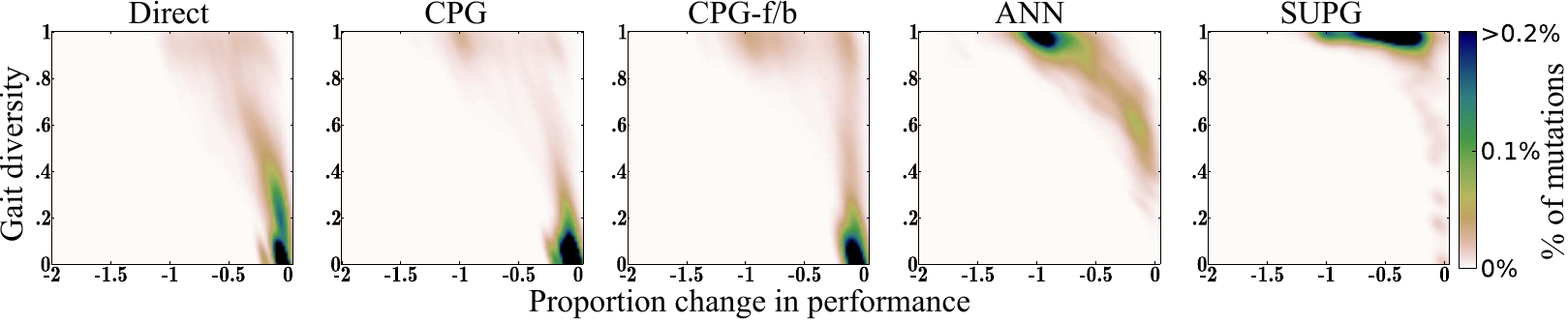}}\\
\subfloat[Medium mutation intensity]{\includegraphics[width=16.5cm]{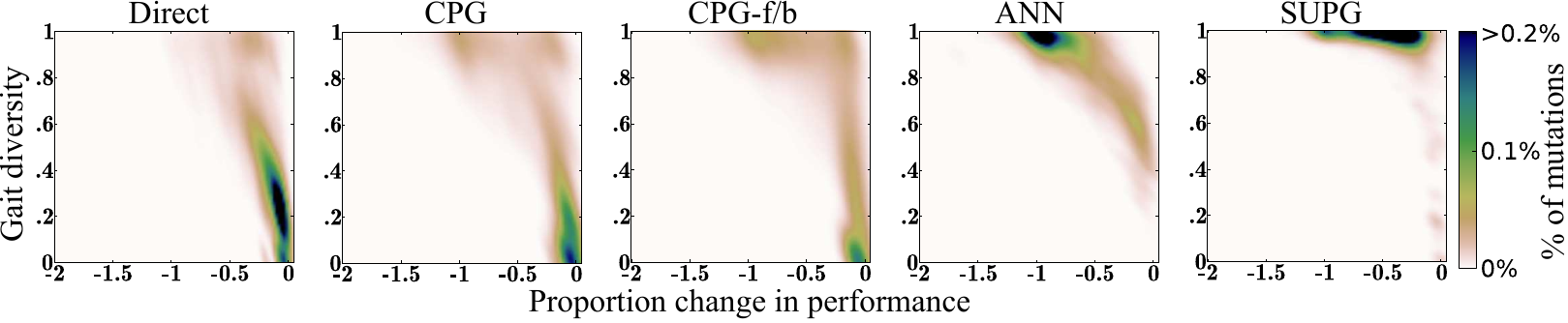}}\\
\subfloat[High mutation intensity]{\includegraphics[width=16.5cm]{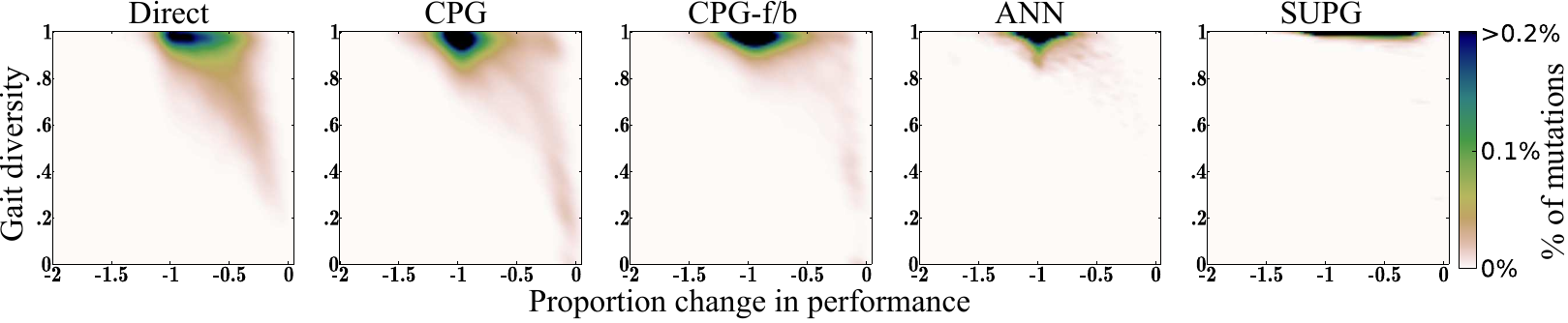}}
\end{center}
\caption{Gait diversity and the proportion decrease in performance, following $20,000$ independent mutations of different
intensities, for the Direct, CPG, CPG-f/b, ANN and SUPG encoding schemes after $8,000$ generations of selection, pooled
from all $20$ replicates. Mutations at low, medium and high intensity, corresponded to $0.25$, $1$ and $4$ times,
respectively, the mutation rate and step-size used during selection.}
\label{fig:evolvability}
\end{figure*}

\begin{figure*}[h]
\centering
\subfloat[Direct]{\includegraphics[width=3.4cm]{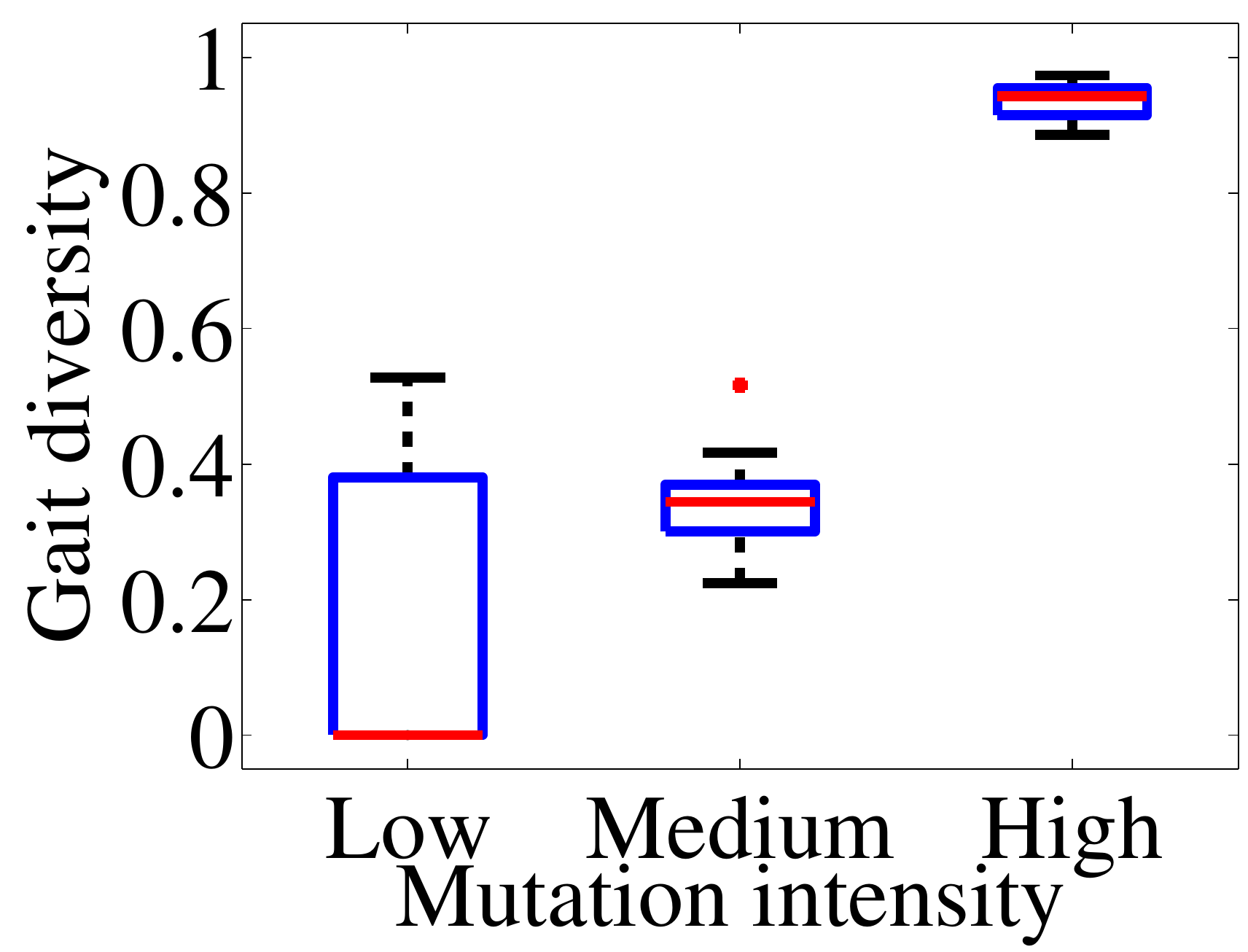}}
\subfloat[CPG]{\includegraphics[width=3.4cm]{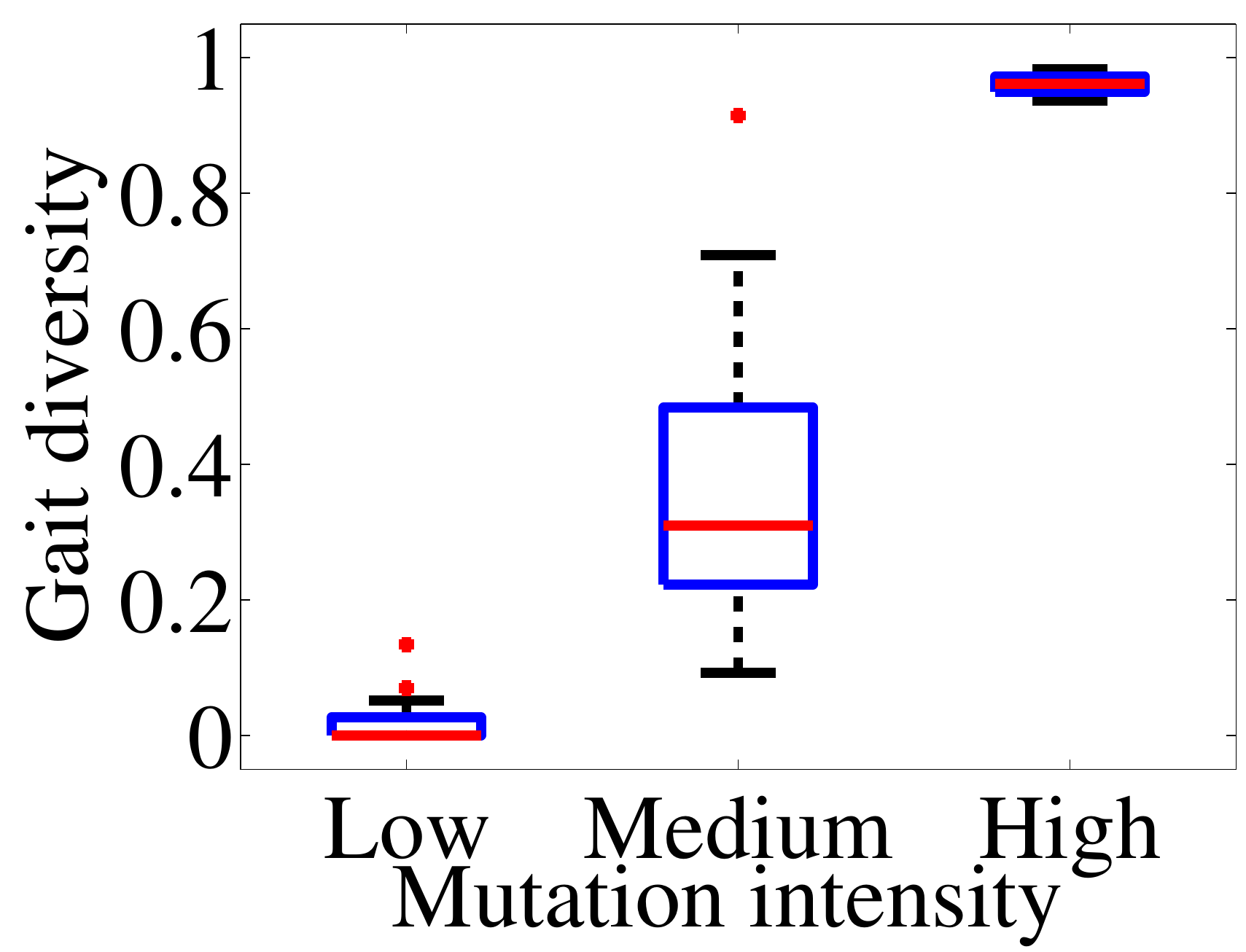}}
\subfloat[CPG-f/b]{\includegraphics[width=3.4cm]{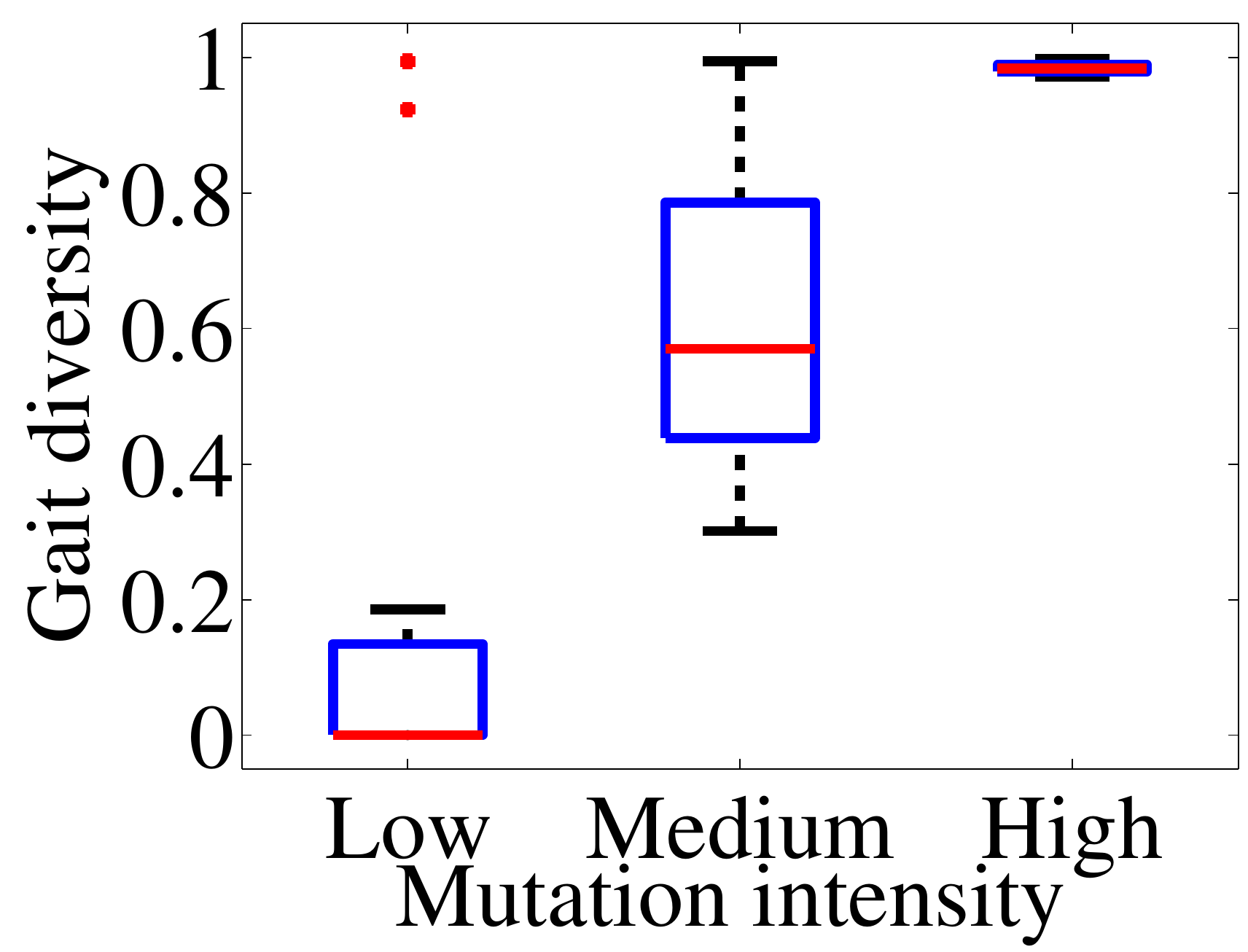}}
\subfloat[ANN]{\includegraphics[width=3.4cm]{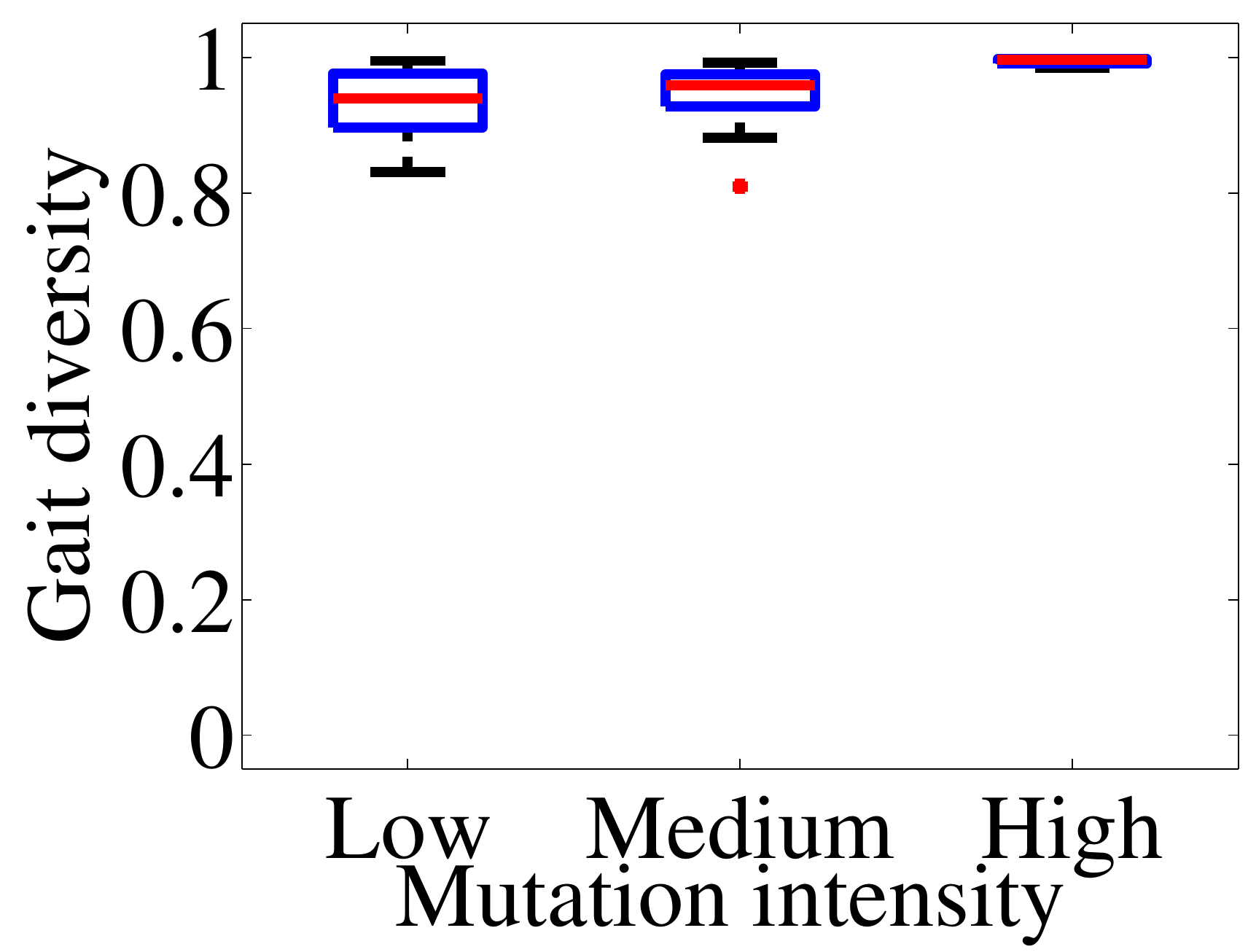}}
\subfloat[SUPG]{\includegraphics[width=3.4cm]{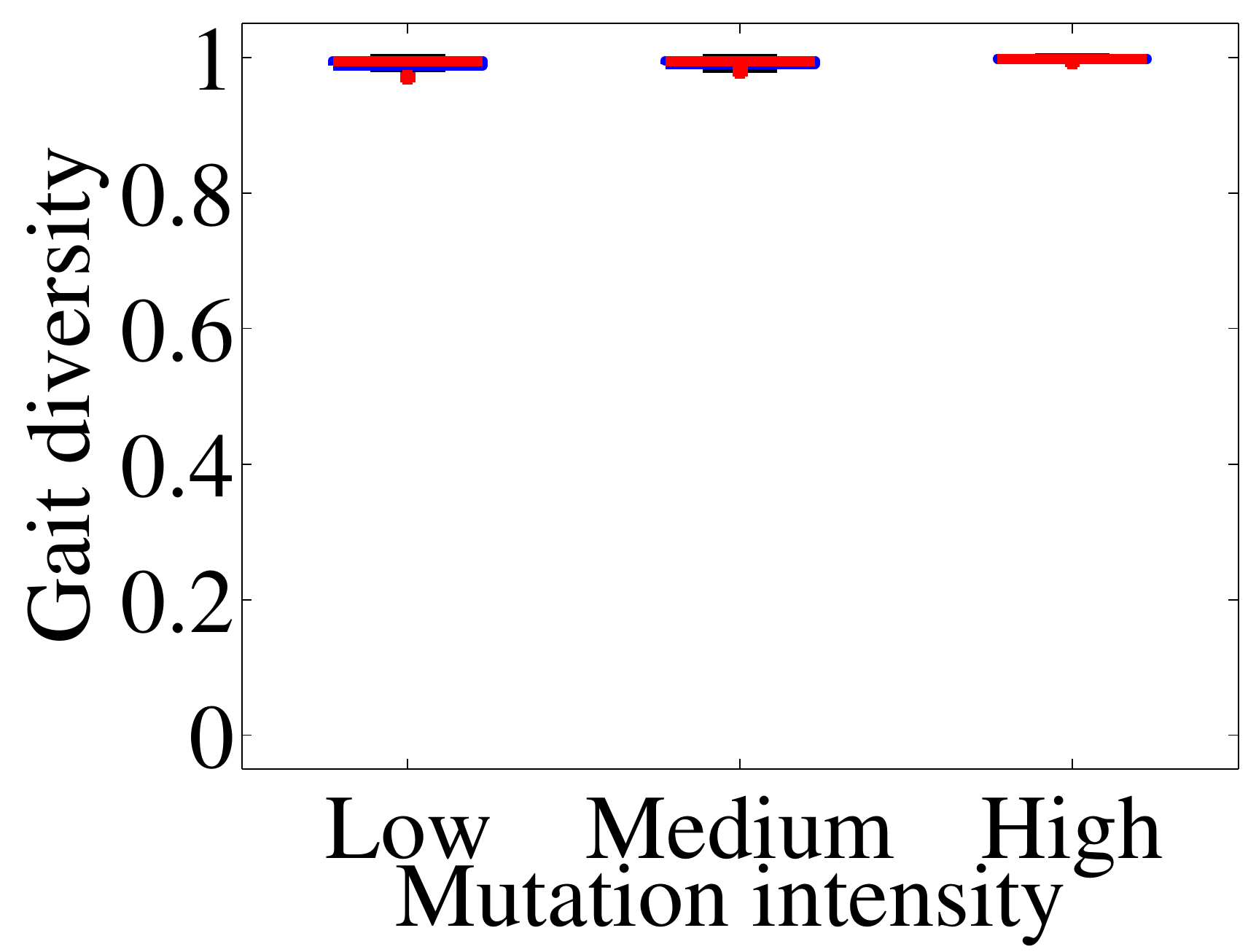}}\\
\subfloat[Direct]{\includegraphics[width=3.4cm]{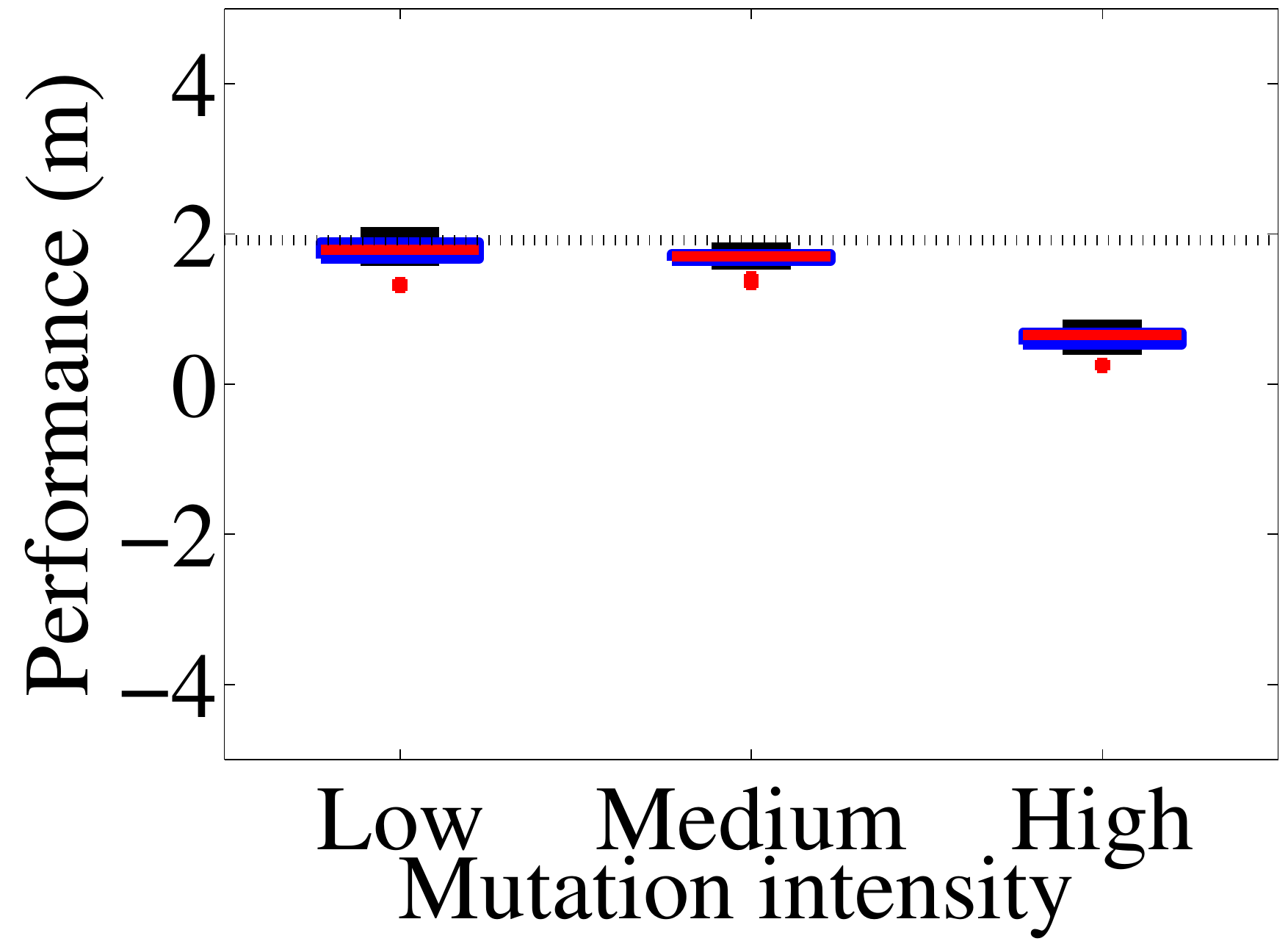}}
\subfloat[CPG]{\includegraphics[width=3.4cm]{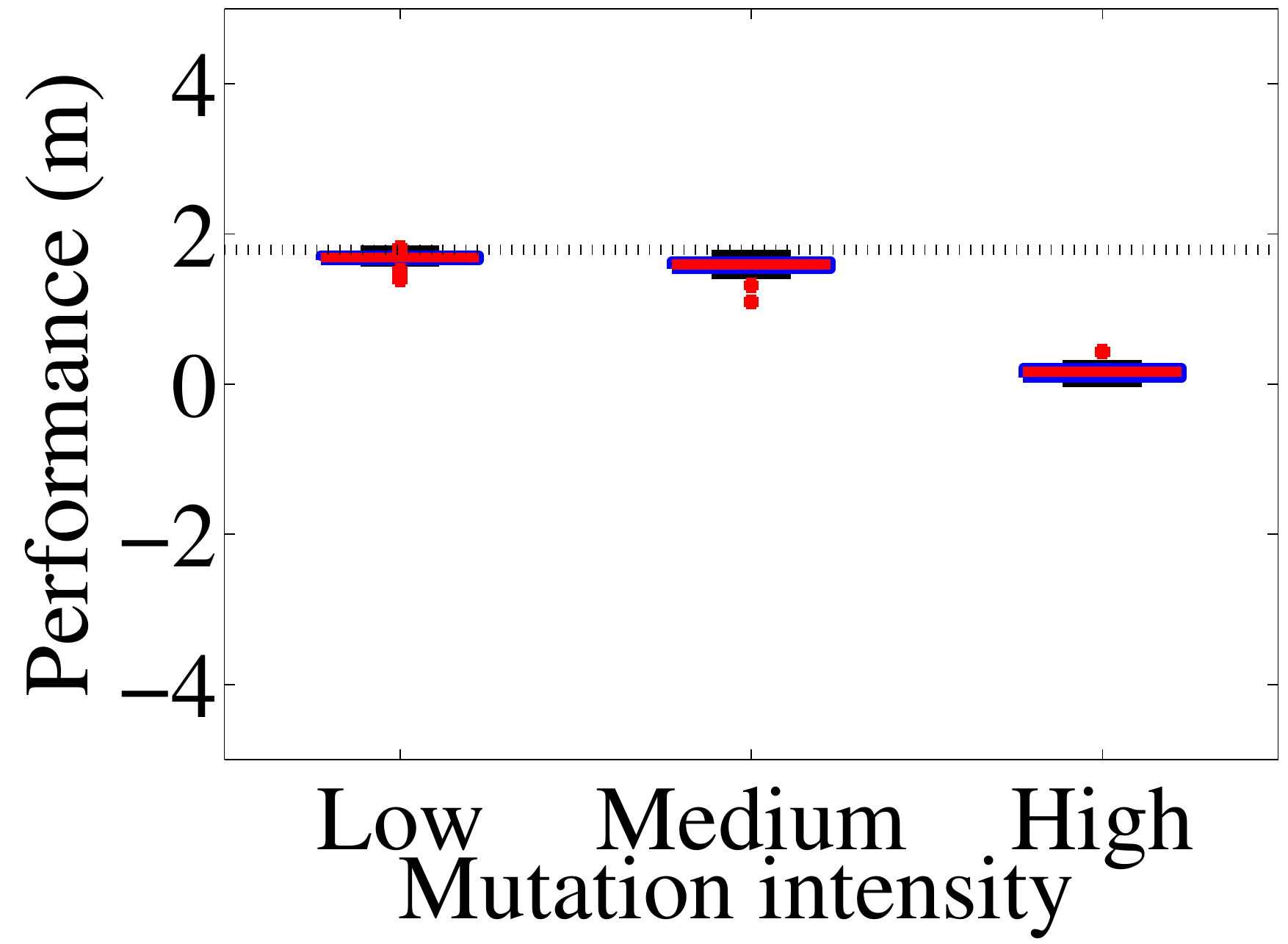}}
\subfloat[CPG-f/b]{\includegraphics[width=3.4cm]{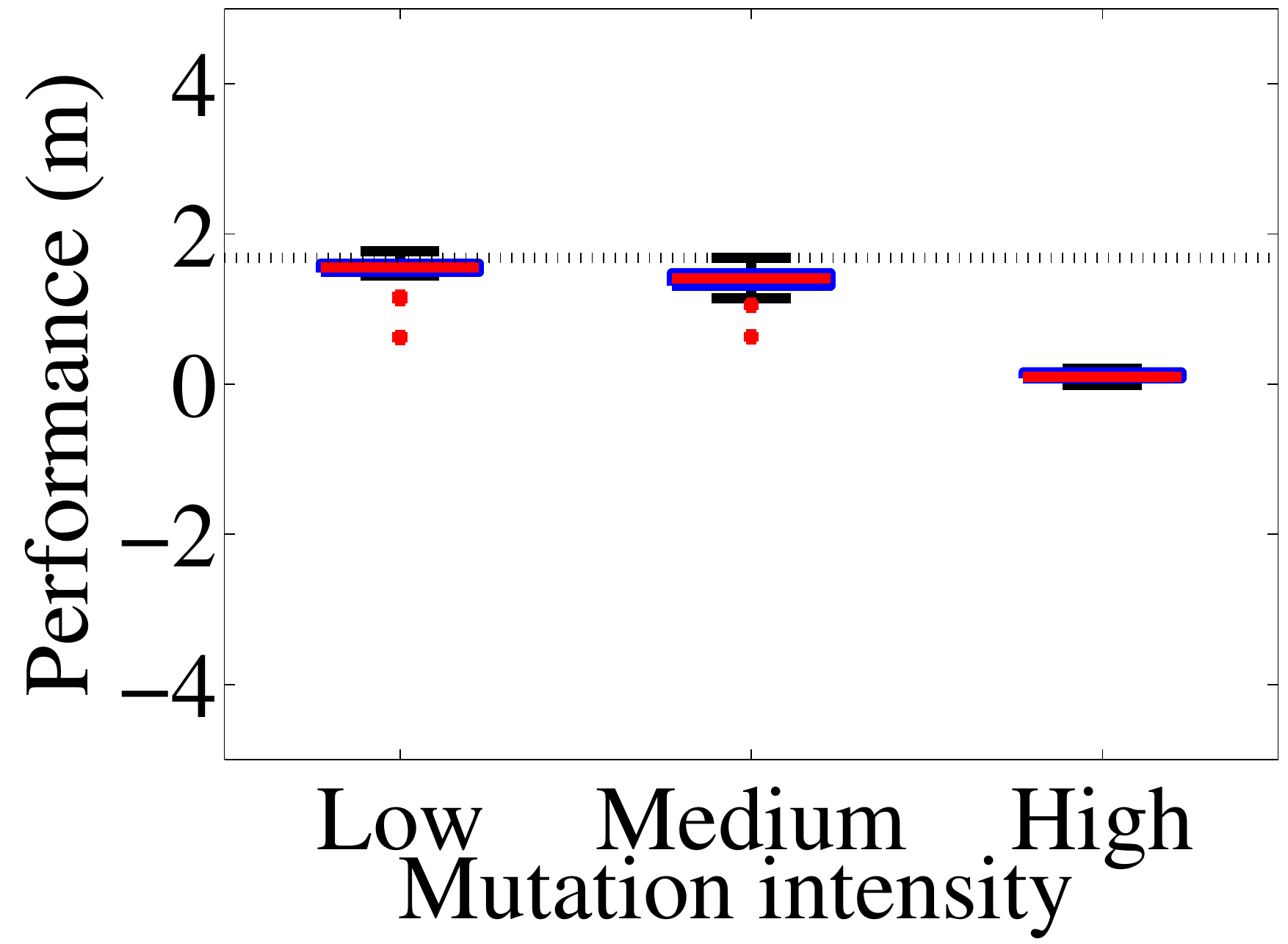}}
\subfloat[ANN]{\includegraphics[width=3.4cm]{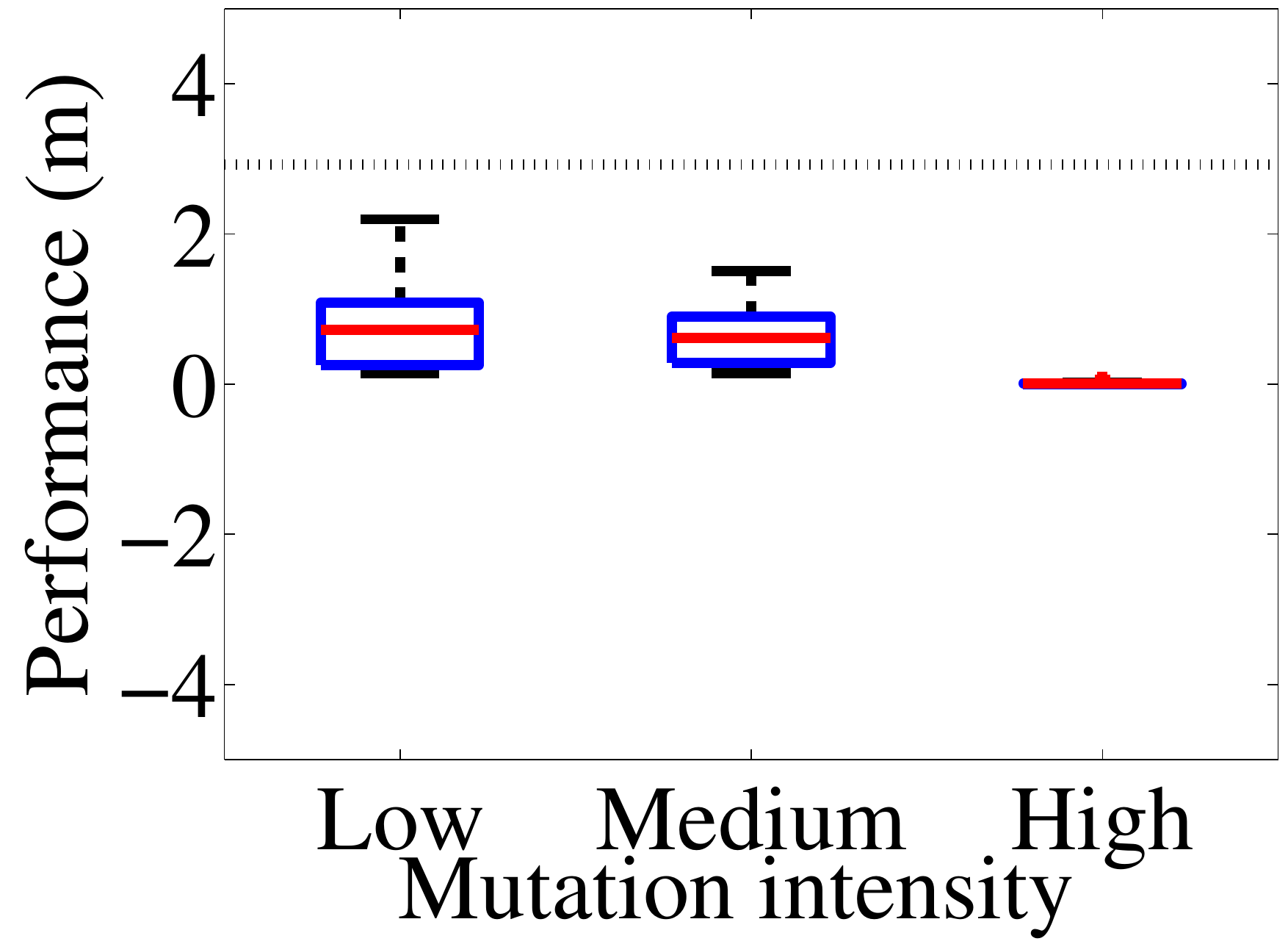}}
\subfloat[SUPG]{\includegraphics[width=3.4cm]{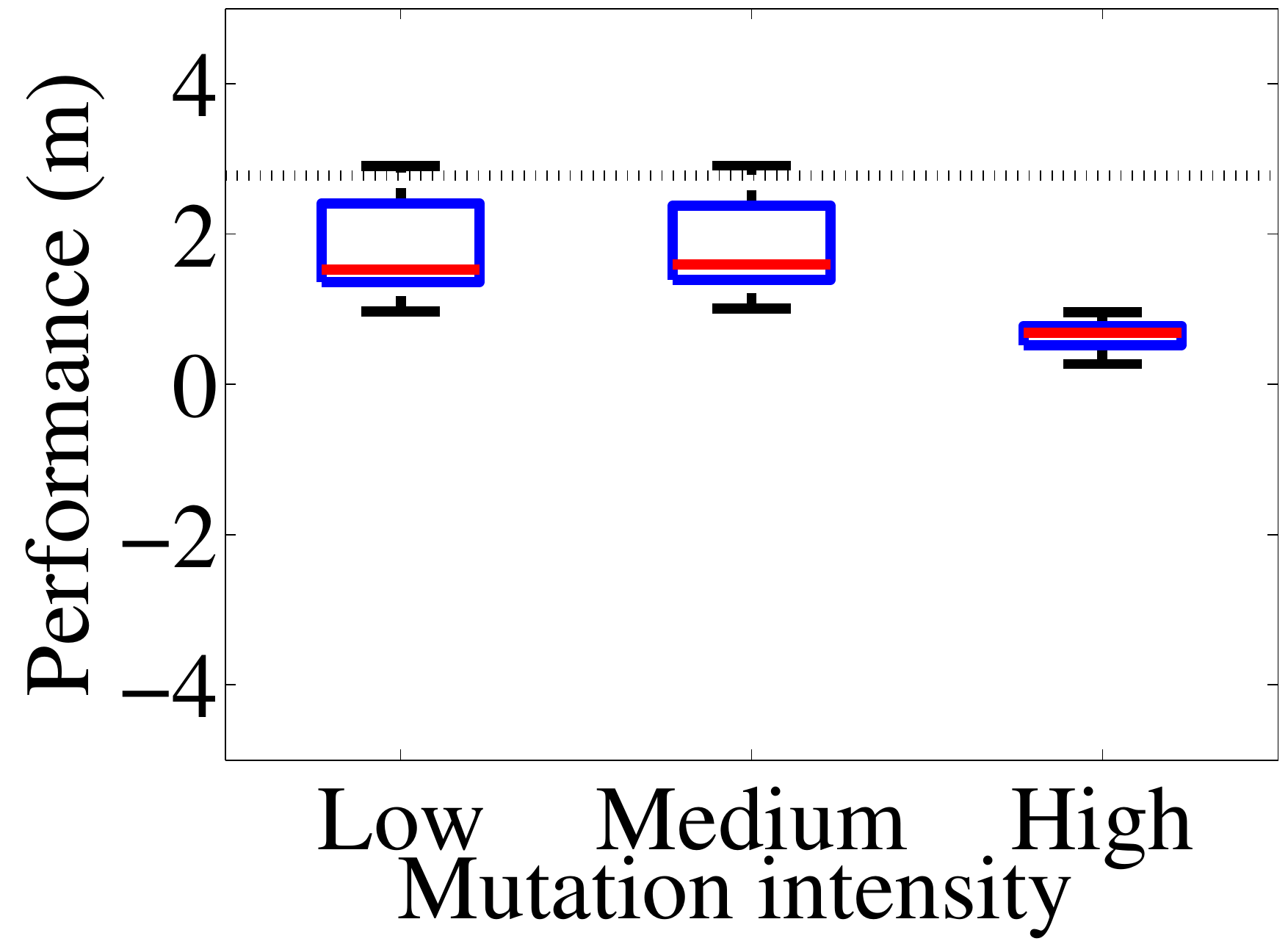}}
\caption{Mutant performance in forward displacement and the gait diversity resulting from mutation, for the Direct, CPG,
CPG-f/b, ANN and SUPG encoding schemes. Each box corresponds to the median of $1,000$ mutants, in each of $20$ replicates.
Mutations were generated at low ($\times 0.25$), medium ($\times 1$) and high ($\times 4$) times mutation rate and step-size
used during selection, and applied to the best individuals at generation $8,000$ of selection. Dotted lines (f-j) indicate
the median performance of best individuals across $20$ replicates, after $8,000$ generation of selection.}
\label{fig:evolvability_1D}
\end{figure*}

\begin{figure*}[h]
\centering
\subfloat[Scenario 1: Removal of right-middle leg.]{\includegraphics[width=0.3\textwidth, height=0.237\textwidth]{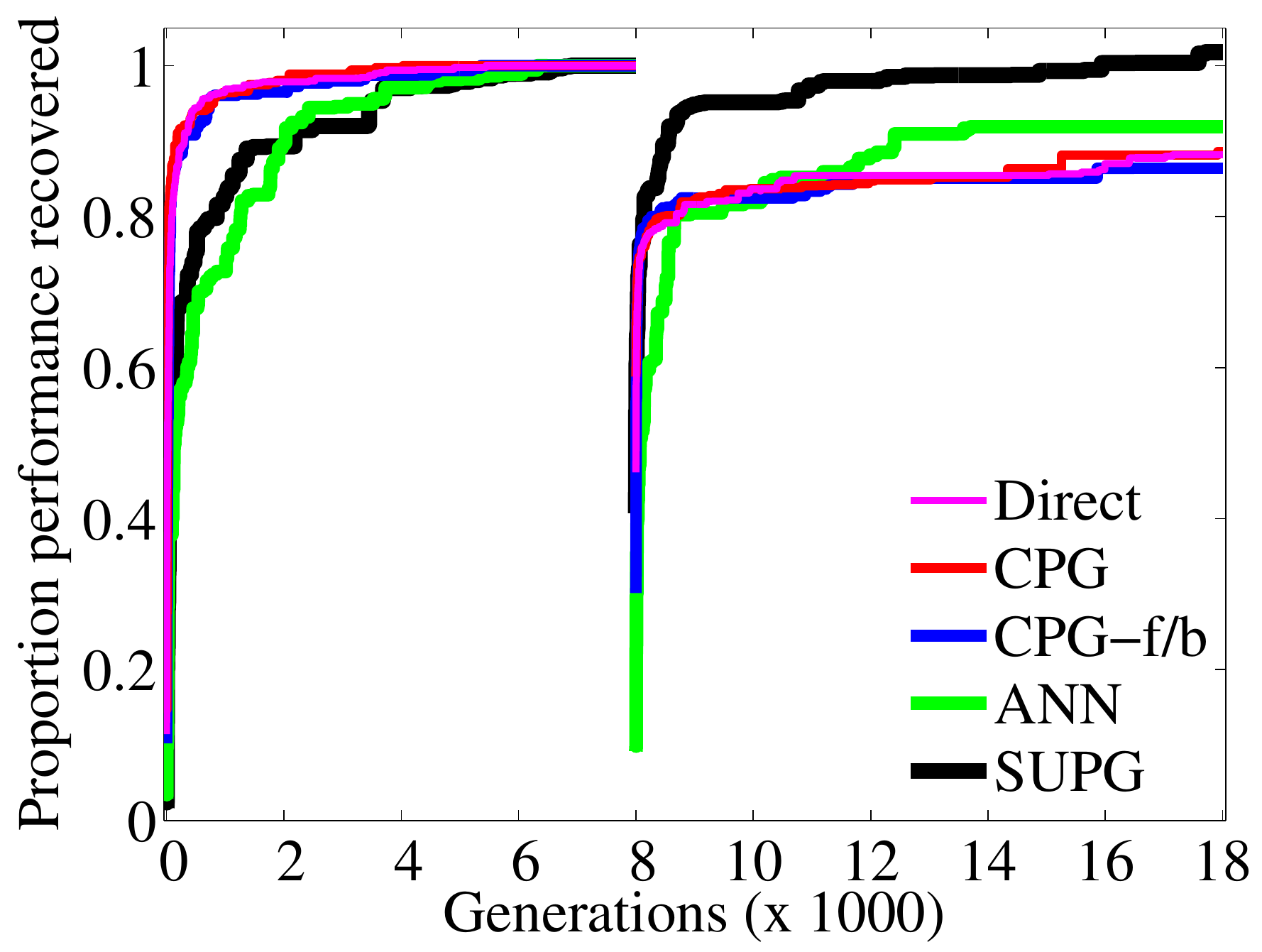}}\hfil
\subfloat[Scenario 2: Removal of right-middle and left-middle legs.]{\includegraphics[width=0.3\textwidth, height=0.237\textwidth]{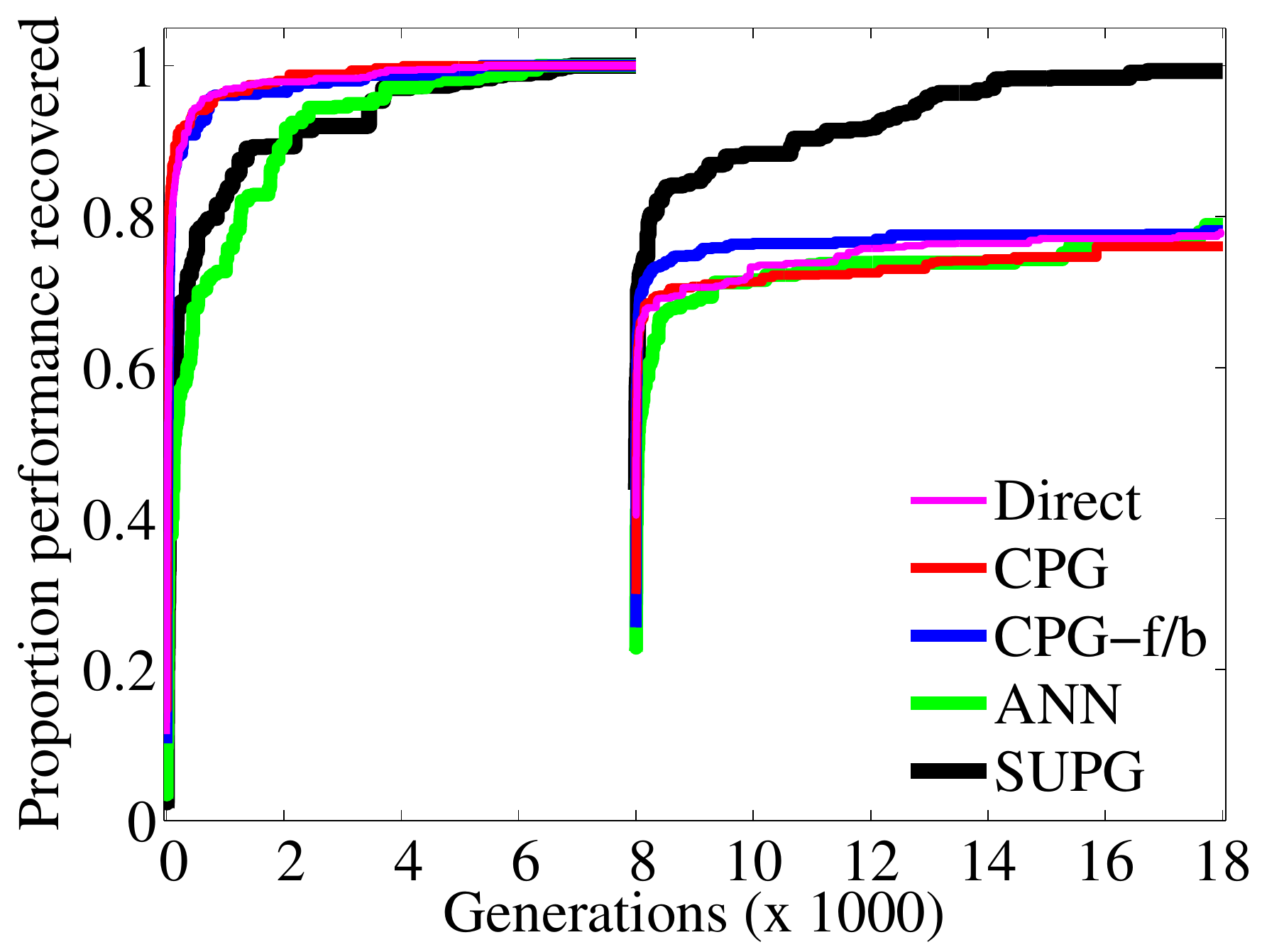}}\hfil
\subfloat[Scenario 3: Removal of right-middle and left-rear legs.]{\includegraphics[width=0.3\textwidth, height=0.237\textwidth]{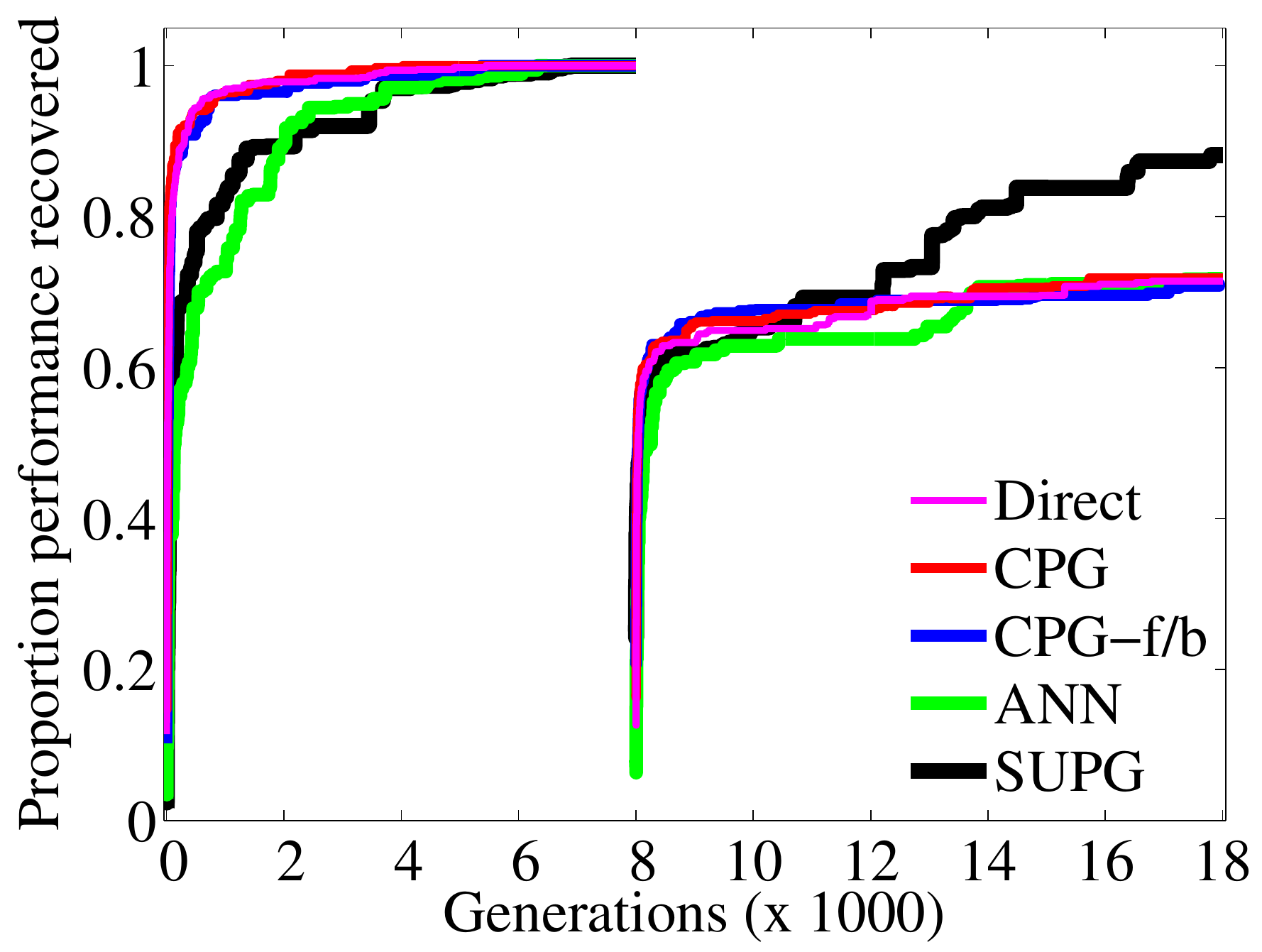}}
\caption{The proportion of performance in forward displacement before (generations $0$ to $8,000$) and after (generations $8001$ to $18,000$) the three robot-damage scenarios, for the Direct, CPG, CPG-f/b, ANN and SUPG encoding schemes.}
\label{fig:recov_performance_timeline}
\end{figure*}

\newpage
\begin{table*}[h]
\caption{Performance comparison of the Direct, Open-loop CPG, Closed-loop CPG, HyperNEAT, and SUPG encoding schemes.\label{tab:performance}}
\centering
\begin{scriptsize}
\setlength{\extrarowheight}{3pt}
\begin{tabular}{||p{3cm}|p{2cm}|p{2cm}||} 
\hline
\textbf{Encodings} & \textbf{$p$-Value}\footnotesize{*} & \textbf{Effect size}\footnotesize{**}\\
\hline
Direct &  \multirow{2}{*}{$8.0 \times 10^{-5}$} & \multirow{2}{*}{$0.85$}\\
\cline{1-1}
Open-loop CPG &&\\
\hline

Direct &  \multirow{2}{*}{$7.0 \times 10^{-7}$} & \multirow{2}{*}{$0.90$}\\
\cline{1-1}
Closed-loop CPG &&\\
\hline

Direct &  \multirow{2}{*}{$0.0142$} & \multirow{2}{*}{$0.30$}\\
\cline{1-1}
HyperNEAT &&\\
\hline

Direct &  \multirow{2}{*}{$1.2 \times 10^{-7}$} & \multirow{2}{*}{$0.02$}\\
\cline{1-1}
SUPG &&\\
\hline

Open-loop CPG &  \multirow{2}{*}{$3.0 \times 10^{-4}$} & \multirow{2}{*}{$0.81$}\\
\cline{1-1}
Closed-loop CPG &&\\
\hline

Open-loop CPG &  \multirow{2}{*}{$4.2 \times 10^{-4}$} & \multirow{2}{*}{$0.21$}\\
\cline{1-1}
HyperNEAT &&\\
\hline

Open-loop CPG &  \multirow{2}{*}{$1.7 \times 10^{-8}$} & \multirow{2}{*}{$0.0$}\\
\cline{1-1}
SUPG &&\\
\hline

Closed-loop CPG &  \multirow{2}{*}{$4.8 \times 10^{-6}$} & \multirow{2}{*}{$0.09$}\\
\cline{1-1}
HyperNEAT &&\\
\hline

Closed-loop CPG &  \multirow{2}{*}{$3.4 \times 10^{-8}$} & \multirow{2}{*}{$0.0$}\\
\cline{1-1}
SUPG &&\\
\hline

HyperNEAT &  \multirow{2}{*}{$0.1860$} & \multirow{2}{*}{$0.42$}\\
\cline{1-1}
SUPG &&\\
\hline
\multicolumn{3}{p{8cm}}{\footnotesize *Significance computed with the Mann-Whitney-Wilcoxon nonparametric test, at $0.05$ significance level.}\\
\multicolumn{3}{p{8cm}}{\footnotesize **Effect size estimated as ratio of Mann-Whitney $U$ statistic and the product of two sample sizes.}
\end{tabular}
\end{scriptsize}
\end{table*}

\begin{table*}[h]
\caption{Comparison of the number of beneficial mutations generated by the Direct, Open-loop CPG, Closed-loop CPG, HyperNEAT, and SUPG encoding schemes.\label{tab:benef_mut}}
\centering
\begin{scriptsize}
\setlength{\extrarowheight}{3pt}
\begin{tabular}{||p{3cm}|p{2cm}|p{2cm}||} 
\hline
\textbf{Encodings} & \textbf{$p$-Value}\footnotesize{*} & \textbf{Effect size}\footnotesize{**}\\
\hline
Direct &  \multirow{2}{*}{$0.0304$} & \multirow{2}{*}{$0.30$}\\
\cline{1-1}
Open-loop CPG &&\\
\hline

Direct &  \multirow{2}{*}{$1.5 \times 10^{-5}$} & \multirow{2}{*}{$0.10$}\\
\cline{1-1}
Closed-loop CPG &&\\
\hline

Direct &  \multirow{2}{*}{$6.8 \times 10^{-8}$} & \multirow{2}{*}{$0.0$}\\
\cline{1-1}
HyperNEAT &&\\
\hline

Direct &  \multirow{2}{*}{$6.7 \times 10^{-8}$} & \multirow{2}{*}{$0.0$}\\
\cline{1-1}
SUPG &&\\
\hline

Open-loop CPG &  \multirow{2}{*}{$5.1 \times 10^{-4}$} & \multirow{2}{*}{$0.18$}\\
\cline{1-1}
Closed-loop CPG &&\\
\hline

Open-loop CPG &  \multirow{2}{*}{$6.8 \times 10^{-8}$} & \multirow{2}{*}{$0.0$}\\
\cline{1-1}
HyperNEAT &&\\
\hline

Open-loop CPG &  \multirow{2}{*}{$6.8 \times 10^{-8}$} & \multirow{2}{*}{$0.0$}\\
\cline{1-1}
SUPG &&\\
\hline

Closed-loop CPG &  \multirow{2}{*}{$2.5 \times 10^{-4}$} & \multirow{2}{*}{$0.16$}\\
\cline{1-1}
HyperNEAT &&\\
\hline

Closed-loop CPG &  \multirow{2}{*}{$3.4 \times 10^{-7}$} & \multirow{2}{*}{$0.03$}\\
\cline{1-1}
SUPG &&\\
\hline

HyperNEAT &  \multirow{2}{*}{$9.1 \times 10^{-8}$} & \multirow{2}{*}{$0.01$}\\
\cline{1-1}
SUPG &&\\
\hline
\multicolumn{3}{p{8cm}}{\footnotesize *Significance computed with the Mann-Whitney-Wilcoxon nonparametric test, at $0.05$ significance level.}\\
\multicolumn{3}{p{8cm}}{\footnotesize **Effect size estimated as ratio of Mann-Whitney $U$ statistic and the product of two sample sizes.}

\end{tabular}
\end{scriptsize}
\end{table*}

\begin{table*}[h]
\caption{Comparison of number of generations of selection required to restore $85\%$ of the original performance of the undamaged hexapod, for the Direct, CPG, CPG-f/b, ANN, and SUPG encoding schemes, in the first damage scenario (removal of right-middle leg).\label{tab:recov_dmg1}}
\centering
\begin{scriptsize}
\setlength{\extrarowheight}{3pt}
\begin{tabular}{||p{3cm}|p{2cm}|p{2cm}||} 
\hline
\textbf{Encodings} & \textbf{$p$-Value}\footnotesize{*} & \textbf{Effect size}\footnotesize{**}\\
\hline
Direct &  \multirow{2}{*}{$0.4182$} & \multirow{2}{*}{$0.48$}\\
\cline{1-1}
Open-loop CPG &&\\
\hline
Direct &  \multirow{2}{*}{$0.3538$} & \multirow{2}{*}{$0.47$}\\
\cline{1-1}
Closed-loop CPG &&\\
\hline
Direct &  \multirow{2}{*}{$0.4393$} & \multirow{2}{*}{$0.52$}\\
\cline{1-1}
HyperNEAT &&\\
\hline
Direct &  \multirow{2}{*}{$0.0003$} & \multirow{2}{*}{$0.81$}\\
\cline{1-1}
SUPG &&\\
\hline
Open-loop CPG &  \multirow{2}{*}{$0.4505$} & \multirow{2}{*}{$0.49$}\\
\cline{1-1}
Closed-loop CPG &&\\
\hline
Open-loop CPG &  \multirow{2}{*}{$0.3045$} & \multirow{2}{*}{$0.55$}\\
\cline{1-1}
HyperNEAT &&\\
\hline
Open-loop CPG &  \multirow{2}{*}{$0.0010$} & \multirow{2}{*}{$0.81$}\\
\cline{1-1}
SUPG &&\\
\hline
Closed-loop CPG &  \multirow{2}{*}{$0.0045$} & \multirow{2}{*}{$0.55$}\\
\cline{1-1}
HyperNEAT &&\\
\hline
Closed-loop CPG &  \multirow{2}{*}{$3.4 \times 10^{-7}$} & \multirow{2}{*}{$0.79$}\\
\cline{1-1}
SUPG &&\\
\hline
HyperNEAT &  \multirow{2}{*}{$9.1 \times 10^{-8}$} & \multirow{2}{*}{$0.74$}\\
\cline{1-1}
SUPG &&\\
\hline
\multicolumn{3}{p{8cm}}{\footnotesize *Significance computed with the Mann-Whitney-Wilcoxon nonparametric test, at $0.05$ significance level.}\\
\multicolumn{3}{p{8cm}}{\footnotesize **Effect size estimated as ratio of Mann-Whitney $U$ statistic and the product of two sample sizes.}
\end{tabular}
\end{scriptsize}
\end{table*}

\begin{table*}[h]
\caption{Comparison of number of generations of selection required to restore $85\%$ of the original performance of the undamaged hexapod, for the Direct, CPG, CPG-f/b, ANN, and SUPG encoding schemes, in the second damage scenario (removal of right-middle and left-middle legs).\label{tab:recov_dmg3}}
\centering
\begin{scriptsize}
\setlength{\extrarowheight}{3pt}
\begin{tabular}{||p{3cm}|p{2cm}|p{2cm}||} 
\hline
\textbf{Encodings} & \textbf{$p$-Value}\footnotesize{*} & \textbf{Effect size}\footnotesize{**}\\
\hline
Direct &  \multirow{2}{*}{$0.4768$} & \multirow{2}{*}{$0.50$}\\
\cline{1-1}
Open-loop CPG &&\\
\hline
Direct &  \multirow{2}{*}{$0.1979$} & \multirow{2}{*}{$0.45$}\\
\cline{1-1}
Closed-loop CPG &&\\
\hline
Direct &  \multirow{2}{*}{$0.0286$} & \multirow{2}{*}{$0.65$}\\
\cline{1-1}
HyperNEAT &&\\
\hline
Direct &  \multirow{2}{*}{$3.9 \times 10^{-6}$} & \multirow{2}{*}{$0.90$}\\
\cline{1-1}
SUPG &&\\
\hline
Open-loop CPG &  \multirow{2}{*}{$0.2231$} & \multirow{2}{*}{$0.46$}\\
\cline{1-1}
Closed-loop CPG &&\\
\hline
Open-loop CPG &  \multirow{2}{*}{$0.0286$} & \multirow{2}{*}{$0.65$}\\
\cline{1-1}
HyperNEAT &&\\
\hline
Open-loop CPG &  \multirow{2}{*}{$1.8 \times 10^{-6}$} & \multirow{2}{*}{$0.91$}\\
\cline{1-1}
SUPG &&\\
\hline
Closed-loop CPG &  \multirow{2}{*}{$0.0052$} & \multirow{2}{*}{$0.69$}\\
\cline{1-1}
HyperNEAT &&\\
\hline
Closed-loop CPG &  \multirow{2}{*}{$4.8 \times 10^{-7}$} & \multirow{2}{*}{$0.93$}\\
\cline{1-1}
SUPG &&\\
\hline
HyperNEAT &  \multirow{2}{*}{$0.0026$} & \multirow{2}{*}{$0.76$}\\
\cline{1-1}
SUPG &&\\
\hline
\multicolumn{3}{p{8cm}}{\footnotesize *Significance computed with the Mann-Whitney-Wilcoxon nonparametric test, at $0.05$ significance level.}\\
\multicolumn{3}{p{8cm}}{\footnotesize **Effect size estimated as ratio of Mann-Whitney $U$ statistic and the product of two sample sizes.}
\end{tabular}
\end{scriptsize}
\end{table*}

\begin{table*}[h]
\caption{Comparison of number of generations of selection required to restore $85\%$ of the original performance of the undamaged hexapod, for the Direct, CPG, CPG-f/b, ANN, and SUPG encoding schemes, in the third damage scenario (removal of right-middle and left-rear legs).\label{tab:recov_dmg2}}
\centering
\begin{scriptsize}
\setlength{\extrarowheight}{3pt}
\begin{tabular}{||p{3cm}|p{2cm}|p{2cm}||} 
\hline
\textbf{Encodings} & \textbf{$p$-Value}\footnotesize{*} & \textbf{Effect size}\footnotesize{**}\\
\hline
Direct &  \multirow{2}{*}{$0.1711$} & \multirow{2}{*}{$0.48$}\\
\cline{1-1}
Open-loop CPG &&\\
\hline
Direct &  \multirow{2}{*}{$0.1711$} & \multirow{2}{*}{$0.47$}\\
\cline{1-1}
Closed-loop CPG &&\\
\hline
Direct &  \multirow{2}{*}{$0.0082$} & \multirow{2}{*}{$0.67$}\\
\cline{1-1}
HyperNEAT &&\\
\hline
Direct &  \multirow{2}{*}{$0.0001$} & \multirow{2}{*}{$0.78$}\\
\cline{1-1}
SUPG &&\\
\hline
Open-loop CPG &  \multirow{2}{*}{$0.5$} & \multirow{2}{*}{$0.50$}\\
\cline{1-1}
Closed-loop CPG &&\\
\hline
Open-loop CPG &  \multirow{2}{*}{$0.0023$} & \multirow{2}{*}{$0.67$}\\
\cline{1-1}
HyperNEAT &&\\
\hline
Open-loop CPG &  \multirow{2}{*}{$3.3 \times 10^{-5}$} & \multirow{2}{*}{$0.80$}\\
\cline{1-1}
SUPG &&\\
\hline
Closed-loop CPG &  \multirow{2}{*}{$0.0023$} & \multirow{2}{*}{$0.68$}\\
\cline{1-1}
HyperNEAT &&\\
\hline
Closed-loop CPG &  \multirow{2}{*}{$3.3 \times 10^{-5}$} & \multirow{2}{*}{$0.80$}\\
\cline{1-1}
SUPG &&\\
\hline
HyperNEAT &  \multirow{2}{*}{$0.2460$} & \multirow{2}{*}{$0.56$}\\
\cline{1-1}
SUPG &&\\
\hline
\multicolumn{3}{p{8cm}}{\footnotesize *Significance computed with the Mann-Whitney-Wilcoxon nonparametric test, at $0.05$ significance level.}\\
\multicolumn{3}{p{8cm}}{\footnotesize **Effect size estimated as ratio of Mann-Whitney $U$ statistic and the product of two sample sizes.}
\end{tabular}
\end{scriptsize}
\end{table*}

\end{document}